%% file: acl_latex.tex
\documentclass[11pt]{article}

\usepackage[preprint]{acl}

\usepackage{times}
\usepackage{latexsym}
\usepackage{microtype}
\usepackage{graphicx}
\usepackage{subcaption}
\usepackage{booktabs} %

\usepackage[T1]{fontenc}

\usepackage[utf8]{inputenc}

\usepackage{microtype}

\usepackage{inconsolata}

\usepackage{graphicx}
\usepackage{amsmath}
\usepackage{amssymb}
\usepackage{mathtools}
\usepackage{amsthm}
\usepackage[table]{xcolor}
\usepackage{multirow}

\usepackage{caption}
\usepackage{subcaption} 

\usepackage{cuted}      
\usepackage[utf8]{inputenc}
\usepackage[most]{tcolorbox}
\tcbuselibrary{breakable}

\usepackage{tikz}

\usepackage{makecell}

\usepackage{titletoc}  
\usepackage{hyperref}  

\newcommand{\vpara}[1]{\vspace{0.5em}\noindent\textbf{#1}\quad}
\usepackage{amssymb}
\newcommand{\cmark}{\checkmark}
\newcommand{\xmark}{\texttimes}

\usepackage[capitalize,noabbrev]{cleveref}

\title{LongWebBench: Evaluating Structural and Functional Webpage Generation in Long-Horizon Settings}

\author{
 \textbf{Yi Zhao\textsuperscript{1*}},
 \textbf{Zhen Yang\textsuperscript{1*}},
 \textbf{Mengpan Chen\textsuperscript{2}},
\textbf{Mingde Xu\textsuperscript{3}},
 \textbf{Shanghui Gong\textsuperscript{2}},
 \\
 \textbf{Xijun Liu\textsuperscript{4}},
 \textbf{Jibing Gong\textsuperscript{2$\dagger$}},
 \textbf{Jie Tang\textsuperscript{1$\dagger$}},
\\
 \textsuperscript{1}Tsinghua University,
 \textsuperscript{2}Yanshan University,
 \textsuperscript{3}University of Waterloo,
  \textsuperscript{4}Beihang University
}

\begin{document}
\maketitle

\begin{abstract}

    Recent vision-language models (VLMs) have shown promising progress in generating webpages from visual inputs, yet existing evaluations mainly focus on short, single-screen, and largely static webpages. 
    We introduce \textbf{LongWebBench}, a benchmark for evaluating long-horizon webpage generation from both structural and functional perspectives. 
    LongWebBench contains 490 real-world long webpages for structural fidelity evaluation and 507 goal-oriented interaction tasks over 129 webpages for functional evaluation. It employs two complementary protocols: a multi-dimensional VLM-based metric for assessing long-range structural coherence, and a DOM-augmented agent-based pipeline for end-to-end functional verification. We further examine the automatic evaluation protocols through human agreement analysis.
    Experiments with state-of-the-art open-source and proprietary VLMs under single-image and multi-image settings reveal that structural fidelity degrades as webpage length increases, while visually plausible generations often fail to support executable multi-step interactions. 
    These results highlight the need to evaluate long webpage generation beyond visual similarity, with executable interaction as a core criterion.
    Our code and data are available at \url{https://github.com/zheny2751-dotcom/LongWebBench}.
\end{abstract}

\let\thefootnote\relax\footnotetext{
$^*$ Equal contributions}

{\let\thefootnote\relax\footnotetext{
$^\dagger$ Corresponding authors}
}
\input{0-introduction}

\input{2-bench}

\input{3-evaluation-pipeline}

\input{4-experiments}

\input{5-related-works}

\input{6-conclusion}

\clearpage

\section*{Limitations}
LongWebBench evaluates stabilized rendered webpages and frontend-observable interactions. This design makes the benchmark controllable and reproducible, but excludes authentication, personalization, live backend services, and unbounded infinite scrolling. Therefore, the benchmark targets long-page visual structure and executable frontend behavior rather than the full complexity of deployed web applications.
The functional tasks in W-FFR are restricted to user goals whose success can be verified through browser execution, DOM states, and rendered page states. Tasks requiring external databases, payment systems, account-specific states, or hidden server-side logic are outside the current scope. In addition, although we validate the automatic evaluators through human agreement analysis and cross-judge robustness checks, automatic evaluation may still miss subtle visual or behavioral errors. Future work can extend LongWebBench toward controlled backend environments and broader real-world interaction scenarios.




\bibliography{references}

\appendix


\input{8-appendix}

\end{document}

%% file: 0-introduction.tex
\section{Introduction}

Recent vision-language models (VLMs)~\cite{liu2023visual,wang2024cogvlm,yang2025qwen3,wang2025internvl3,team2025kimi} have shown promising progress in generating front-end code from visual inputs such as webpage screenshots. Given a rendered interface, these models can produce HTML, CSS, and JavaScript that visually resemble the target page~\cite{jiang2025screencoder,liang2025waffle}. However, existing efforts~\cite{yun2024web2code,tan2025chartmaster,niu2025chart2code53,zhao2025chartcoder,wu2025mllm} largely focus on short, single-screen, and mostly static webpages, where success is primarily measured by local visual similarity. Such settings provide limited evidence about whether models can generate long webpages that preserve global structure and support executable user interactions.

\begin{figure}[t!]
    \centering
    \includegraphics[width=\linewidth]{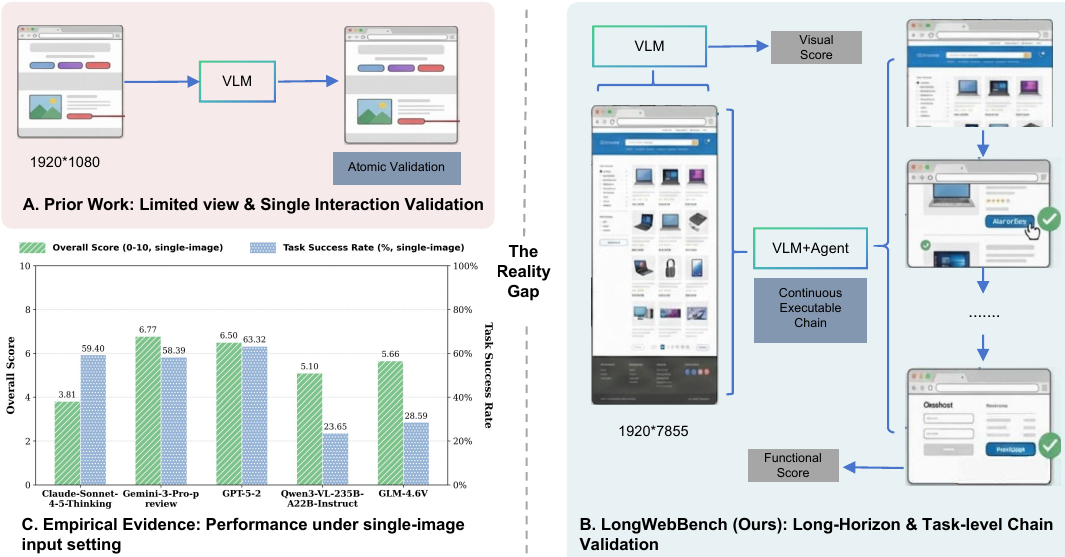}
    \caption{Motivation of LongWebBench. Existing evaluations mainly consider limited-view reconstruction or isolated interactions, while LongWebBench evaluates long webpages through both structural fidelity and task-level functional verification. Empirical results under the single-image setting illustrate that current VLMs still face challenges in both long-range visual reconstruction and executable task completion.}
    \vspace{-6mm}
    \label{fig:generate_result}
\end{figure}

\begin{table*}[t!]
    \centering
    \caption{Comparison with existing webpage-to-code benchmarks. Existing benchmarks mainly focus on short-page visual validation or isolated interaction validation, while LongWebBench supports long webpages, multi-image inputs, and goal-driven functional validation.}
    \resizebox{\textwidth}{!}{
    \begin{tabular}{c|c|c|c|c|c}
    \toprule
    Benchmark & Visual Validation & Single-Interaction Validation & Functional Validation & Long-WebPage  Support & Multi-Image Support
    \\
    \midrule
        Web2Code   \cite{yun2024web2code} & \cmark & \xmark & \xmark  & \xmark  & \xmark  \\
        Interaction2Code \cite{xiao2024interaction2code} & \cmark & \cmark  & \xmark  & \xmark  & \xmark  \\
        Webcode2M \cite{gui2025webcode2m} & \cmark &\xmark  & \xmark  & \xmark  & \xmark \\
        Design2Code \cite{si2025design2code} & \cmark & \xmark & \xmark  & \xmark  & \xmark  \\
        WebGen-Bench \cite{lu2025webgen}  & \cmark & \cmark & \cmark  & \xmark  & \xmark  \\
        Sketch2Code \cite{li2025sketch2code} & \cmark & \xmark & \xmark  & \xmark  & \xmark \\
        \rowcolor{blue!8}
        LongWebBench  & \cmark & \cmark  & \cmark  & \cmark  & \cmark \\
    \bottomrule
    \end{tabular}
    }
    \label{tab:comparison_with_others}
    \vspace{-3mm}
\end{table*}

As illustrated in Figure~\ref{fig:generate_result}, this results in an evaluation gap between limited-view webpage reconstruction and long-horizon webpage generation with executable interactions.
Real-world webpages often span multiple screens, contain repeated components, require style consistency across distant sections, and support multi-step user goals. 
These properties introduce two challenges beyond short static reconstruction. Structurally, models must preserve page scale, global layout, section hierarchy, visual styling, and information density across the full webpage. Functionally, models must synthesize interaction logic that supports user goals under browser execution. 
A page may look plausible while still failing to open menus, filter content, navigate across sections, or submit forms, suggesting that long-horizon webpage generation should be evaluated beyond static visual similarity.

Despite growing interest in webpage-to-code evaluation, existing benchmarks do not jointly cover these requirements. As summarized in Table~\ref{tab:comparison_with_others}, prior benchmarks mainly emphasize visual fidelity on short webpages~\cite{yun2024web2code,gui2025webcode2m,si2025design2code,li2025sketch2code} or isolated interaction validation~\cite{xiao2024interaction2code}. Although WebGen-Bench~\cite{lu2025webgen} evaluates goal-driven functionality, it does not target screenshot-driven long webpage generation or long-scroll structural consistency. Consequently, existing benchmarks do not jointly assess global structure across long webpages and executable multi-step user interactions.

To address this gap, we introduce \textbf{LongWebBench}, a benchmark for evaluating long-horizon webpage generation from both structural and functional perspectives. LongWebBench contains two complementary tasks. \textit{Webpage Visual Fidelity Replication} (W-VFR) evaluates whether a model can reproduce the global structure and visual organization of long webpages, using 490 real-world long webpages across diverse categories. \textit{Webpage Functional Fidelity Realization} (W-FFR) evaluates whether a model can generate executable webpages that support goal-oriented user interactions, using 507 interaction tasks over 129 webpages. These two tasks separate visual fidelity from functional correctness, enabling a more diagnostic evaluation of webpage generation.

LongWebBench further provides corresponding evaluation protocols: a multi-dimensional VLM-based metric for long-range structural coherence, and a DOM-augmented agent-based pipeline that executes generated webpages in a browser environment to verify user-goal completion. For structural evaluation, we additionally incorporate DINO-based feature similarity as an auxiliary signal for local visual styling consistency. The benchmark supports both single-image and multi-image input settings, allowing us to analyze model behavior under different long-page input constraints.

We evaluate state-of-the-art open-source and proprietary VLMs on LongWebBench under both single-image and multi-image input settings. Results show that long-horizon webpage generation remains challenging: structural fidelity generally degrades as webpage length increases, while visually plausible generations often fail to support executable multi-step interactions. These findings highlight the need to evaluate webpage generation beyond static visual similarity, with executable interaction as a core criterion.

Our contributions are summarized as follows:
\begin{itemize}
    \item We introduce \textbf{LongWebBench}, a benchmark for long-horizon webpage generation with structural and functional evaluation.

    \item We construct two complementary tasks: W-VFR with 490 real-world long webpages and W-FFR with 507 goal-oriented tasks over 129 webpages.

    \item  We design evaluation protocols for long-range structural fidelity and end-to-end functional verification.
    
    \item We evaluate state-of-the-art VLMs under single-image and multi-image settings, revealing challenges in structural consistency and executable functional interaction.
\end{itemize}

%% file: 2-bench.tex
\section{LongWebBench}

We present the design of \textbf{LongWebBench}, a benchmark for evaluating long-horizon webpage generation from both structural and functional perspectives. We first define long webpages and the two evaluation tasks, then describe the construction of W-VFR and W-FFR, and finally analyze the dataset coverage and complexity.

\subsection{Task Definition}

\paragraph{Long Webpage.}
We define a long webpage as a webpage whose vertical length exceeds three viewport heights under a fixed $1920 \times 1080$ browser resolution, requiring models to aggregate visual information across multiple screens.

\paragraph{Webpage Visual Fidelity Replication (W-VFR).} Given a target long webpage screenshot, W-VFR evaluates whether a model can preserve the target visual structure, including page scale, global layout, section hierarchy, visual styling, and information density. Functional correctness is not considered, and interactive elements are treated as static visual components.

\paragraph{Webpage Functional Fidelity Realization (W-FFR).} Given a target webpage screenshot and a set of user-goal tasks, W-FFR evaluates whether a model can generate an executable webpage that supports the specified user goals. The screenshot provides component and state context, while evaluation focuses on whether prescribed actions lead to expected outcomes under browser execution. Visual similarity is not required.

\subsection{Data Construction}

\vpara{W-VFR: Long Webpage Collection.} We construct W-VFR from real-world webpages with diverse long-form structures. We define seven structural categories: informational articles, reference and documentation, product and service pages, community and discussion pages, multimedia and interactive pages, governmental and institutional pages, and data and resource aggregation pages. For each category, annotators collect candidate URLs from representative public websites to cover diverse layout patterns, content densities, and section organizations.

We then apply a multi-stage curation pipeline to obtain high-quality long-webpage snapshots. Candidate pages are rendered with Playwright under a fixed $1920 \times 1080$ browser resolution, and pages are filtered out if they fail to render, do not satisfy the length requirement, require authentication or personalization, contain excessive overlays or sensitive personal content, or involve unbounded infinite scrolling. We remove near-duplicates using perceptual hashing over rendered screenshots and capture each remaining page as a stabilized rendered snapshot, isolating long-page generation from external backend services and uncontrolled personalization. All screenshots are manually verified to ensure correct rendering and absence of capture artifacts. After filtering and verification, W-VFR contains 490 long webpages, with 70 webpages in each category. More details can be found in Appendix~\ref{appendix:w-vfr_cat}

\begin{figure}[t!]
    \vspace{-2mm}
    \centering
    \includegraphics[width=\linewidth]{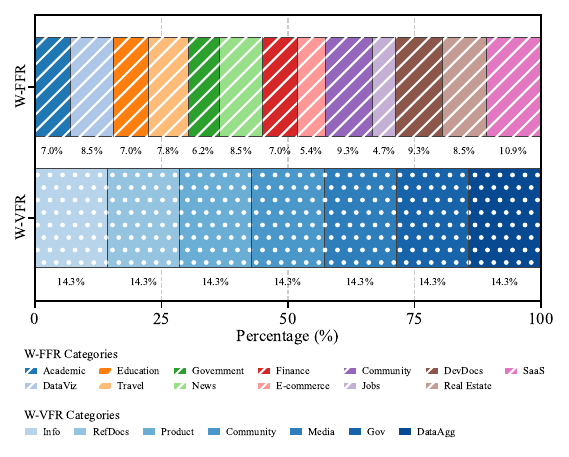}
    \caption{Category distribution of LongWebBench. W-VFR is balanced across 7 structural webpage categories, while W-FFR covers 13 user-goal-oriented functional categories.}
    \label{fig:category_distribution}
    \vspace{-6mm}
\end{figure}

\paragraph{W-FFR: Functional Task Construction.} While W-VFR is organized around visual structure, W-FFR is organized around user goals. We first define 13 functional categories based on the primary services provided by modern webpages, including academic search, data visualization, education, travel, government services, news, finance, e-commerce, community, jobs, developer documentation, real estate, and SaaS. Candidate webpages are collected and curated using the same rendering and quality-control pipeline as W-VFR, ensuring that each selected page provides meaningful frontend-observable interactions.

For each selected webpage, we construct 3--4 user-centered tasks that reflect realistic interaction goals on that page. We follow a draft--review--rewrite--verify workflow: task drafts are first created following predefined task-design guidelines, and annotators then rewrite, validate, and finalize them based on the rendered webpage. During verification, annotators check whether each task is clear, feasible, non-redundant, executable in the rendered webpage, and grounded in frontend-observable state changes rather than external backend services. Tasks that are ambiguous, backend-dependent, redundant, or infeasible are removed or revised. This process yields 129 webpages paired with 507 functional tasks, where each task serves as an independent evaluation instance. More details can be found in Appendix~\ref{appendix:w-ffr_cat}

\subsection{Data Analysis}

\vpara{Category Coverage.} LongWebBench covers diverse webpage structures and user goals. For W-VFR, the dataset is balanced across seven structural webpage categories, with 70 webpages per category. For W-FFR, webpages are organized into 13 user-goal-oriented functional categories, covering a broad range of real-world service objectives. As shown in Figure~\ref{fig:category_distribution}, LongWebBench provides broad coverage across both visual structures and functional scenarios.

\begin{figure}[t!]
    \vspace{-1mm}
    \centering
    \begin{subfigure}[t]{0.23\textwidth}
        \centering
        \includegraphics[width=\linewidth]{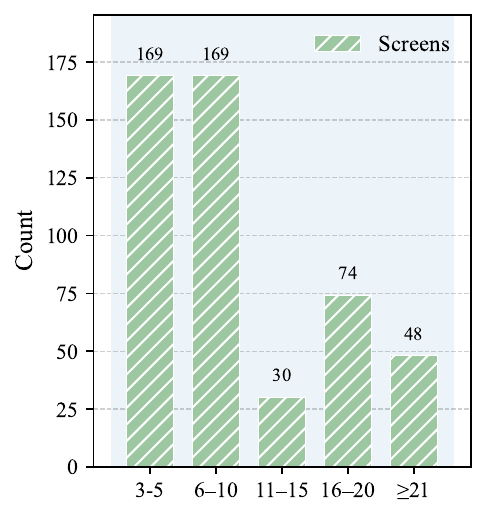}
        \caption{Distribution of vertical viewport counts in W-VFR.}
        \label{fig:screen_count_wvfr}
    \end{subfigure}
    \hfill
    \begin{subfigure}[t]{0.23\textwidth}
        \centering
        \includegraphics[width=\linewidth]{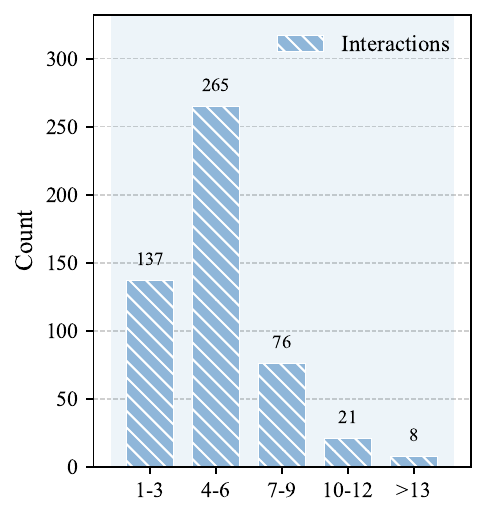}
        \caption{Distribution of interaction steps in W-FFR.}
        \label{fig:screen_count_wffr}
    \end{subfigure}
    \caption{Analysis of dataset complexity.}
    \label{fig:longweb_screencount}
    \vspace{-3mm}
\end{figure}

\vpara{Long-Webpage Complexity.} To characterize the long-horizon nature of W-VFR, we measure the number of vertical viewports required to cover each webpage under the fixed $1920 \times 1080$ resolution. As shown in Figure~\ref{fig:screen_count_wvfr}, webpages exhibit a long-tailed distribution. Most webpages span 3--10 viewports, while pages exceeding 10 viewports account for about 31\% of the dataset, with some extending up to 30 viewports. This distribution supports evaluation across different degrees of long-range visual structure and layout complexity.

\vpara{Interaction Complexity.} For W-FFR, we report the number of explicit user actions required to complete each task. As shown in Figure~\ref{fig:screen_count_wffr}, most tasks require medium-to-long interaction sequences: 52.27\% require 4--6 steps, 20.71\% require at least 7 steps, and 27.02\% require 1--3 steps. This distribution reflects the benchmark's emphasis on coordinated, multi-step interactions rather than isolated UI actions.

\subsection{Benchmark Design Principles}
LongWebBench is designed to disentangle two capabilities that are often conflated in webpage-to-code evaluation: long-range visual-structural reconstruction and executable interaction realization. W-VFR evaluates whether models can maintain coherent structure across extended page contexts, including repeated sections, cross-section consistency, and long-range layout organization, among others. W-FFR evaluates whether generated webpages can support user-goal completion under browser execution, rather than merely rendering visually plausible interactive elements. 
We evaluate stabilized rendered snapshots and frontend-observable interactions, avoiding dependence on external backends while preserving the long-page structure and multi-step interaction patterns needed for reproducible evaluation.

%% file: 3-evaluation-pipeline.tex

\section{Evaluation Protocols}

\begin{figure*}
    \centering
    \includegraphics[width=0.98\linewidth]{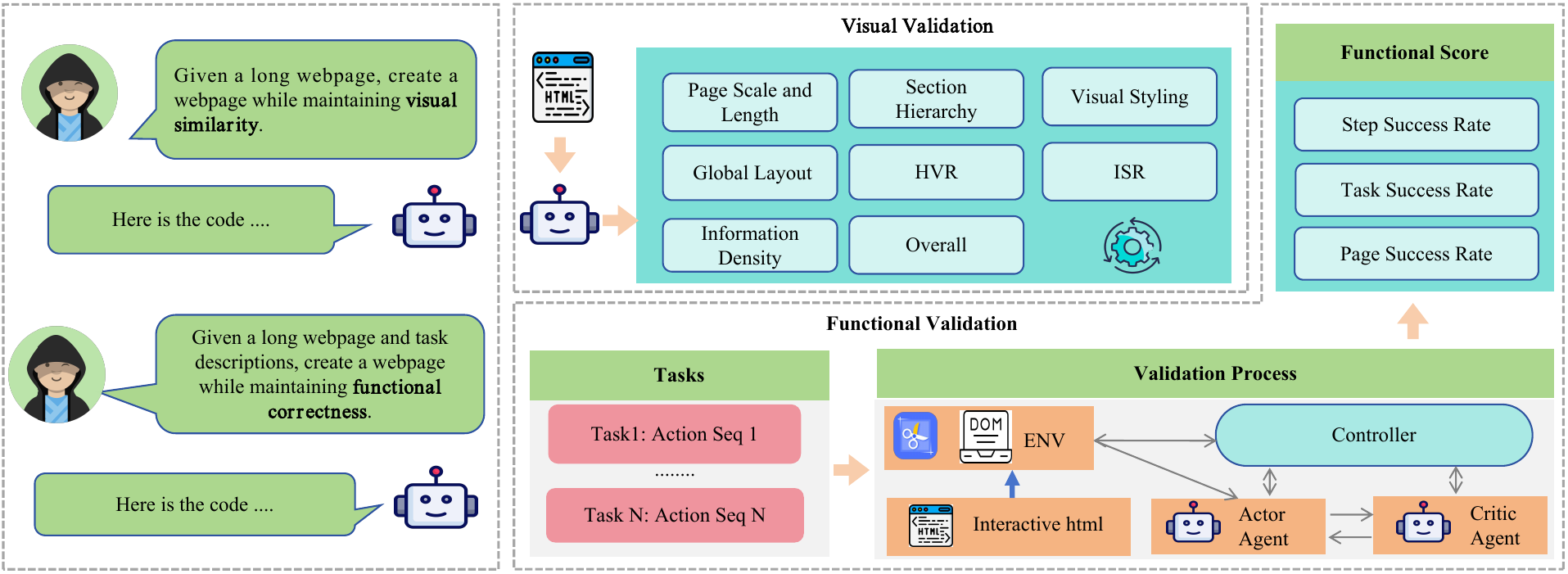}
    \vspace{-2mm}
    \caption{Overview of LongWebBench evaluation. W-VFR evaluates structural fidelity across five dimensions: page scale, global layout, section hierarchy, visual styling, and information density. W-FFR evaluates functional fidelity by executing goal-oriented interaction tasks in a browser environment and reporting step-, task-, and page-level success rates.}
    \label{fig:evaluation_pipeline}
    \vspace{-3mm}
\end{figure*}

We design two evaluation protocols for LongWebBench, as illustrated in Figure~\ref{fig:evaluation_pipeline}. W-VFR evaluates long-range structural fidelity by comparing generated renderings with reference long webpages, while W-FFR verifies whether generated webpages support executable user goals under browser execution.

\subsection{Evaluation Settings}

LongWebBench supports two input settings. In the \textit{single-image} setting, the full webpage screenshot is provided as one long image, testing whether models can process long visual contexts under their native image-input constraints. In the \textit{multi-image} setting, the screenshot is split into consecutive viewport-sized tiles and provided in top-to-bottom order, reducing resolution constraints while requiring cross-image integration. These settings help separate input-size limitations from long-context structural reasoning challenges.

\subsection{Structural Fidelity Validation}

To evaluate W-VFR, we use a VLM-based relative evaluator that compares a generated full-page rendering with the reference long webpage screenshot. The evaluator scores the generated page based only on visually observable evidence, explicitly excluding interactive behavior. We use \textit{GPT-4o} as the primary structural evaluator with deterministic decoding, and provide the full prompt, model version, and scoring rubric in Appendix~\ref{appendix:details_wvfr}.

The evaluator assigns scores on a 0--10 scale across five dimensions: \textit{page scale and scroll consistency}, \textit{global layout consistency}, \textit{section-level structure and hierarchy}, \textit{local visual styling consistency}, and \textit{information density}. The overall W-VFR score is the average of the five dimension scores. To better capture low-level appearance consistency, the local visual styling dimension combines VLM-based scoring with DINO-based feature similarity.

We also report three availability metrics to separate input-side constraints from generation quality. \textit{Input Support Rate} (ISR) measures the fraction of test cases accepted by the model or API before generation, \textit{HTML Validity Rate} (HVR) measures, among supported cases, the proportion of outputs that contain a valid renderable HTML artifact, and \textit{Code Quality} (CQ) measures the overall quality of the generated code. Detailed definitions and scoring procedures are provided in Appendix~\ref{appendix:metrics}.

\subsection{Functional Fidelity Validation}

To evaluate W-FFR, we use a task-driven execution-based protocol that verifies whether a generated webpage supports specified user goals under browser execution. Unlike W-VFR, functional fidelity does not evaluate visual similarity to the reference screenshot. The screenshot and task descriptions provide component and state context, while evaluation focuses on whether prescribed interactions can be executed and lead to the expected outcomes.

Each task is evaluated independently in a controlled browser environment. A controller coordinates an actor agent, which executes the prescribed operations, and a critic agent, which checks whether the resulting DOM and rendered page state satisfy the task specification; both agents are powered by \textit{Gemini-3-Pro}. A task is successful only if all required interaction steps are executable and collectively lead to the expected outcome. Implementation details and prompts are provided in Appendix~\ref{appendix:details_wffr}.

We report three functional metrics. \textit{Step Success Rate} (SSR) measures the fraction of interaction steps that are executable and trigger the expected intermediate state change. \textit{Task Success Rate} (TSR) measures the fraction of tasks completed successfully and serves as the primary W-FFR metric. \textit{Page Success Rate} (PSR) measures the fraction of webpages for which all associated tasks are completed successfully, providing a stricter estimate of page-level functional robustness.

%% file: 4-experiments.tex
\section{Experiments}

\begin{table*}[t]
\centering
\caption{Performance comparison of VLMs on LongWebBench under single-image and
multi-image input settings. ISR, HVR, and CQ are reported as percentages (\%);
category-wise and overall scores are on a 0--10 scale (macro-average).}
\resizebox{\textwidth}{!}{%
\setlength{\tabcolsep}{1mm}
\begin{tabular}{llcc|cccccccc|c}
\toprule
& & \multicolumn{2}{c|}{\textbf{Availability (\%)}}
& \multicolumn{9}{c}{\textbf{Average Score (0--10)}} \\
\cmidrule(lr){3-4}\cmidrule(lr){5-13}
\textbf{Type} & \textbf{Model}
& \textbf{ISR} & \textbf{HVR}
& \textbf{Comm.} & \textbf{Data} & \textbf{Gov.} & \textbf{Info.}
& \textbf{Prod.} & \textbf{Ref.} & \textbf{Trans.}
& \textbf{Overall} & \textbf{CQ} \\

\midrule
\multicolumn{13}{c}{\textbf{Single-Image Input}} \\
\midrule
\textit{Open-source}
& Kimi-VL-A3B-Thinking        & 100.00 & 78.98 & 2.69 & 2.64 & 3.36 & 2.20 & 3.02 & 2.20 & 3.95 & 2.87 & 92.50 \\
& Qwen3-VL-8B-Instruct        & 95.71  & 54.37 & 3.72 & 3.64 & 3.92 & 3.17 & 3.56 & 2.02 & 4.28 & 3.47 & 88.38 \\
& InternVL3-78B               & 100.00 & 81.84 & 3.48 & 3.43 & 3.85 & 3.42 & 3.35 & 2.78 & 3.50 & 3.40 & 94.41 \\
& GLM-4.1V-9B-Thinking-Flash  & 100.00 & 84.90 & 5.32 & 4.27 & 5.15 & 5.32 & 5.18 & 3.34 & 5.54 & 4.87 & 92.95 \\
& GLM-4.6V                    & 100.00 & 84.29 & 5.85 & 5.55 & 6.56 & 5.64 & 5.62 & 3.93 & 6.44 & 5.66 & 94.06 \\
& Qwen3-VL-235B-A22B-Instruct & 99.80  & 80.78 & 5.29 & 5.19 & 6.23 & 5.72 & 4.44 & 3.40 & 5.45 & 5.10 & 89.20 \\
\cmidrule(l){2-13}
\textit{Closed-source}
& Claude-Sonnet-4-5-Thinking  & 56.12  & 99.64 & 5.97 & 3.49 & 4.74 & 2.58 & 3.32 & 0.29 & 6.27 & 3.81 & 93.75 \\
& Claude-Opus-4-5-20251101    & 56.12  & 99.64 & 5.95 & 3.34 & 4.70 & 2.49 & 3.40 & 0.34 & 6.67 & 3.84 & 93.14 \\
& GPT-5.2                     & 95.71  & 100.00& 7.07 & 6.53 & 6.57 & 6.52 & 6.07 & 5.52 & \textbf{7.20} & 6.50 & 94.14 \\
& GPT-4o                      & 100.00 & 99.39 & 3.15 & 3.12 & 3.62 & 3.05 & 3.21 & 2.60 & 3.84 & 3.23 & \textbf{95.59} \\
& Gemini-3-Pro-preview        & 100.00 & 100.00& \textbf{7.39} &\textbf{ 6.89} & \textbf{6.88} & \textbf{6.88} & \textbf{6.66} & \textbf{5.83} & 6.89 &\textbf{ 6.77 }& 93.00 \\
& Gemini-3-Flash-preview      & 100.00 & 100.00& 6.97 & 5.95 & 6.84 & 6.77 & 6.32 & 4.87 & 6.78 & 6.36 & 93.31 \\
& Doubao-Seed-1-6             & 76.33  & 98.93 & 4.35 & 2.89 & 4.25 & 4.51 & 4.13 & 0.45 & 5.05 & 3.66 & 94.58 \\

\midrule
\multicolumn{13}{c}{\textbf{Multi-Image Input}} \\
\midrule
\textit{Open-source}
& Kimi-VL-A3B-Thinking        & 100.00 & 80.20 & 3.00 & 3.05 & 3.52 & 2.70 & 3.33 & 2.56 & 3.63 & 3.12 & 89.88 \\
& Qwen3-VL-8B-Instruct        & 100.00 & 54.69 & 4.14 & 3.81 & 4.94 & 3.31 & 4.16 & 3.23 & 4.02 & 3.95 & 86.72 \\
& InternVL3-78B               & 99.39  & 77.21 & 3.53 & 3.73 & 3.51 & 3.51 & 3.65 & 3.67 & 3.72 & 3.62 & \textbf{94.81} \\
& GLM-4.1V-9B-Thinking-Flash  & 100.00 & 83.06 & 1.46 & 2.71 & 1.12 & 0.39 & 0.31 & 0.31 & 0.28 & 0.94 & 92.59 \\
& GLM-4.6V                    & 100.00 & 88.16 & 5.76 & 6.16 & 6.50 & 5.62 & 5.65 & 5.79 & 6.25 & 5.96 & 93.54 \\
& Qwen3-VL-235B-A22B-Instruct & 99.80  & 80.57 & 6.08 & 5.68 & 6.38 & 5.98 & 5.00 & 6.03 & 5.83 & 5.85 & 87.43 \\
\cmidrule(l){2-13}
\textit{Closed-source}
& Claude-Sonnet-4-5-Thinking  & 100.00 & 99.39 & 6.73 & 6.61 & 7.05 & 6.97 & \textbf{7.07} & 6.04 & 6.69 & 6.74 & 93.79 \\
& Claude-Opus-4-5-20251101    & 100.00 & 97.96 & 6.83 & 6.28 & 6.51 & 6.13 & 6.51 & 6.25 & 6.49 & 6.43 & 93.07 \\
& GPT-5.2                     & 100.00 & 99.59 & 6.96 & 6.24 & 5.97 & 5.85 & 5.67 & 5.92 & 6.36 & 6.14 & 93.95 \\
& GPT-4o                      & 100.00 & 99.80 & 3.39 & 3.40 & 3.91 & 3.50 & 3.44 & 3.13 & 3.92 & 3.53 & 94.51 \\
    & Gemini-3-Pro-preview        & 100.00 & 99.59 & \textbf{7.51} & \textbf{7.07} & \textbf{7.43} & \textbf{7.08} & 6.99 & \textbf{7.35} & \textbf{7.08} & \textbf{7.22} & 92.25 \\
& Gemini-3-Flash-preview      & 100.00 & 100.00& 6.83 & 6.46 & 6.88 & 6.81 & 6.57 & 5.48 & 6.70 & 6.53 & 92.25 \\
& Doubao-Seed-1-6             & 100.00 & 99.39 & 4.50 & 4.57 & 5.01 & 4.96 & 4.80 & 4.39 & 5.14 & 4.77 & 94.59 \\
\bottomrule
\end{tabular}}
\vspace{-3mm}
\label{tab:results_combined}
\end{table*}

\subsection{Experimental Setup}

\paragraph{Evaluated Models.}
We evaluate a broad set of state-of-the-art VLMs, covering both proprietary models (e.g., GPT-5.2 \cite{singh2025openai}, GPT-4o \cite{openai2024gpt4technicalreport}, Claude-Opus-4.5, Claude Sonnet-4-5-Thinking, Gemini-3-Pro \cite{comanici2025gemini}, Gemini-3-Flash, and Doubao-Seed-1-6) and open-source models (e.g., GLM-4.6V , GLM-4.1V-9B-Thinking-Flash \cite{vteam2025glm45vglm41vthinkingversatilemultimodal}, Qwen3-VL-235B-A22B-Instruct \cite{yang2025qwen3},
Qwen3-VL-8B-Instruct, InternVL3-78B \cite{wang2025internvl3}, and Kimi-VL-A3B-Thinking \cite{team2025kimi}). All models are evaluated with the same generation prompts.

\paragraph{Evaluation Settings.}
We evaluate models under both single-image and multi-image settings. W-VFR is evaluated on the full model set when the input format is supported. W-FFR is evaluated on a representative subset due to the higher cost of browser-based execution, covering both proprietary and open-source models with different input capabilities and performance levels. Additional implementation details are provided in Appendix~\ref{appendix:model_settings}.

\subsection{Structural Fidelity Results}

Table~\ref{tab:results_combined} reports W-VFR results under both single-image and multi-image input settings. Overall, current VLMs still achieve only moderate structural fidelity on long webpages, showing that long-horizon webpage reconstruction remains challenging beyond local visual matching. Under the single-image setting, Gemini-3-Pro achieves the highest overall score, followed by GPT-5.2 and Gemini-3-Flash. Strong open-source models such as GLM-4.6V and Qwen3-VL-235B-A22B-Instruct are competitive, but still lag behind the best proprietary systems.
Multi-image input improves input availability and often improves structural fidelity by reducing image-size constraints, especially for Claude models. However, the gains remain limited: even under the multi-image setting, the best model still reaches only moderate fidelity. This suggests that multi-image input mainly alleviates input-resolution constraints, but long-webpage generation still requires robust cross-image integration and global layout planning.

\subsection{Functional Fidelity Results}

Table~\ref{tab:results_fun} reports W-FFR results under both input settings. Due to the high cost of browser execution and agent-based verification, we evaluate a representative subset covering strong closed-source and competitive open-source models from W-VFR. Functional fidelity is substantially more challenging than step-level execution alone: models with high Step Success Rate (SSR) still show much lower Task Success Rate (TSR) and Page Success Rate (PSR), indicating error accumulation across multi-step workflows and multiple tasks on the same page. Claude-Sonnet-4.5-Thinking and Gemini-3-Pro perform best, reaching SSR above 86\% and TSR around 58--59\% under the single-image setting, while their PSR remains below 22\%. Open-source models such as GLM-4.6V and Qwen3-VL-235B-A22B-Instruct achieve much lower page-level success, showing limited robustness in end-to-end functional realization. Multi-image input provides only limited and inconsistent gains, suggesting that executable interaction depends not only on input resolution, but also on correct interaction logic and state transitions.

\begin{table}[t!]
\centering
\caption{W-FFR results under single-image and multi-image input settings.
SSR, TSR, and PSR denote step, task, and page success rates, respectively.}
\small
\begin{tabular}{lrrr}
\toprule
\textbf{Model} & \textbf{SSR} & \textbf{TSR} & \textbf{PSR} \\
\midrule
\multicolumn{4}{c}{\textbf{Single-Image Input}} \\
\midrule
GLM-4.6V                    & 62.59 & 28.39 &  5.56 \\
Gemini-3-Pro-preview                & \textbf{87.30} & 58.38 & 17.83 \\
Qwen3-VL-235B-A22B-Instruct & 60.52 & 23.65 &  3.13 \\
Claude-Sonnet-4-5-Thinking  & 86.91 & \textbf{59.40} & \textbf{21.78} \\
\midrule
\multicolumn{4}{c}{\textbf{Multi-Image Input}} \\
\midrule
GLM-4.6V                    & 64.20 & 30.52 &  7.03 \\
Gemini-3-Pro-preview                & \textbf{87.91 }& \textbf{59.24}& \textbf{21.09 }\\
Qwen3-VL-235B-A22B-Instruct & 60.34 & 23.27 &  6.20 \\
Claude-Sonnet-4-5-Thinking  & 86.46 & 55.47 & 17.19 \\
\bottomrule
\end{tabular}
\vspace{-3mm}
\label{tab:results_fun}
\end{table}

\subsection{Effect of Length on Structural and Functional Fidelity}

\begin{figure}[h]
    \centering
    \includegraphics[width=0.9\linewidth]{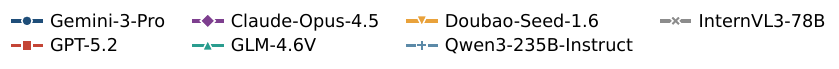}\\[-2pt]
    \begin{subfigure}[t]{0.49\linewidth}
        \centering
        \includegraphics[width=\linewidth]{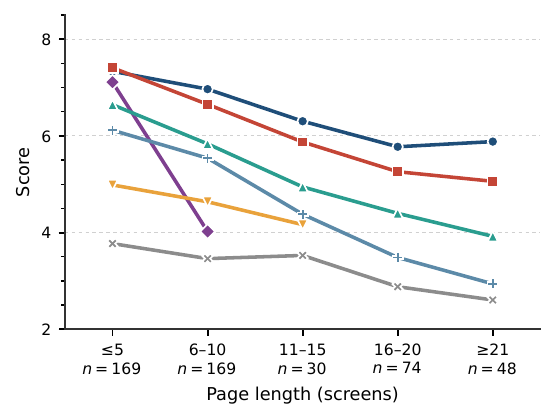}
        \caption{Single-image input.}
        \label{fig:screen_counts_single}
    \end{subfigure}
    \hfill
    \begin{subfigure}[t]{0.49\linewidth}
        \centering
        \includegraphics[width=\linewidth]{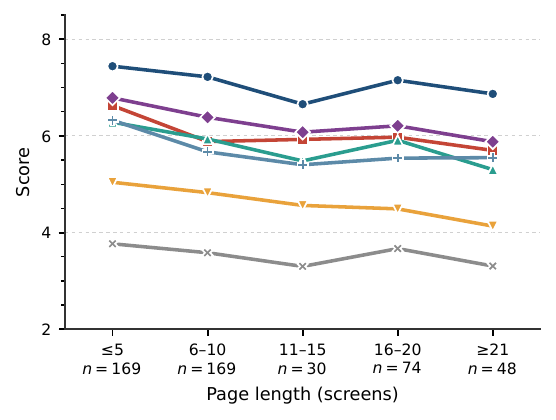}
        \caption{Multi-image input.}
        \label{fig:screen_counts_tiles}
    \end{subfigure}
    \vspace{-2mm}
    \caption{W-VFR performance across different webpage lengths.}
    \label{fig:screen_counts}
    \vspace{-3mm}
\end{figure}

\vpara{Webpage Length.} Figure~\ref{fig:screen_counts} shows W-VFR performance across webpages with different numbers of vertical viewports. Structural fidelity generally decreases as webpage length increases, indicating that long webpages require global layout planning, section ordering, and style consistency beyond local visual reconstruction. Multi-image input alleviates this degradation by preserving finer-grained local details, but performance still drops on very long webpages, suggesting that robust cross-segment integration remains challenging.

\begin{figure}[h]
    \centering
    \includegraphics[width=0.85\linewidth]{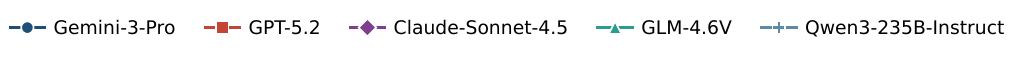}\\[-2pt]
    \begin{subfigure}[t]{0.48\linewidth}
        \centering
        \includegraphics[width=\linewidth]{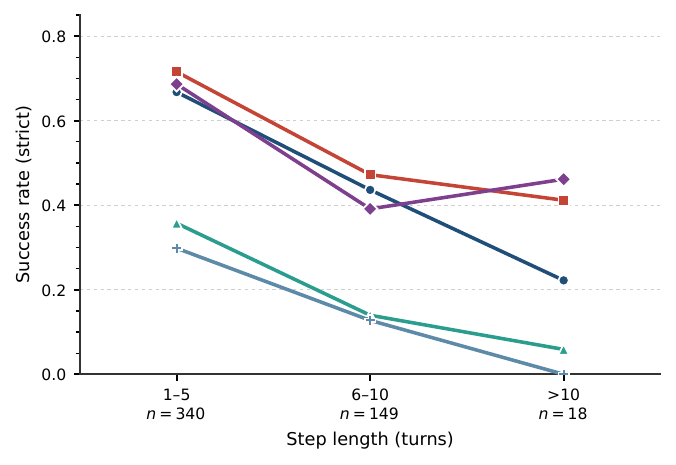}
        \caption{Single-image input.}
        \label{fig:succ_rate_single}
    \end{subfigure}
    \hfill
    \begin{subfigure}[t]{0.48\linewidth}
        \centering
        \includegraphics[width=\linewidth]{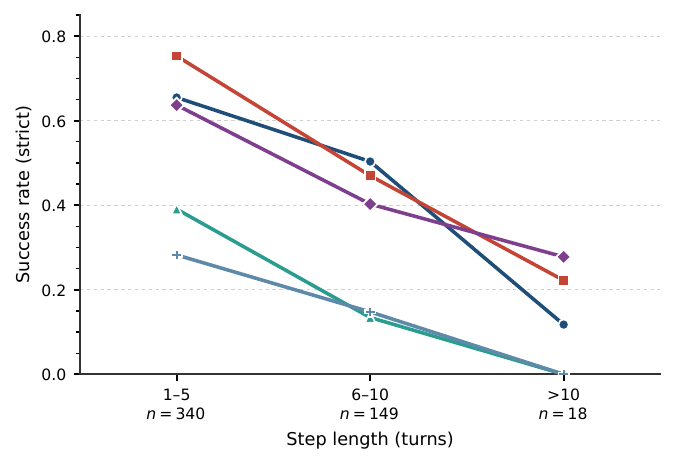}
        \caption{Multi-image input.}
        \label{fig:succ_rate_tiles}
    \end{subfigure}
    \caption{W-FFR success rate across different interaction lengths.}
    \label{fig:succ_rate_by_step}
    \vspace{-3mm}
\end{figure}

\vpara{Interaction Length.} Figure~\ref{fig:succ_rate_by_step} shows W-FFR success rates across tasks with different numbers of interaction steps. Success rates generally decrease as interaction length increases, indicating that longer workflows amplify error accumulation and require more stable state tracking. These results show that W-FFR evaluates more than isolated clickability. Generated webpages must maintain correct interaction logic and state transitions across multi-step user goals, where early mistakes can propagate and prevent task completion.

\subsection{Robustness of Structural Evaluation}

\begin{figure}[h]
    \centering
    \includegraphics[width=0.65\linewidth]{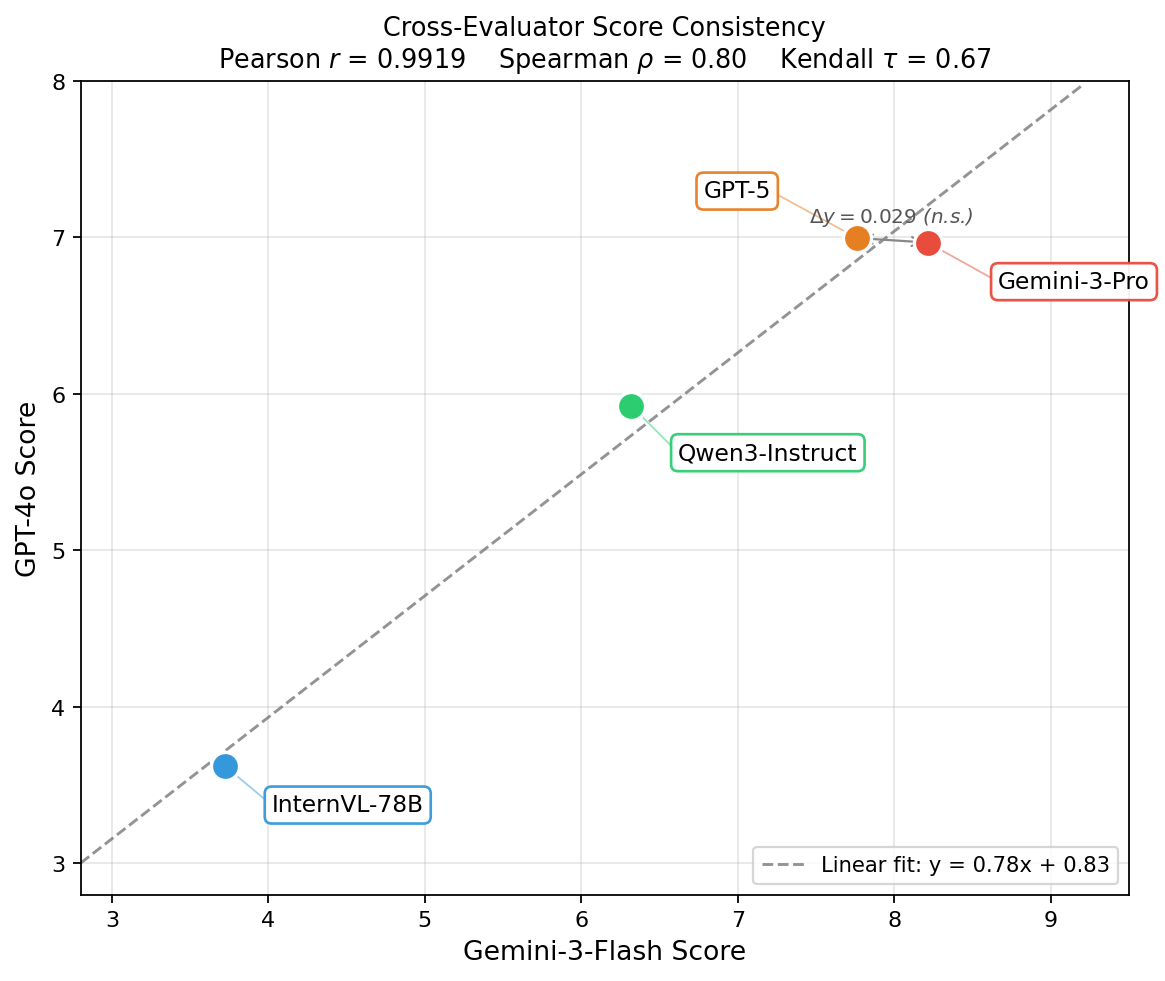}
    \caption{Per-model mean fidelity scores from two independent judge models. The near-collinear arrangement ($r=0.992$) indicates that the relative ranking is invariant to the choice of judge.}
    \label{fig:crossmodelevalution}
    \vspace{-3mm}
\end{figure}

To examine whether W-VFR scores depend on a single VLM evaluator, we conduct a cross-judge robustness analysis. We sample 100 test cases, each consisting of a reference screenshot and the corresponding generated rendering, from four representative models: GPT-5.2, Gemini-3-Pro, Qwen3-VL-235B-A22B-Instruct and InternVL3-78B. Using the same prompt and rubric, we score these samples with both GPT-4o and Gemini-3-Flash.
As shown in Figure~\ref{fig:crossmodelevalution}, the two judges produce highly consistent results. Their per-model mean scores are nearly collinear, with Pearson $r=0.992$. Pairwise ranking agreement is $5/6$ ($83.3\%$), with the only inversion occurring between GPT-5.2 and Gemini-3-Pro, whose scores differ by only $0.03$ points.
These results indicate that our structural evaluation is robust to the choice of judge model.

\subsection{Agreement with Human Evaluation}

\begin{figure}[t]
    \centering
    \begin{subfigure}[t]{0.45\linewidth}
        \centering
        \includegraphics[width=\linewidth]{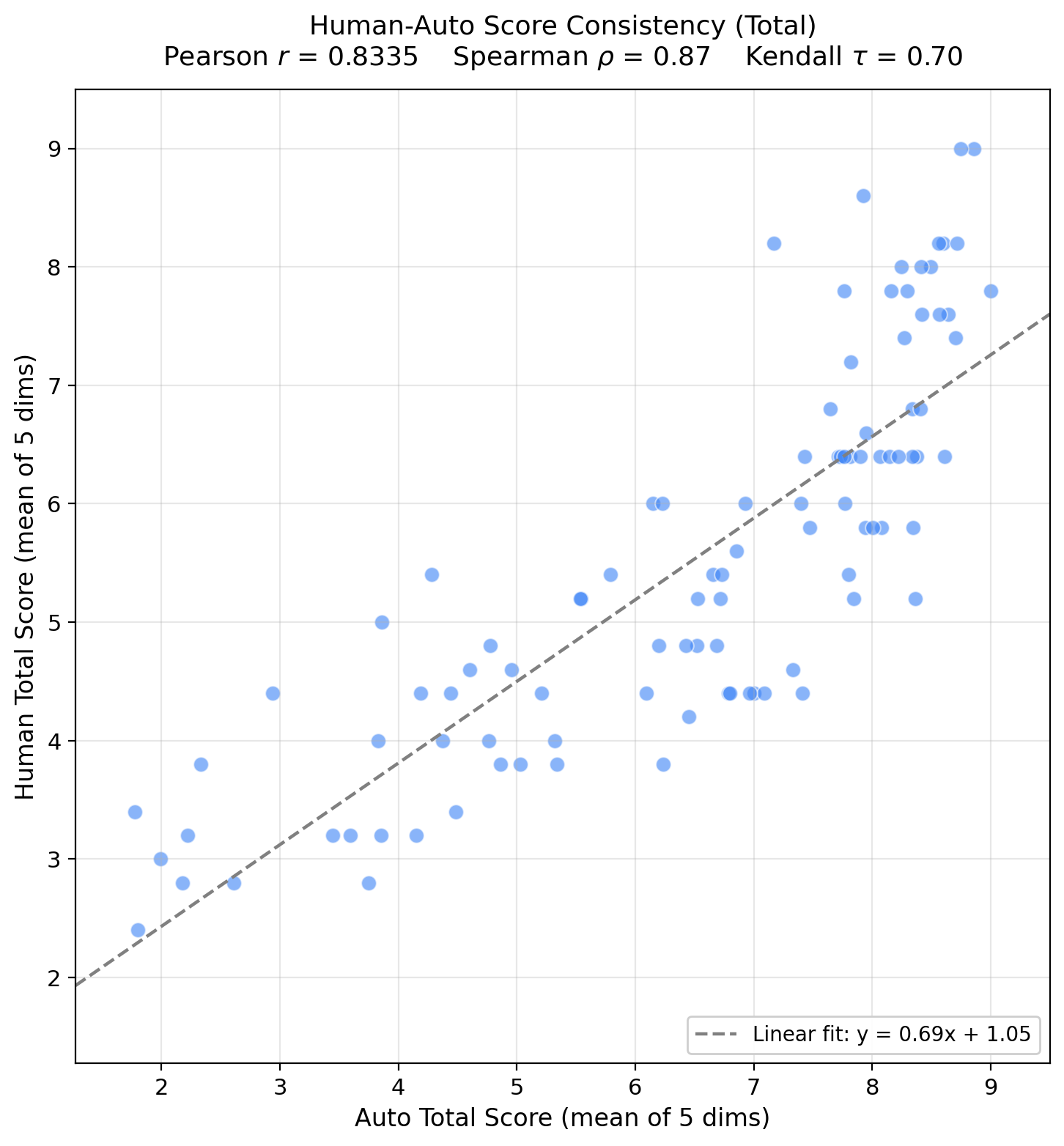}
        \caption{Overall-score correlation.}
        \label{fig:human-align-total}
    \end{subfigure}
    \hfill
    \begin{subfigure}[t]{0.48\linewidth}
        \centering
        \includegraphics[width=\linewidth]{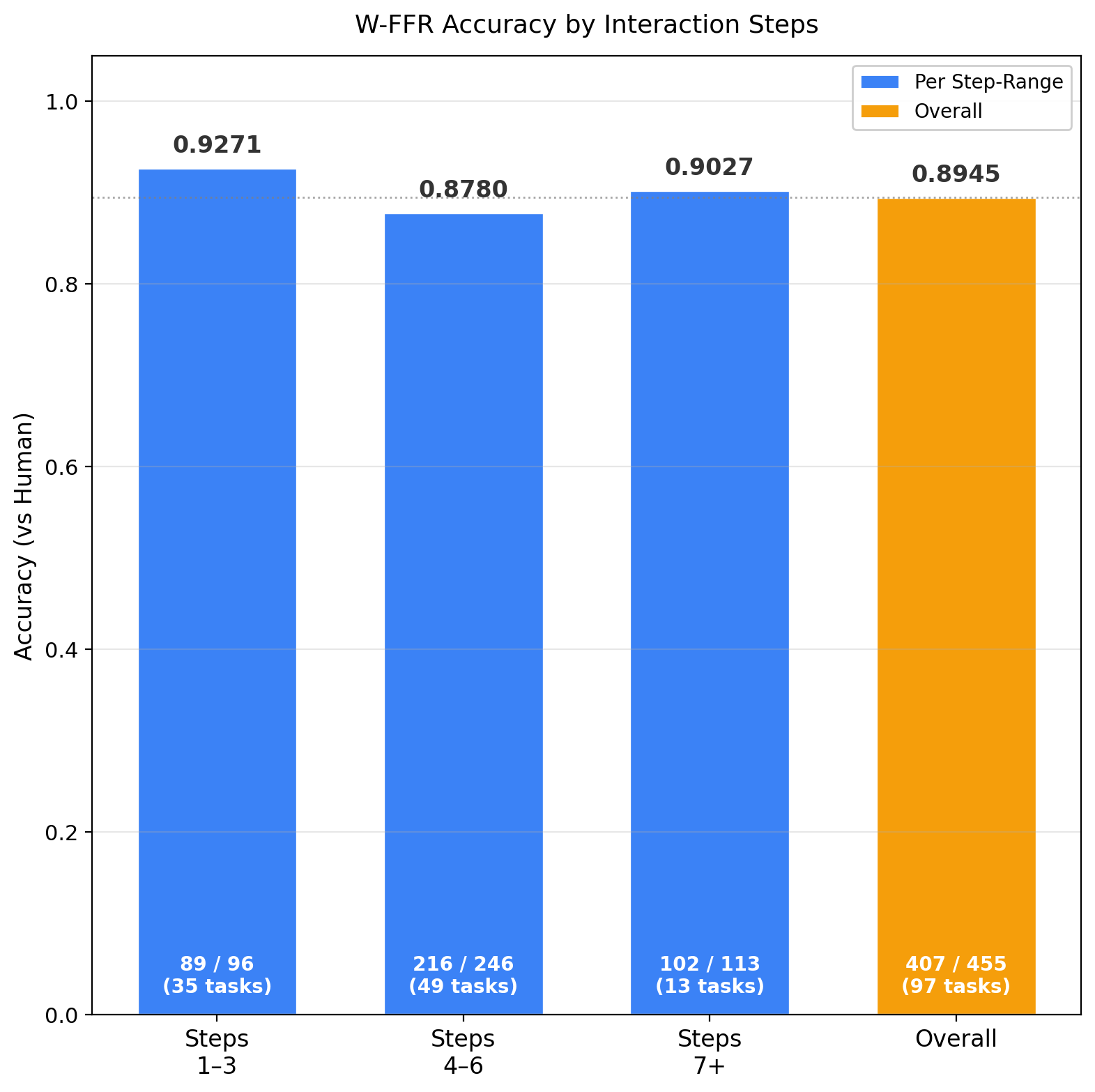}
        \caption{Per step-range and overall W-FFR accuracy.}
        \label{fig:human-align-dim}
    \end{subfigure}
    \caption{Agreement between automatic evaluation and human judgments. (a) W-VFR score-level agreement on $100$ sampled outputs. (b) W-FFR task-level consistency across interaction-step lengths.}
    \label{fig:human-alignment}
    \vspace{-3mm}
\end{figure}

To assess the reliability of LongWebBench's automatic evaluation protocols, we compare them with human judgments on sampled model outputs. For W-VFR, three annotators independently rate each generated webpage against the reference screenshot on the same 0--10 structural fidelity scale, and their average rating is used as the human score. For W-FFR, annotators manually inspect browser execution traces and assign binary success labels for executed interaction steps.
As shown in Figure~\ref{fig:human-alignment}, the automatic protocols align well with human judgments. On W-VFR, automatic structural scores achieve strong agreement with human ratings over $100$ sampled outputs, with Pearson $r=0.8335$ and Spearman $\rho=0.87$. On W-FFR, the automatic verifier reaches an overall consistency of $0.8945$ over $455$ evaluated steps from $97$ tasks, and remains above $0.87$ across different interaction-length groups. These results support the use of the automatic protocols for scalable evaluation of both long-range structural fidelity and executable functional correctness. Details of annotation and agreement analysis are provided in Appendix~\ref{appendix:human_evaluation}.

%% file: 5-related-works.tex
\section{Related Works}


\subsection{Visual-to-Code Generation}

Visual-to-code generation aims to convert visual interface inputs into executable code. Early work such as pix2code~\cite{beltramelli2018pix2code} studied GUI screenshot-to-code, while recent VLM-based systems extend this paradigm to webpage and front-end code generation~\cite{wang2024cogvlm,wang2025internvl3,jiang2025screencoder}. Since HTML/CSS implementations are non-unique and original source code is often unavailable, existing benchmarks commonly adopt render-based evaluation, including Web2Code~\cite{yun2024web2code}, Design2Code~\cite{si2025design2code}, Sketch2Code~\cite{li2025sketch2code}, and WebCode2M~\cite{gui2025webcode2m}. However, these benchmarks mainly focus on short, single-screen, or mostly static pages. Long webpages introduce additional challenges, including global layout consistency, section ordering, repeated modules, information density, and style coherence across multiple viewports. LongWebBench complements existing benchmarks by treating long-range structural fidelity as a central evaluation target and supporting both single-image and multi-image input settings.

\subsection{Functional Webpage Generation}

Realistic webpage generation requires executable interaction behavior beyond visual reconstruction. Recent benchmarks begin to evaluate interaction-aware generation: Interaction2Code~\cite{xiao2024interaction2code} focuses on isolated interactions such as clicks, hovers, and toggles, FullFront~\cite{sun2025fullfront} uses DOM-level signals to assess front-end workflow components, and WebGen-Bench~\cite{lu2025webgen} studies functional website generation beyond static appearance. Related web-agent benchmarks such as WebArena~\cite{zhou2023webarena}, Mind2Web~\cite{deng2023mind2web}, BrowserGym~\cite{chezelles2024browsergym}, and WebVIA~\cite{xu2025webvia} evaluate agents operating on existing webpages or browser environments. In contrast, LongWebBench evaluates whether a generated webpage itself contains the interaction logic and state transitions needed to support executable multi-step user goals.

%% file: 6-conclusion.tex
\section{Conclusion}

In this paper, we introduced \textbf{LongWebBench}, a benchmark for evaluating long-horizon webpage generation from structural and functional perspectives. Unlike short, static visual reconstruction benchmarks, LongWebBench tests whether generated webpages preserve global structure across multiple viewports and support executable multi-step user interactions. Experiments with state-of-the-art VLMs show that structural fidelity degrades as webpage length increases, while visually plausible generations often fail to complete functional user goals. These findings highlight the need to evaluate webpage generation beyond static visual similarity and provide a diagnostic testbed for long-context multimodal code generation.

%% file: 8-appendix.tex
\clearpage
\onecolumn       
\section*{Appendix}
\startcontents[sections]
\printcontents[sections]{l}{1}{\setcounter{tocdepth}{2}}
\twocolumn                          
\newpage

\label{sec:appendix}

\section{Data Annotation}
\begin{table*}[]
\centering
\caption{The descriptions of W-VFR dataset categories.}
\renewcommand{\arraystretch}{1.3}
\begin{tabular}{>{\centering\arraybackslash}p{1.2cm}
                >{\raggedright\arraybackslash}p{3.2cm}
                >{\raggedright\arraybackslash}p{3.4cm}
                >{\raggedright\arraybackslash}p{6.1cm}}
\toprule
\# & Category & Content Driver & Comment \\
\midrule
1 & Informational Articles     & Narrative-driven              & Users primarily gain information by reading a continuous textual narrative. \\
2 & Reference \& Documentation & Lookup-driven                 & Users search with a clear goal to find and locate specific entries or instructions. \\
3 & Product \& Service Pages   & Offering-driven               & The page aims to present and drive the selection of a specific product or service. \\
4 & Community \& Discussion    & UGC-driven                    & The main information comes from user-generated content and interaction structures. \\
5 & Governmental / Institutional & Authority-driven            & The page centers on information published by official or authoritative institutions. \\
6 & Data \& Resource Aggregation & Listing-driven              & The core is a systematic listing of large amounts of data, resources, or items for retrieval. \\
7 & Transportation \& Travel Services & Decision + Transaction-driven & Users search, compare, and book travel options under multiple constraints. \\
\bottomrule
\end{tabular}

\label{tab:W-VFR}
\end{table*}

\subsection{More details about W-VFR categories}
\textbf{W-VFR Categories}
\label{appendix:w-vfr_cat}
Table \ref{tab:W-VFR} provides a taxonomy of page types. We define categories using two complementary criteria: (1) the page’s primary functional role in the information ecosystem (e.g., explaining, lookup, offering/transaction, or interaction), and (2) the dominant mechanism by which users obtain content on the page (“content driver”), such as narrative reading, goal-directed lookup, offering-driven choice, or UGC-mediated discussion. This scheme has three main advantages: it is explicit and operational, enabling consistent labeling across domains and websites; it aligns naturally with user intent and page structure, making it easy to map categories to observable features (e.g., navigation depth, modular layout, interaction density, and conversion components) for downstream detection, evaluation, and stratified experimentation; and it is extensible—new page forms can be incorporated by adding categories or drivers without undermining comparability with the existing taxonomy.

\subsection{More details about W-FFR categories}
\textbf{W-FFR Categories}
\label{appendix:w-ffr_cat}
Table \ref{tab:w-vfr-categories} summarizes the category taxonomy of the W-VFR dataset, covering a diverse range of real-world web scenarios. Each category is characterized by its dominant content driver, reflecting the primary user intent and interaction pattern, such as information retrieval, decision making, learning, or user-generated interaction. This categorization highlights the heterogeneity of web pages in terms of content structure and functional objectives, providing a structured foundation for analyzing model performance across different webpage types.

\begin{table*}[htbp]
\centering
\caption{The descriptions of W-VFR dataset categories.}
\renewcommand{\arraystretch}{1.3}
\begin{tabular}{>{\centering\arraybackslash}p{0.5cm}
                >{\raggedright\arraybackslash}p{3.8cm}
                >{\raggedright\arraybackslash}p{3.0cm}
                >{\raggedright\arraybackslash}p{8.0cm}}
\toprule
\# & Category & Content Driver & Comment \\
\midrule
1  & Academic Search                  & Retrieval-driven    & Users search for academic papers, authors, or citations with precise information needs, emphasizing accurate retrieval and relevance ranking. \\
2  & Community \& Discussion          & UGC-driven          & The primary content is generated by users through questions, answers, and discussions, where information emerges from collective interaction. \\
3  & Data Visualization \& Analytics  & Insight-driven      & Pages focus on presenting data through visual or analytical forms to support interpretation, comparison, and exploratory analysis. \\
4  & Developer Docs \& Cloud Console  & Lookup-driven       & Users seek technical documentation, APIs, or configuration instructions to support development and system operation tasks. \\
5  & E-commerce Product Selection     & Decision-driven     & The page assists users in comparing products, prices, and attributes to facilitate informed purchasing decisions. \\
6  & Education \& Course              & Learning-driven     & Content is structured to support systematic learning, including tutorials, courses, and educational materials. \\
7  & Finance \& Market Analysis       & Analysis-driven     & Pages provide financial data, trends, and analytical insights to support investment, risk assessment, and market understanding. \\
8  & Government \& Open Data          & Authority-driven    & Information is published by official institutions, focusing on policies, regulations, and publicly released datasets. \\
9  & Jobs \& Recruitment              & Matching-driven     & Users search and compare job opportunities or candidates based on structured requirements and qualifications. \\
10 & News \& Interactive Media        & Narrative-driven    & Content delivers timely information through news reporting or interactive storytelling, emphasizing readability and engagement. \\
11 & Real Estate Search               & Decision-driven     & Pages support searching, filtering, and comparing properties under multiple constraints such as price, location, and amenities. \\
12 & SaaS Pricing                     & Comparison-driven   & The focus is on pricing structures, feature tiers, and cost comparison to support service selection decisions. \\
13 & Travel \& Transportation         & Planning-driven     & Pages support users in searching, comparing, and booking travel options such as flights, hotels, and transit routes, emphasizing multi-constraint filtering and itinerary planning. \\
\bottomrule
\end{tabular}
\label{tab:w-vfr-categories}
\end{table*}

\subsection{Distribution of webpage lengths across categories}
\label{appendix:screenshots_lenghts_distribution}
\begin{figure*}
    \centering
    \includegraphics[width=0.98\linewidth]{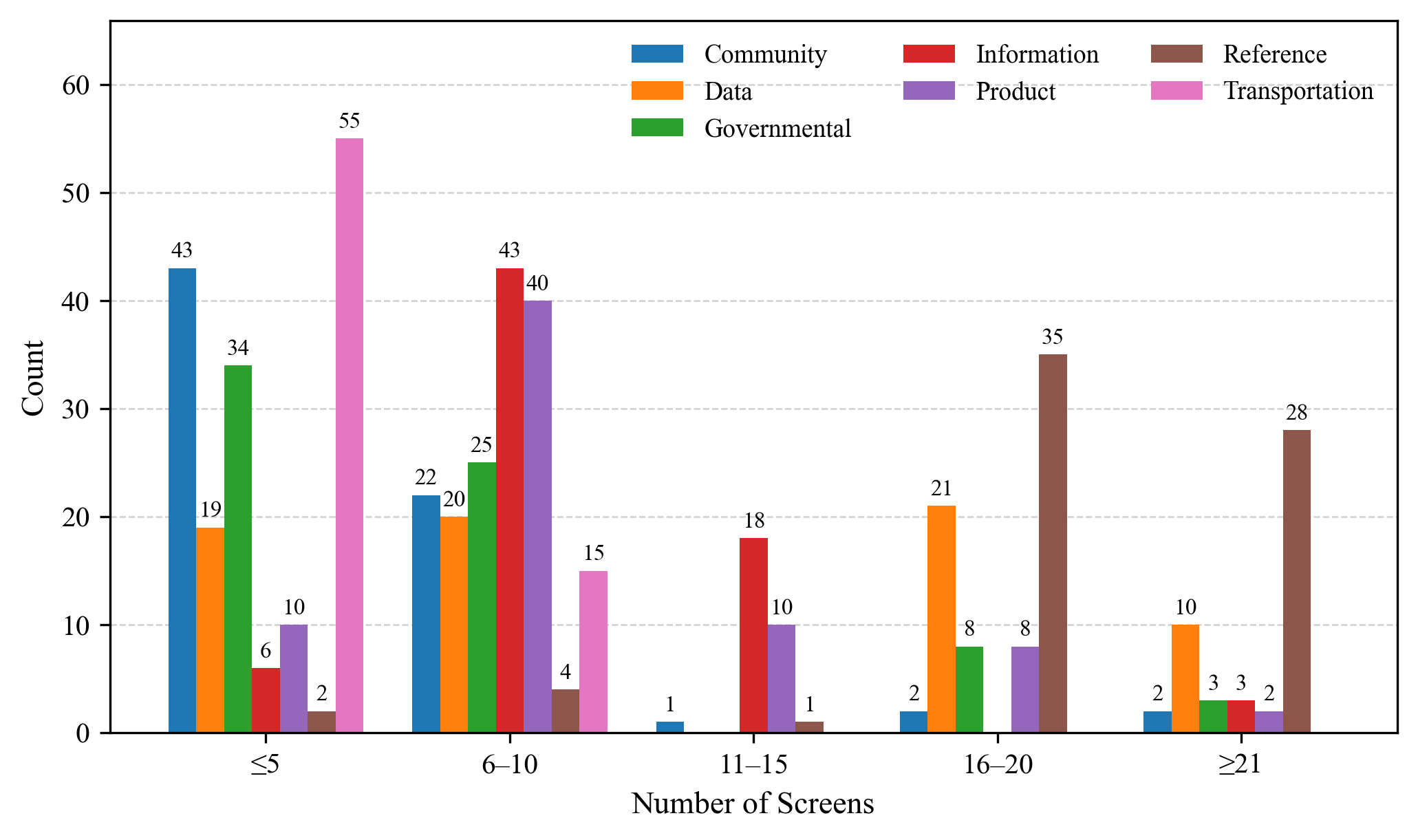}
    \caption{Distribution of Webpage Screenshot Lengths across Categories}
    \label{fig:screen_dist_by_category}
\end{figure*}

Figure~\ref{fig:screen_dist_by_category} illustrates the distribution of webpage lengths, measured by the number of viewports, across different webpage categories. Overall, the distribution exhibits clear category-dependent patterns, indicating substantial structural diversity among real-world webpages.

Short webpages ($\leq$5 screens) are predominant in categories such as \textit{Transportation}, \textit{Community}, and \textit{Governmental}. In particular, Transportation pages are heavily concentrated in this range, suggesting that such webpages are often designed for rapid access and concise presentation. Community and Governmental pages also tend to be relatively short, reflecting task-oriented or modular layouts.

Medium-length webpages (6--10 screens) are most common in the \textit{Information} and \textit{Product} categories. Information pages show a pronounced peak in this range, indicating that explanatory or editorial-style content typically spans multiple sections while remaining within a moderate overall length. Product pages exhibit a similar pattern, consistent with layouts that combine descriptions, specifications, and auxiliary content.

In contrast, long and very long webpages ($\geq$16 screens) are predominantly observed in the \textit{Reference} category. Reference pages display a clear long-tail distribution, with a substantial number of instances extending beyond 16 screens and a notable portion exceeding 21 screens. This reflects the encyclopedic and documentation-oriented nature of reference webpages, which often contain dense and hierarchically structured content.

The \textit{Data} category exhibits a more balanced distribution across screen ranges, spanning from short to long webpages. This variability suggests that data-centric webpages differ widely in presentation style, ranging from compact dashboards to extensive analytical or tabular pages.

Overall, these category-specific length distributions highlight the heterogeneity of webpage structures captured in the benchmark and motivate the need for evaluation protocols and model designs that can robustly handle diverse page lengths in long-webpage understanding and generation tasks.

 \section{Details of Evaluation Process}
\subsection{More details about our validation process on W-VFR task}
\label{appendix:details_wvfr}
The prompt for Structural Fidelity Validation is provided in Figure \ref{fig:w-ffr-evaluation-1}-\ref{fig:w-ffr-evaluation-2}. We employ \texttt{gpt-4o-2024-11-20} as the judge model for automated evaluation. For models that do not support input settings, we assign a default score of 0. The final score is computed as a weighted sum of scores across all evaluation dimensions. Specifically, for the \textit{Visual Style} dimension, the score is computed as a weighted combination of $0.8 \times \text{VLM score} + 0.2 \times \text{normalized DINO score}$, a weighting scheme we empirically validated to achieve the highest agreement with human judgments.
\subsection{More metric details about W-VFR}
\label{appendix:metrics}
\paragraph{Input Support Rate (ISR).}
ISR measures the fraction of test cases that a model or its serving API accepts for generation, isolating input-side constraints (e.g., image-size limits or resolution-size limits) from downstream generation quality. Let $\mathcal{T}$ denote the full test set and $\mathcal{S}_m \subseteq \mathcal{T}$ the subset that model $m$ accepts and proceeds to generate for. ISR is defined as
\begin{equation}
\textsc{ISR}(m) \;=\; \frac{|\mathcal{S}_m|}{|\mathcal{T}|}.
\end{equation}
A low ISR reflects upstream limitations that prevent the model from being applied to certain inputs at all.

\paragraph{HTML Validity Rate (HVR).}
Given the supported subset $\mathcal{S}_m$, HVR measures the proportion of generations that yield a valid, renderable HTML artifact. An output is counted as valid if it is non-empty and contains a well-formed \texttt{<html>} document parseable by a standard HTML parser (with no truncation or non-HTML content). Let $\mathcal{V}_m \subseteq \mathcal{S}_m$ denote the valid outputs from model $m$. HVR is defined as
\begin{equation}
\textsc{HVR}(m) \;=\; \frac{|\mathcal{V}_m|}{|\mathcal{S}_m|}.
\end{equation}
HVR is defined as the model's ability to reliably emit a renderable artifact on supported inputs.

\noindent \textbf{Code Quality (CQ)}
\label{sec:code-quality}
We assess the engineering quality of model-generated HTML using \textbf{Google Lighthouse} (v12), a widely adopted, fully automated, and reproducible web auditing tool that scores a page on a $[0, 100]$ scale across multiple categories.
\paragraph{Evaluation dimensions.}
We adopt Lighthouse's four default categories:
\begin{itemize}
    \item \textbf{Performance}: page load efficiency measured by core Web Vitals.
    \item \textbf{Accessibility}: powered by the \texttt{axe-core} rule set, covering image \texttt{alt} attributes, form \texttt{<label>} associations, color contrast, the \texttt{<html lang>} declaration, ARIA landmark usage, and so on.
    \item \textbf{Best Practices}: HTML document conventions (\texttt{<!DOCTYPE>}, \texttt{<meta charset>}, \texttt{<meta viewport>}), absence of console errors, deprecated APIs, CSP declaration, and image aspect-ratio consistency.
    \item \textbf{SEO}: presence of \texttt{<title>}, \texttt{<meta description>}, crawler-friendly links, and \texttt{robots.txt} compliance.
\end{itemize}

\paragraph{Audit pipeline.}
Each HTML file is served via a local static HTTP server bound to \texttt{127.0.0.1} and audited through the Lighthouse Node API driving a headless Chromium instance launched by Puppeteer. 

\paragraph{Input normalization.}
To prevent surface artifacts from affecting Lighthouse parsing, we apply a lightweight normalization step on raw outputs: residual Markdown code fences (e.g., leading/trailing \texttt{\textasciigrave\textasciigrave\textasciigrave html}) are stripped, and missing \texttt{<!DOCTYPE html>} declarations are restored for files whose HTML body is otherwise complete. Outputs that already failed at the generation stage (empty files, OOM placeholders, or streaming truncations yielding non-HTML content) are \emph{left untouched} so that they are scored as genuine failures by Lighthouse, contributing to the model's robustness profile.

\paragraph{Final score.}
For each file, we average the four Lighthouse category scores to obtain a single overall score $S_{m, f} \in [0, 100]$. The code-quality score of model $m$ is the mean of $S_{m, f}$ over the common subset:
\begin{equation}
\textsc{CodeQuality}(m) \;=\; \frac{1}{N} \sum_{f \in \mathcal{F}_{\text{common}}} S_{m, f}.
\end{equation}
where $N= |\mathcal{F}_{\text{common}}|$.
\subsection{More details about W-FFR process}
\label{appendix:details_wffr}
\noindent \textbf{W-FFR Evaluation Pipeline}
To assess whether a model-generated HTML page supports correct user interactions, we design an automated GUI evaluation pipeline that combines a browser-based environment with an LLM-as-judge agent.

\paragraph{Environment.}
Each HTML file is served by a dedicated environment process,
which launches a headless Chromium browser via Playwright and exposes a local HTTP
API\@. All operations are addressed by
numeric element IDs assigned at runtime: a JavaScript traversal of the DOM
identifies interactive elements (standard form controls, elements with
\texttt{role=button}, elements styled with \texttt{cursor:pointer}, etc.) and
builds an $\text{id} \to \text{XPath}$ mapping used to locate and manipulate
elements.

\paragraph{Tasks.}
Each HTML file is paired with a JSON task file specifying one or more tasks, each
decomposed into a sequence of atomic steps. Each step has a type, a natural-language description of the intended interaction, and an optional value (for input/select steps).

\paragraph{Evaluation Loop.}
For each task, the page is reset to its initial state before execution begins.
Steps are executed sequentially. Each step proceeds in three stages:

\begin{enumerate}
  \item \textbf{Predict.} A screenshot and the serialized DOM tree (with numeric
  IDs) are passed to an LLM\@. The model is prompted to identify the target element
  and output a structured action command in \texttt{\textbackslash boxed\{\}}
  format (e.g., \texttt{\textbackslash boxed\{click[42]\}} or
  \texttt{\textbackslash boxed\{enter[7][hello]\}}). The command is parsed via
  regex to extract the action type, element ID, and value.

  \item \textbf{Execute.} The parsed action is dispatched to the browser
  environment via the corresponding HTTP endpoint. The environment resolves the
  element by XPath and performs the operation.

  \item \textbf{Verify.} Screenshots captured before and after execution are
  passed to a second LLM call. The model compares the two images and judges
  whether the intended change occurred (e.g., a field was filled, a dropdown
  changed, a new element appeared), outputting \texttt{\textbackslash
  boxed\{success\}} or \texttt{\textbackslash boxed\{fail\}}. 
\end{enumerate}

The prompts used in this process are provided in Appendix \ref{appendix:prompts}.

\section{Details about Evaluation Performance}
\label{appendix:each-dimension_vali_visul}
Tables~\ref{tab:single_image_evaluation_each_dimension} and~\ref{tab:multiple_images_evaluation} summarize model performance across different evaluation dimensions under the single-image and multi-image input settings, respectively. Overall, the results show that the multi-image setting consistently improves model performance across most dimensions and models, particularly for long and structurally complex webpages.
\begin{table*}[h!]
\centering
\caption{The evaluation result of each dimension under single-image input setting}
\begin{tabular}{lcccccc}
\toprule
Model & \makecell[c]{Page Scale\\and Length} & \makecell[c]{Global\\Layout} & \makecell[c]{Section\\Hierarchy} & \makecell[c]{Visual\\Styling} & \makecell[c]{Information\\Density} & \makecell[c]{Overall\\Score} \\
\midrule
\multicolumn{7}{c}{\textbf{Open-source VLM}} \\
\midrule
Kimi-VL-A3B-Thinking & 2.06 & 3.40 & 2.67 & 3.86 & 2.34 & 2.87 \\
Qwen3-VL-8B-Instruct & 3.32 & 4.02 & 3.36 & 3.75 & 2.91 & 3.47 \\
InternVL3-78B & 2.41 & 4.25 & 3.30 & 4.22 & 2.82 & 3.40 \\
GLM-4.1V-9B-Thinking-Flash & 5.04 & 5.62 & 4.67 & 4.92 & 4.11 & 4.87 \\
GLM-4.6V & 5.54 & 6.58 & 5.63 & 5.59 & 4.94 & 5.66 \\
Qwen3-VL-235B-A22B-Instruct & 5.29 & 5.75 & 5.08 & 4.95 & 4.44 & 5.10 \\
\midrule
\multicolumn{7}{c}{\textbf{Closed-source VLM}} \\
\midrule
Claude-Sonnet-4-5-Thinking & 3.85 & 4.39 & 3.91 & 3.49 & 3.39 & 3.81 \\
Claude-Opus-4-5-20251101 & 3.95 & 4.38 & 3.92 & 3.52 & 3.43 & 3.84 \\
GPT-5.2 & 7.02 & 7.33 & 6.37 & 6.00 & 5.77 & 6.50 \\
GPT-4o & 1.92 & 4.16 & 3.01 & 4.60 & 2.45 & 3.23 \\
Gemini-3-Pro-preview & 6.65 & 7.87 & 6.95 & 6.38 & 5.99 & 6.77 \\
Gemini-3-Flash-preview & 6.28 & 7.43 & 6.44 & 6.05 & 5.58 & 6.36 \\
Doubao-Seed-1-6 & 2.97 & 4.60 & 3.62 & 4.02 & 3.10 & 3.66 \\
\bottomrule
\end{tabular}
\label{tab:single_image_evaluation_each_dimension}
\end{table*}

\begin{table*}[h!]
\centering
\caption{The evaluation result of each dimension under multi-image input setting}
\begin{tabular}{lcccccc}
\toprule
Model & \makecell[c]{Page Scale\\and Length} & \makecell[c]{Global\\Layout} & \makecell[c]{Section\\Hierarchy} & \makecell[c]{Visual\\Styling} & \makecell[c]{Information\\Density} & \makecell[c]{Overall\\Score} \\
\midrule
\multicolumn{7}{c}{\textbf{Open-source VLM}} \\
\midrule
Kimi-VL-A3B-Thinking & 2.45 & 3.71 & 2.87 & 4.03 & 2.52 & 3.12 \\
Qwen3-VL-8B-Instruct & 3.69 & 4.56 & 3.89 & 4.24 & 3.35 & 3.95 \\
InternVL3-78B & 2.62 & 4.51 & 3.51 & 4.40 & 3.05 & 3.62 \\
GLM-4.1V-9B-Thinking-Flash & 0.56 & 0.93 & 0.65 & 1.97 & 0.58 & 0.94 \\
GLM-4.6V & 5.65 & 7.07 & 6.03 & 5.89 & 5.18 & 5.96 \\
Qwen3-VL-235B-A22B-Instruct & 5.50 & 6.90 & 5.93 & 5.79 & 5.13 & 5.85 \\
\midrule
\multicolumn{7}{c}{\textbf{Closed-source VLM}} \\
\midrule
Claude-Sonnet-4-5-Thinking & 6.56 & 7.83 & 6.90 & 6.41 & 5.97 & 6.74 \\
Claude-Opus-4-5-20251101 & 6.14 & 7.59 & 6.58 & 6.20 & 5.63 & 6.43 \\
GPT-5.2 & 5.33 & 7.46 & 6.18 & 6.45 & 5.28 & 6.14 \\
GPT-4o & 2.02 & 4.58 & 3.30 & 4.93 & 2.79 & 3.53 \\
Gemini-3-Pro-preview & 7.12 & 8.26 & 7.48 & 6.72 & 6.50 & 7.22 \\
Gemini-3-Flash-preview & 6.15 & 7.68 & 6.58 & 6.53 & 5.71 & 6.53 \\
Doubao-Seed-1-6 & 3.58 & 6.05 & 4.69 & 5.48 & 4.03 & 4.77 \\
\bottomrule
\end{tabular}
\label{tab:multiple_images_evaluation}
\end{table*}
\subsection{Overall trends}
From the Overall Score, most models benefit substantially from the multi-image setting. By segmenting long webpage screenshots into multiple images, models are able to perceive finer-grained visual details and reduce information loss caused by resolution compression in single-image inputs. This improvement is especially pronounced for strong closed-source models, such as the Gemini and Claude families, whose overall scores increase consistently across settings.

\subsection{Page scale and length} 
Page Scale and Length results show different trends across models. Under roughly comparable image processing conditions, the GPT-5.2 model actually experiences a performance drop under multi-image input settings, whereas Gemini-3-Pro shows some improvement. In contrast, the Claude model exhibits a significant performance increase, as the number of successfully processed images is greatly enhanced under the multi-image input setting.

\subsection{Global layout and  section hierarchy }
Performance on Global Layout and Section Hierarchy also improves noticeably in the multi-image setting in most cases. These gains indicate that access to multiple localized views helps models better infer high-level structural organization, including module ordering and hierarchical relationships between sections. The effect is particularly evident for the Claude and Gemini models, which show substantial improvements on these structure-oriented dimensions.

\subsection{Visual styling}
In contrast, improvements on  Visual Styling are still evident. Multi-image inputs provide additional visual details, which help models better capture fine-grained stylistic attributes—such as font choices, color schemes, and spacing—even though these aspects remain challenging to infer from screenshots alone. Consequently, this dimension also shows positive performance gains.

\subsection{Information density}
Information Density lies between structural and stylistic dimensions in terms of evaluation complexity. Most models achieve modest but consistent improvements in the multi-image setting, reflecting the combined influence of enhanced structural perception and the inherent ambiguity of density judgments from visual inputs.

\subsection{Open-source vs. Closed-source models.}
Almost all Closed-source models consistently outperform open-source models under both settings, and their advantage becomes more pronounced in the multi-image scenario. Notably, Gemini-3-Pro-preview achieves the highest overall performance and demonstrates strong, balanced improvements across all dimensions.

\subsection{Anomalous behavior.}
An exception is observed for GLM-4.1V-9B-Thinking-Flash, which exhibits a sharp performance degradation under the multi-image setting. Manual inspection reveals that, when presented with multiple images, this model tends to misinterpret the task as site-level webpage generation rather than long-page reconstruction, resulting in severely degraded outputs. This highlights that effective utilization of multi-image inputs requires robust cross-image integration and task alignment, which are not uniformly supported across models.

\subsection{Summary}
In summary, these results demonstrate that multi-image input is a critical design choice for evaluating and generating long webpages, offering substantial benefits for scale- and structure-related dimensions. While stylistic fidelity remains challenging due to limited design-level information, the overall performance gains underscore the importance of segmented visual inputs for long-context webpage understanding and generation.

\section{Human Evaluation}
\label{appendix:human_evaluation}
\subsection{The score of each dimension on human relevance}
Table \ref{tab:correlation_ai} reports the correlation between the automatic evaluation scores and human judgments averaged across annotators. Overall, the automatic scores exhibit a strong correlation with human assessments, achieving a Pearson correlation coefficient of 0.8335, which indicates a high degree of alignment between the proposed evaluation protocol and human perception.

Among the individual dimensions, Page Scale and Length show a very strong correlation (0.9150), suggesting that both humans and models consistently assess page completeness and overall length. Global Layout (0.7706) and Chapter Hierarchy (0.7393) also demonstrate strong correlations, reflecting reliable agreement on the structural organization and hierarchical arrangement of webpage content. Information Density achieves a moderately high correlation (0.6807), indicating that the automatic metric can reasonably capture the perceived amount and distribution of information on long webpages. Visual Style achieves a correlation of 0.6035, reflecting a moderately strong agreement between automatic scores and human judgments. This is expected given the screenshot-only input setting adopted in our experiments, as fine-grained stylistic attributes—such as font families, color schemes, spacing, and design system constraints—are not explicitly accessible without the original HTML or CSS sources, making stylistic fidelity inherently more challenging to assess from rendered images alone. Nevertheless, the correlation remains meaningful, suggesting that the automatic metric can capture overall visual style consistency to a reasonable degree.

Overall, these results indicate that the proposed automatic evaluation aligns well with human judgments, while the reduced correlation on visual style reflects the intrinsic subjectivity of aesthetic assessment and the limited availability of design-level information in the current task setting.
\begin{table}[ht]
\centering
\caption{Correlation between person\_average and AI Models}
\begin{tabular}{lcc}
\toprule
\textbf{Metric} & \textbf{Coefficient} & \textbf{p-value} \\
\midrule
Overall Score & 0.8335 & $<$0.0001 \\
Page Scale and Length & 0.9150 & $<$0.0001 \\
Global Layout & 0.7706 & $<$0.0001 \\
Chapter Hierarchy & 0.7393 & $<$0.0001 \\
Visual Style & 0.6035 & $<$0.0001 \\
Information Density & 0.6807 & $<$0.0001 \\
\bottomrule
\end{tabular}
\label{tab:correlation_ai}
\end{table}
\subsection{Inter-annotator agreement analysis}
\begin{figure*}[t]
\centering

\begin{subfigure}[t]{0.32\textwidth}
    \centering
    \includegraphics[width=\linewidth]{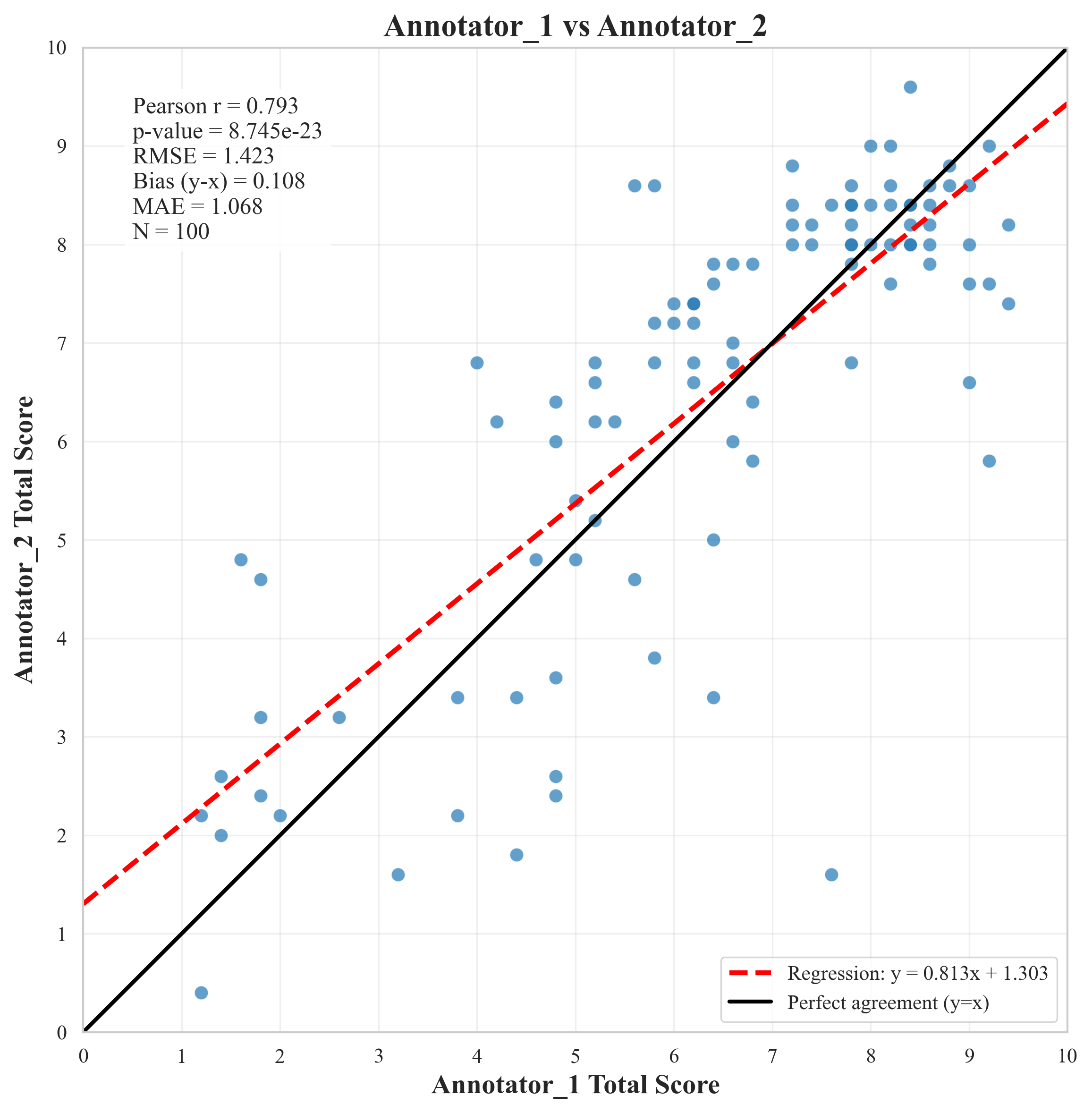}
    \caption{Annotator 1 vs Annotator 2}
\end{subfigure}
\hfill
\begin{subfigure}[t]{0.32\textwidth}
    \centering
    \includegraphics[width=\linewidth]{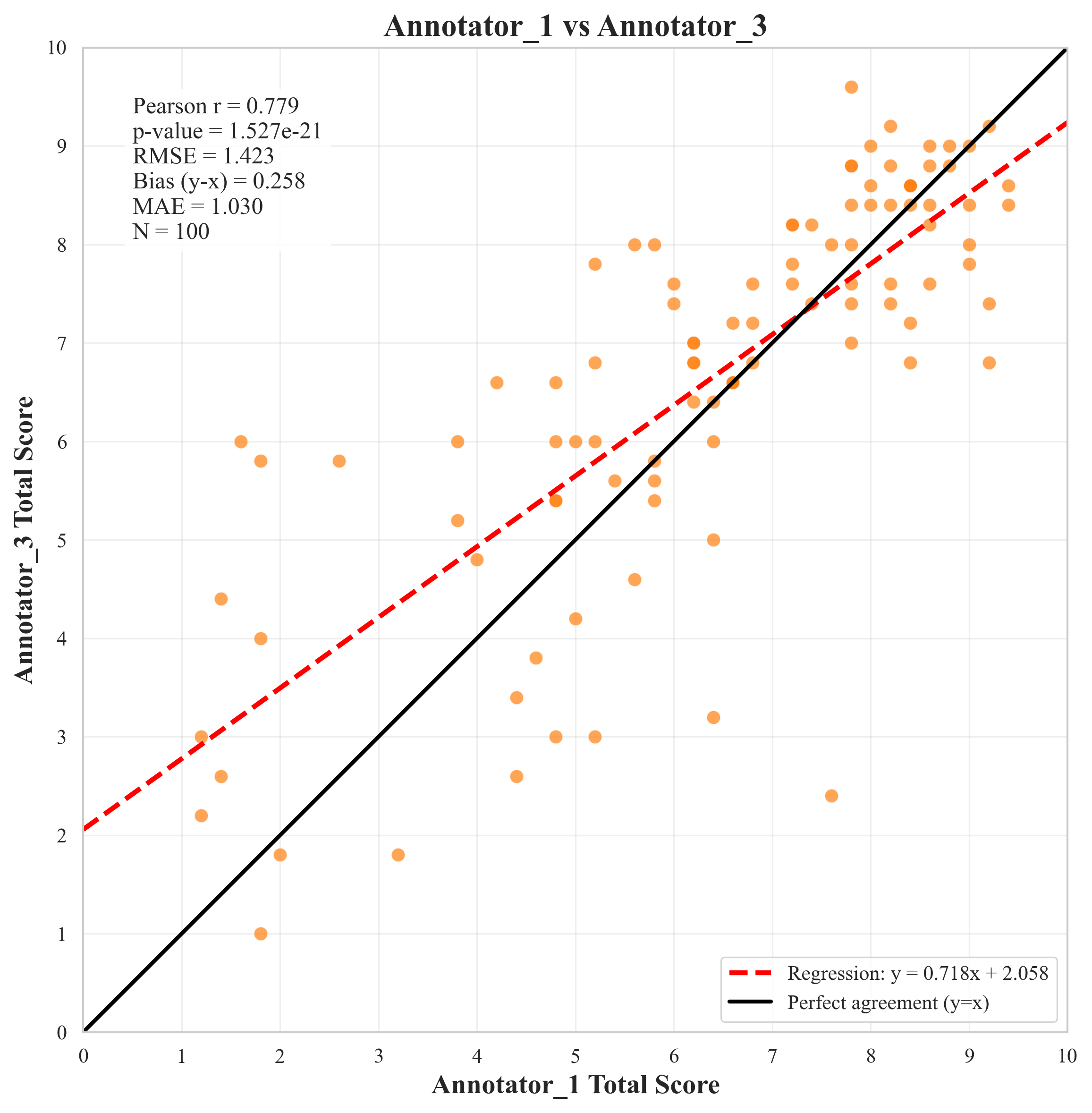}
    \caption{Annotator 1 vs Annotator 3}
\end{subfigure}
\hfill
\begin{subfigure}[t]{0.32\textwidth}
    \centering
    \includegraphics[width=\linewidth]{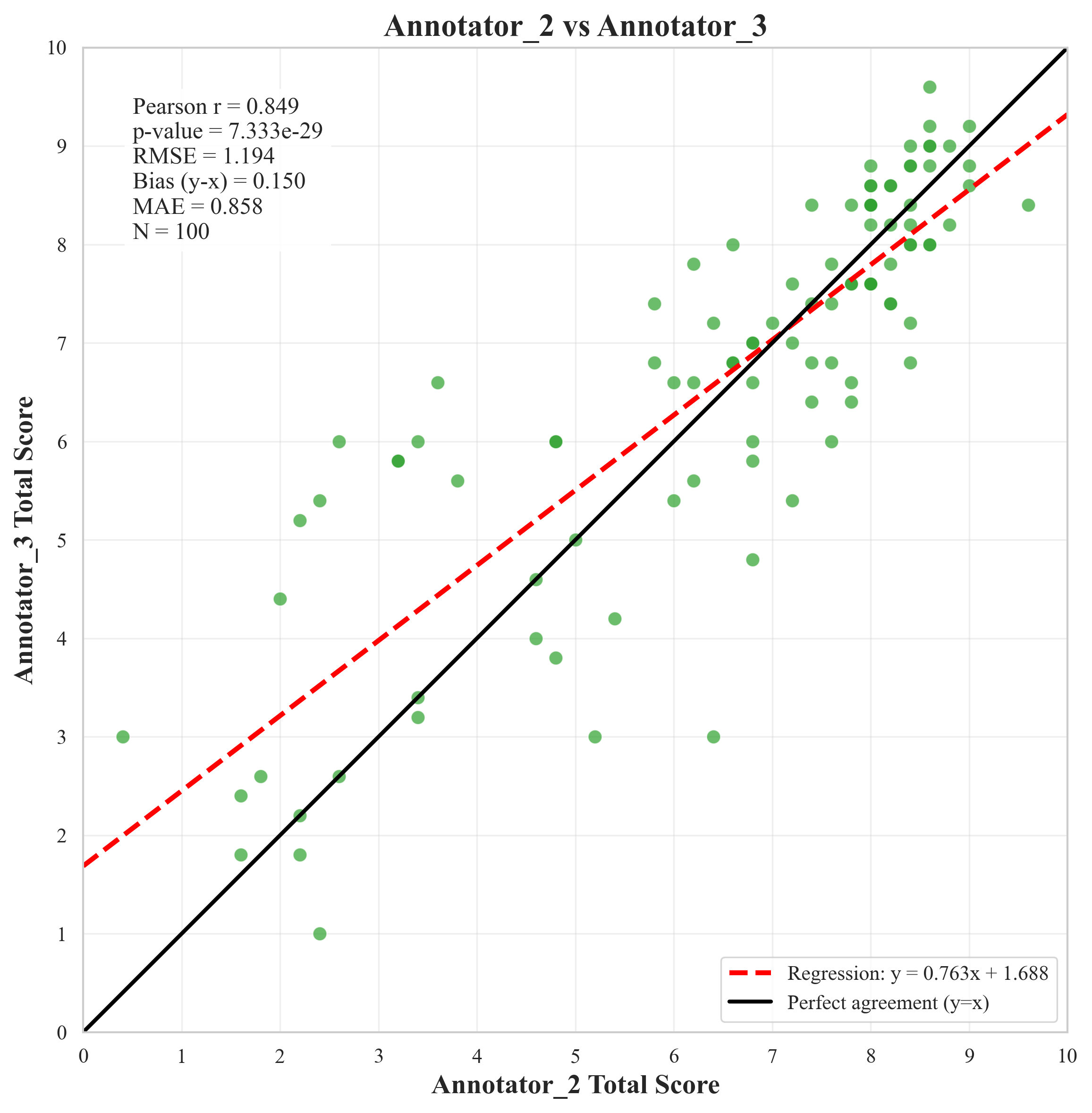}
    \caption{Annotator 2 vs Annotator 3}
\end{subfigure}

\vspace{2mm}

\begin{subfigure}[t]{0.32\textwidth}
    \centering
    \includegraphics[width=\linewidth]{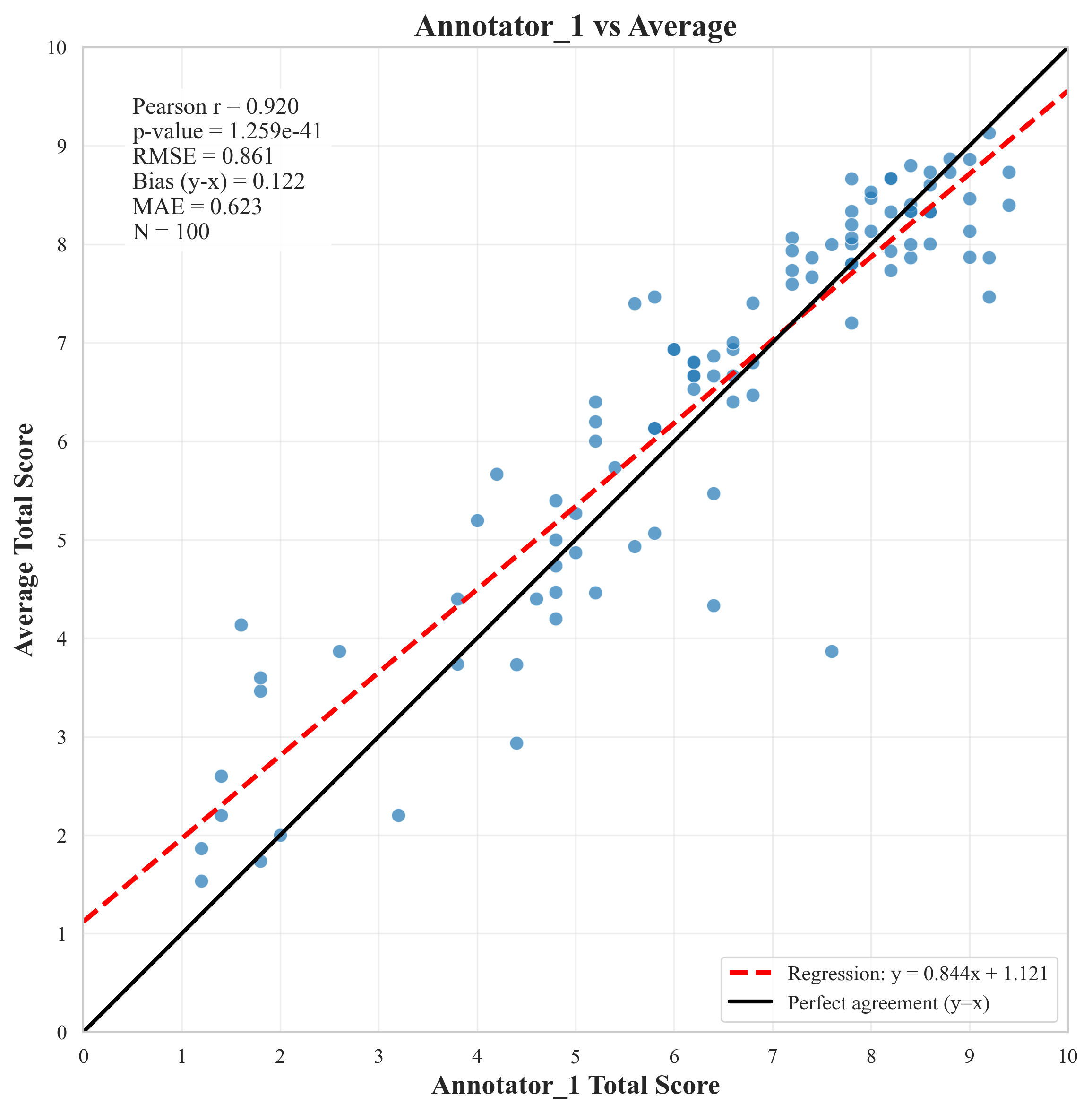}
    \caption{Annotator 1 vs Avg Score}
\end{subfigure}
\hfill
\begin{subfigure}[t]{0.32\textwidth}
    \centering
    \includegraphics[width=\linewidth]{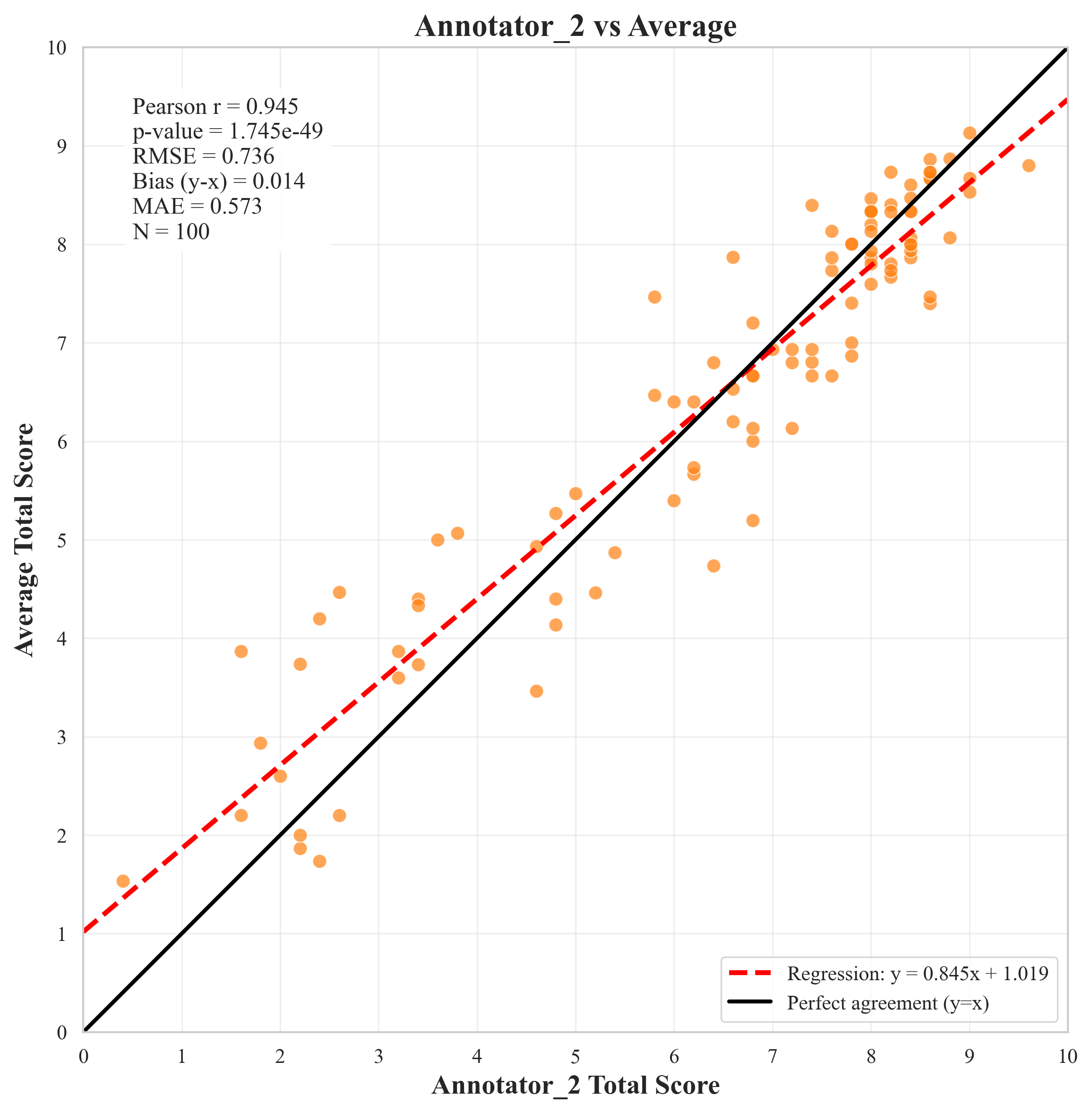}
    \caption{Annotator 2 vs Avg Score}
\end{subfigure}
\hfill
\begin{subfigure}[t]{0.32\textwidth}
    \centering
    \includegraphics[width=\linewidth]{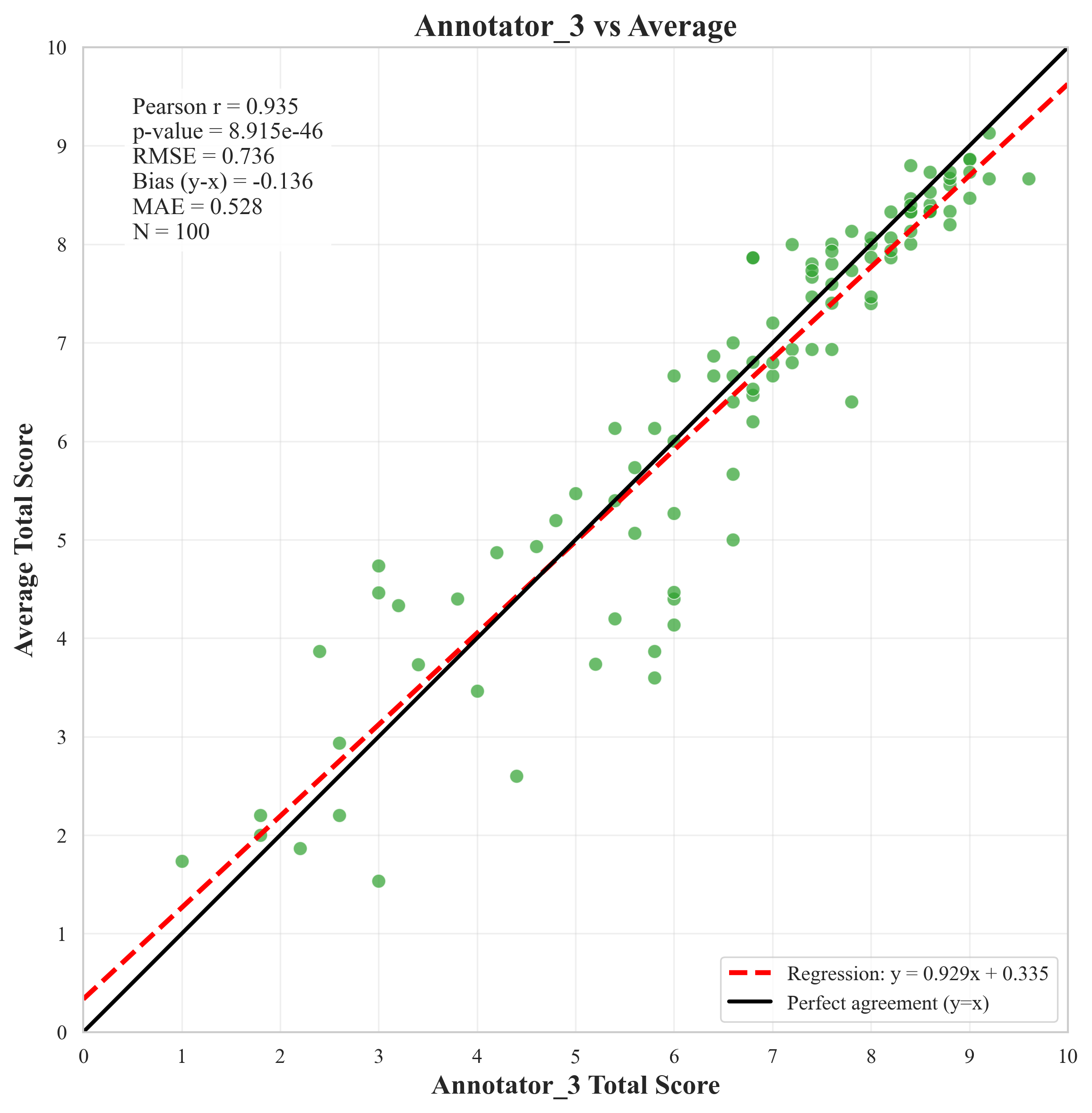}
    \caption{Annotator 3 vs Avg Score}
\end{subfigure}

\caption{Inter-Annotator Agreement Analysis.}
\label{fig:annotator_agreement}
\end{figure*}

Figure~\ref{fig:annotator_agreement} presents pairwise comparisons of total scores assigned by three human annotators, as well as comparisons between each annotator and the averaged score across annotators. Each point corresponds to one webpage instance, the black diagonal line indicates perfect agreement ($y=x$), and the red dashed line denotes the fitted linear regression.

Overall, the results demonstrate a high level of inter-annotator consistency. Pairwise correlations between individual annotators are strong, with Pearson correlation coefficients of $r=0.793$ (Annotator~1 vs.\ Annotator~2), $r=0.779$ (Annotator~1 vs.\ Annotator~3), and $r=0.849$ (Annotator~2 vs.\ Annotator~3), all with statistically significant $p$-values ($p<10^{-20}$). The corresponding RMSE and MAE values indicate moderate absolute disagreement, reflecting the inherent subjectivity of fine-grained visual fidelity assessment.

Comparisons between individual annotators and the averaged score further reveal substantially higher agreement. Annotator~1, Annotator~2, and Annotator~3 achieve Pearson correlations of $r=0.920$, $r=0.945$, and $r=0.935$ with the average score, respectively. In addition, both RMSE and MAE are markedly reduced in these comparisons, suggesting that score averaging effectively suppresses individual annotation noise and yields a more stable reference signal.

The regression lines across all plots exhibit slopes slightly below unity and small intercept offsets, indicating mild systematic differences in score calibration among annotators. Nevertheless, the bias values remain close to zero, suggesting no severe over- or under-estimation tendencies for any individual annotator.

Taken together, these results confirm that the proposed evaluation protocol yields reliable and reproducible human judgments. The strong inter-annotator agreement and the improved consistency obtained through averaging support the use of the averaged human score as the ground-truth reference in subsequent correlation analyses with model predictions.

\section{Prompts and Model Settings}
\subsection{ Prompts}
\label{appendix:prompts}

\textbf{Task Generation Prompts}
\label{appendix:w-vfr-task-generatioin_prompts}
We present the prompts used for long webpage generation and verification, as shown in Figures \ref{fig:w-ffr-generatioin-task-1}--\ref{fig:w-ffr-generatioin-task-3}.

\noindent \textbf{W-VFR Generation Prompts}
\label{appendix:w-vfr-generatioin_prompts}
Generation prompts for W-VFR under single-image input setting and multi-image input setting are shown in Figures~\ref{fig:w-vfr-generatioin-1}--\ref{fig:w-vfr-generatioin-2} and Figures~\ref{fig:w-vfr-generatioin-multi-1}--\ref{fig:w-vfr-generatioin-multi-2}, respectively.

\noindent \textbf{W-VFR Evaluation Prompts}
\label{appendix:evaluation_prompts}
Evaluation prompts for W-VFR  are shown in Figures~\ref{fig:w-vfr-evaluation-1}--\ref{fig:w-vfr-evaluation-3} .

\noindent \textbf{W-FFR Generation Prompts}
\label{appendix:evaluation_prompts}
Generation prompts for W-FFR under single-image input setting and multi-image input setting are shown in Figures~\ref{fig:w-ffr-generation-1}--\ref{fig:w-ffr-generation-2} and Figures~\ref{fig:w-ffr-generation-1-multi}--\ref{fig:w-ffr-generation-2-multi}, respectively.

\noindent \textbf{W-FFR Evaluation Prompts}
\label{appendix:evaluation_prompts}
Evaluation prompts for W-VFR  are shown in Figures~\ref{fig:w-ffr-evaluation-1}--\ref{fig:w-ffr-evaluation-2} .

\begin{figure*}[t!]
\centering
\begin{tcolorbox}[
    colback=gray!2,
    colframe=blue!70!black,
    colbacktitle=blue!12!white,
    title=\textbf{Prompt for HTML webpage Task generation for W-FFR. (Part 1)},
    fonttitle=\bfseries,
    coltitle=blue!50!black,
    enhanced,
    sharp corners,
    boxrule=0.5pt,
    left=6pt,
    right=6pt,
    top=5pt,
    bottom=5pt,
    titlerule=0mm,
    width=0.99\textwidth, 
    title style={left color=blue!8!white, right color=blue!5!white}
]

You are an expert in designing Goal-Oriented Web Task Benchmarks for long webpages with complex interactions. Your focus is on modeling realistic, end-to-end user goals rather than atomic navigation actions. 

Inputs: \\
- A full-length webpage screenshot (the page may require vertical scrolling to be fully visible). \\
- (Optional) A virtual account credential, which may only be used if login-related UI elements are explicitly visible in the screenshot:
\begin{verbatim}
{
  "benchmark_account": {
    "username": "test_user_001",
    "email": "test_user_001@example.com",
    "password": "P@ssw0rd!001"
  }
}
\end{verbatim}
Critical clarifications (must be followed): \\
- The tasks you design are not atomic UI navigation steps, but complex, real-world user goals. \\
- Each task must include: \\
  - A clearly defined final goal \\
  - Multi-step interactions (e.g., filtering, configuration, comparison, decision-making, submission) \\
  - Interactions spanning multiple page regions or modules  \\
  - A final goal with explicit and verifiable success signals 

Examples (for style reference only; do not replicate): \\
- Selecting the cheapest direct flight among all available options \\
- Configuring a computer with 16GB RAM and 2TB storage and adding it to the cart \\
- Finding the hotel with the highest rating and lowest price in the current city \\
- Locating and confirming the default value of a key parameter in documentation 

Objective: \\
Based on the screenshot, infer a plausible interactive webpage and design 3--4 complex interaction tasks.

Task design requirements:\\
- Each task represents a realistic user goal rather than a single-button action \\
- Each task contains at least three semantically meaningful interaction steps\\
- Tasks are completable within a long webpage (scrolling, filtering, tab switching allowed) \\
- Filtering, comparison, or decision-making tasks must include explicit numeric or discrete outcomes
- Key success values must be derived from task execution through calculation or extraction \\
- All tasks must be automatically verifiable via observable success signals defined in the schema  \\
Authentication task rules: \\
- Authentication-related tasks are allowed only if the screenshot explicitly shows one of the following: \\
  - Login / Log in / Sign in \\
  - Register / Sign up \\
  - Logout / Log out \\
  - User avatar / Account menu / Profile center \\
- If none of the above are visible: \\
  - Do not design any login, registration, or logout tasks \\
  - All tasks must be executable in a logged-out state \\
Output requirements: \\
- Output only one JSON object \\
- Do not include Markdown

\end{tcolorbox}

\captionof{figure}{This prompt is designed for HTML webpage Task generation for W-FFR (Part 1).}
\label{fig:w-ffr-generatioin-task-1}
\end{figure*}

\begin{figure*}[t!]
\centering
\begin{tcolorbox}[
    colback=gray!2,
    colframe=blue!70!black,
    colbacktitle=blue!12!white,
    title=\textbf{Prompt for HTML webpage Task generation for W-FFR. (Part 2)},
    fonttitle=\bfseries,
    coltitle=blue!50!black,
    enhanced,
    sharp corners,
    boxrule=0.5pt,
    left=6pt,
    right=6pt,
    top=5pt,
    bottom=5pt,
    titlerule=0mm,
    width=0.99\textwidth, 
    title style={left color=blue!8!white, right color=blue!5!white}
]

- Do not include explanatory text \\
- The output must strictly conform to the schema below

Schema (complex goal-oriented task version):
\begin{verbatim}
{
  "page_summary": {
    "site_or_product": "...",
    "page_type": "e-commerce | ticketing | travel | documentation
    | SaaS | news | comparison | other",
    "primary_modules": [
      "Clearly visible main modules in the screenshot"
    ],
    "notable_interactives": [
      "Key interactive controls enabling complex goals"
    ],
    "auth_visible": true | false,
    "auth_evidence": [
      "Visible evidence proving authentication presence, if any"
    ]
  },
  "benchmark_account": {
    "username": "test_user_001",
    "email": "test_user_001@example.com",
    "password": "P@ssw0rd!001"
  },
"tasks": [
    {
      "task_id": "T1",
      "task_type": "purchase | configuration | comparison | selection |
      information_verification |
      auth_dependent | mixed",
      "task_name": "Concise summary of the task objective",
      "user_goal": "One-sentence description of the real user’s final goal",
      "instruction": "Goal-oriented instruction to the agent (in English)",
      "constraints": [
        "Mandatory business or logical constraints"
      ],
      "preconditions": [
        "e.g., logged-out state, filter panel expanded, city selected"
      ],
      "steps": [
        {
          "action": "click | type | select | scroll | compare",
          "target_description": "UI target described by location and 
          semantic meaning",
          "value": "Specific input or selection value, or null"
        }
      ],

\end{verbatim}

\end{tcolorbox}

\captionof{figure}{This prompt is designed for HTML webpage Task generation for W-FFR (Part 2).}
\label{fig:w-ffr-generatioin-task-2}
\end{figure*}

\begin{figure*}[t!]
\centering
\begin{tcolorbox}[
    colback=gray!2,
    colframe=blue!70!black,
    colbacktitle=blue!12!white,
    title=\textbf{Prompt for HTML webpage Task generation for W-FFR. (Part 3)},
    fonttitle=\bfseries,
    coltitle=blue!50!black,
    enhanced,
    sharp corners,
    boxrule=0.5pt,
    left=6pt,
    right=6pt,
    top=5pt,
    bottom=5pt,
    titlerule=0mm,
    width=0.99\textwidth, 
    title style={left color=blue!8!white, right color=blue!5!white}
]

\begin{verbatim}
     "success_criteria": [
        "At least two observable, automatically verifiable success 
        signals with explicit quantitative or discrete values"
      ],
     
        "verification_hints": {
        "text_assertions": [
          "Key texts or values expected after successful completion"
        ],
        "layout_assertions": [
          "Observable layout or state changes"
        ],
        "selector_strategy": [
          "Prefer semantic regions for configuration or pricing",
          "Fallback to aria/role",
          "Lastly, structured XPath"
        ],
        "suggested_selectors": [
          {
            "purpose": "Core element for final verification",
            "css": null,
            "xpath": null,
            "aria": null
          }
        ]
      },
      "difficulty": "medium | hard | very_hard",
      "coverage_tags": [
        "filtering",
        "decision_making",
        "multi_step",
        "long_page_navigation"
      ]
    }
  ]
}
\end{verbatim}

\end{tcolorbox}

\captionof{figure}{This prompt is designed for HTML webpage Task generation for W-FFR (Part 3).}
\label{fig:w-ffr-generatioin-task-3}
\end{figure*}

\begin{figure*}[t!]
\centering
\begin{tcolorbox}[
    colback=gray!2,
    colframe=blue!70!black,
    colbacktitle=blue!12!white,
    title=\textbf{Prompt for HTML webpage generation under the single-image input setting for W-VFR. (Part 1)},
    fonttitle=\bfseries,
    coltitle=blue!50!black,
    enhanced,
    sharp corners,
    boxrule=0.5pt,
    left=6pt,
    right=6pt,
    top=5pt,
    bottom=5pt,
    titlerule=0mm,
    width=0.99\textwidth, 
    title style={left color=blue!8!white, right color=blue!5!white}
]
  You are a Web Visual Replication Specialist.

Your task is: given a full long webpage screenshot, generate a standalone HTML file that can be opened directly in a browser, reproducing the screenshot’s visual appearance, page structure, and page scale as accurately as possible.

This is a Visual Fidelity task. Real business interaction logic is not evaluated at this stage.\\
--------------------------------\\
Input:\\
--------------------------------\\
- One full long webpage screenshot (a vertical long image containing the entire page) \\
- The real pixel dimensions of the screenshot:
  \indent- Target page width (pixels): \{TARGET\_PAGE\_WIDTH\_PX\} \\
  \indent- Target page height (pixels): \{TARGET\_PAGE\_HEIGHT\_PX\}\\
-------------------------------- \\
Output Requirements (must strictly follow)\\
--------------------------------\\
- You must output one and only one complete HTML file content \\
- The output must start with $<!doctype html>$ \\
- The output must end with $</html>$ \\
- Do not output any explanatory text, comments, Markdown, or code fences
- The output must be directly savable as index.html and open in a desktop browser\\
--------------------------------\\
Implementation Constraints\\
--------------------------------\\
- Use native HTML / CSS / JavaScript only \\
- All styles must be defined inline within a <style> block \\
- Do not use any third-party libraries, frameworks, CDNs, or build tools\\
- Do not reference any external images, fonts, or resources \\
- For images / avatars / logos in the screenshot:\\
  \indent- Recreate them using SVG or CSS placeholder blocks \\
  \indent- Keep their visual size and position\\
  \indent- Keep semantics as close as possible to the original content \\
-------------------------------- \\
Core Visual Replication Requirements \\
-------------------------------- \\
1) Page Structure and Layout \\
   - Strictly replicate the information architecture and module order shown in the screenshot, such as: \\
    \indent - Top navigation / secondary navigation \\
     \indent- Sidebar (if present) \\
     \indent- Main content area \\
    \indent - Card lists / tables / section modules \\
     \indent- Footer (if present) \\
   - The relative positions, width proportions, and layering relationships of modules must match the screenshot 

2) Style and Visual Elements \\
   - Reproduce the overall color scheme (background, primary, secondary colors) \\

\end{tcolorbox}

\captionof{figure}{This prompt is designed for the single-image input setting, where a single full-page webpage screenshot is provided as input and the model is required to generate a complete HTML webpage that visually replicates the input screenshot.}
\label{fig:w-vfr-generatioin-1}
\end{figure*}

\begin{figure*}[t!]
\centering
\begin{tcolorbox}[
    colback=gray!2,
    colframe=blue!70!black,
    colbacktitle=blue!12!white,
    title=\textbf{Prompt for HTML webpage generation under the single-image input setting for W-VFR. (Part 2)},
    fonttitle=\bfseries,
    coltitle=blue!50!black,
    enhanced,
    sharp corners,
    boxrule=0.5pt,
    left=6pt,
    right=6pt,
    top=5pt,
    bottom=5pt,
    titlerule=0mm,
    width=0.99\textwidth, 
    title style={left color=blue!8!white, right color=blue!5!white}
]
   - Reproduce font hierarchy (title/body/note sizing and weight relationships) \\
   - Reproduce spacing system (padding / margin) \\
   - Reproduce borders, corner radius, shadows, and divider line styles \\
   - The visual density of cards, lists, and tables must match the screenshot\\
3) Page Scale and Dimension Alignment (Critical) \\
   - The generated page must cover all content visible in the screenshot \\
   - The total scroll height should align as closely as possible to \{TARGET\_PAGE\_HEIGHT\_PX\} pixels \\
   - Do not inflate height by meaningless repetition \\
   - If extension beyond the screenshot is necessary to keep the structure complete: \\
     \indent - Only extend using module types and visual styles already present in the screenshot \\
      \indent- Extensions must be reasonable and restrained, and must not break overall proportions \\
     \indent- The overall layout proportions must match the screenshot corresponding to the target width \{TARGET\_PAGE\_WIDTH\_PX\} pixels \\
-------------------------------- \\
Content and Placeholder Rules \\
--------------------------------\\
- Replicate text content and typographic layout as faithfully as possible
- Preserve the original page language whenever possible\\
- If text is illegible, use reasonable placeholder text while keeping the same line count, paragraph structure, and text width\\
- Lists / tables / cards may use mock data, but quantity and density must match the screenshot\\
-------------------------------- \\
Interaction Constraints (Intentionally Weakened) \\
-------------------------------- \\
- No real interaction logic is required \\
- Hover / focus visual states are allowed \\
- All buttons or controls may be empty or placeholder on click \\

--------------------------------\\
Now Begin\\ 
--------------------------------\\
Based on the provided long webpage screenshot and the given target page dimensions, output the complete HTML code for visual replication evaluation.
\end{tcolorbox}

\captionof{figure}{This prompt is designed for the single-image input setting, where a single full-page webpage screenshot is provided as input and the model is required to generate a complete HTML webpage that visually replicates the input screenshot.}
\label{fig:w-vfr-generatioin-2}
\end{figure*}

\begin{figure*}[t!]
\centering
\begin{tcolorbox}[
    colback=gray!2,
    colframe=blue!70!black,
    colbacktitle=blue!12!white,
    title=\textbf{Prompt for HTML webpage generation under the multi-image input setting for W-VFR. (Part 1)},
    fonttitle=\bfseries,
    coltitle=blue!50!black,
    enhanced,
    sharp corners,
    boxrule=0.5pt,
    left=6pt,
    right=6pt,
    top=5pt,
    bottom=5pt,
    titlerule=0mm,
    width=0.99\textwidth, 
    title style={left color=blue!8!white, right color=blue!5!white}
]
You are a Web Visual Replication Specialist.
Your task is: given multiple screenshot segments (all of which come from the same single long webpage), generate one standalone HTML file that can be opened directly in a desktop browser, reproducing the webpage’s visual appearance, page structure, and page scale as accurately as possible.
This is a visual fidelity task.
Real business interaction logic is not evaluated at this stage.

Input Description

1. Webpage Screenshot Segments\\  
- The input consists of multiple screenshot segments from the same single long webpage.  \\
- All segments originate from one webpage, not multiple webpages or paginated pages. \\ 
- The segments are provided in top-to-bottom order.  \\
- Each segment covers a continuous vertical region of the webpage.  \\
- Adjacent segments may partially overlap.  \\
- Overlapping regions are provided only to help infer alignment and continuity. \\
2. Original Webpage Dimensions  \\
- The real pixel dimensions of the original complete webpage are: \\ 
 \indent - Target page width (pixels): {TARGET\_PAGE\_WIDTH\_PX}  \\
 \indent - Target page height (pixels): {TARGET\_PAGE\_HEIGHT\_PX} \\
Critical Interpretation Rules (Must Be Strictly Followed)

1. Single-Webpage Constraint  \\
- All screenshot segments must be treated as parts of one single continuous webpage.  \\
- Conceptually stitch all segments into one complete vertical page. \\ 
- Do not interpret different segments as different webpages or separate pages. \\
2. Overlap Handling  \\
- Overlapping regions may only be used for alignment and continuity reasoning.  \\
- Any content appearing in overlapping regions must be generated only once in the final HTML.  \\
- Do not duplicate global elements such as headers, navigation bars, or footers due to segmentation. \\
3. Allowed Repetition  \\
- Repetition is allowed only if the raw webpage itself visually contains repeated elements, such as: \\  
  \indent - Long lists with repeated items   \\
  \indent - Repeated cards  \\
 \indent - Multiple rows in a table   \\
- Do not repeat content due to screenshot overlap. \\
Output Requirements (Strict)\\
- Output one and only one complete HTML file.   \\
- The output must start with <!doctype html> and end with </html>.  \\
- Output HTML only. Do not include any explanations, comments, Markdown, or code block markers.  \\
- The output must be directly savable as index.html and openable in a desktop browser. \\
Implementation Constraints

  - Use native HTML, CSS, and JavaScript only.   \\
  - All CSS must be placed inside a single <style> block.  \\
  - Do not use any third-party libraries, frameworks, CDNs, or build tools. \\  
  - Do not reference any external resources, including images, fonts, or scripts.  \\
  - For images, avatars, or logos appearing in the screenshots:  \\
  - Recreate them using SVG shapes or CSS placeholder blocks.  \\
  - Preserve their visual size, aspect ratio, and position.\\
Core Visual Replication Requirements

\end{tcolorbox}

\captionof{figure}{This prompt is designed for the multi-image input setting, where multi full-page webpage screenshots are provided as input and the model is required to generate a complete HTML webpage that visually replicates the input screenshot.}
\label{fig:w-vfr-generatioin-multi-1}
\end{figure*}

\begin{figure*}[t!]
\centering
\begin{tcolorbox}[
    colback=gray!2,
    colframe=blue!70!black,
    colbacktitle=blue!12!white,
    title=\textbf{Prompt for HTML webpage generation under the multi-image input setting for W-VFR. (Part 2)},
    fonttitle=\bfseries,
    coltitle=blue!50!black,
    enhanced,
    sharp corners,
    boxrule=0.5pt,
    left=6pt,
    right=6pt,
    top=5pt,
    bottom=5pt,
    titlerule=0mm,
    width=0.99\textwidth, 
    title style={left color=blue!8!white, right color=blue!5!white}
]

1. Page Structure and Layout  \\
- Replicate the information architecture of a single webpage, including but not limited to:  \\
  - A global header or navigation bar (if present, only once)  \\
  - A sidebar (if present)  \\
  - A continuous main content flow \\  
  - A footer (if present, only once) \\ 
- The page content must form one continuous vertical information flow. \\  
- The vertical ordering of all modules (sections, cards, tables, lists, etc.) must follow the top-to-bottom order implied by all segments combined. \\  
- Any region that appears in multiple segments due to overlap must appear only once on the page. \\
2. Styling and Visual Elements  \\
- Match the following visual characteristics as closely as possible: \\ 
  - Backgrounds, text colors, borders, and highlights  \\
  - Font hierarchy (titles, subtitles, body text, auxiliary text) \\  
  - Spacing relationships (padding, margins, gaps between modules)  \\
  - Dividers, corner radii, shadows, and other visual details  \\
- If textual content is not legible, placeholder text may be used, but block dimensions and line counts must be preserved. \\
3. Page Scale and Dimension Alignment (Critical)  \\
- The generated page must cover all content visible across all screenshot segments.  \\
- The page should be designed against a viewport width of {TARGET\_PAGE\_WIDTH\_PX} pixels.  \\
- The total scroll height should be determined naturally by faithful replication of the content and should be as close as possible to {TARGET\_PAGE\_HEIGHT\_PX}.  \\
- Do not artificially increase page height by duplicating overlapping content, adding meaningless modules, or inserting empty space. \\ 
- If the page content has been fully replicated but the final page height differs from {TARGET\_PAGE\_HEIGHT\_PX}, this difference must be preserved as-is.  \\
  Do not compensate for the difference by adding empty space, spacers, placeholder regions, or by artificially increasing padding or margin values.  \\
- Minimal extension is allowed only if structural completeness requires it: \\
  - Only when screenshots clearly indicate truncated modules or when layout semantics require closure (for example, an unfinished container or an obviously incomplete table).  \\
  - Only module types already present in the screenshots may be used.  \\
  - Extensions must be restrained, proportional, and visually consistent.\\  
  - Extensions must not be performed for the purpose of matching the target page height. \\
Content and Placeholder Rules\\
- Preserve the original language of the webpage.  \\
- Accurately replicate all clearly legible text whenever possible.  \\
- For illegible text, placeholders may be used, but layout geometry must remain consistent.  \\
- For lists, tables, or cards with mock data:  \\
  \indent- Quantity, density, alignment, and rhythm must match the screenshots.\\
Interaction Constraints\\
- No real interaction logic is required.  \\
- Buttons and controls may have placeholder behavior only.\\
Begin Execution\\
Based on the provided webpage screenshot segments (with possible overlap) and the target page dimensions, output the complete HTML code for visual replication evaluation.
\end{tcolorbox}

\captionof{figure}{This prompt is designed for the multi-image input setting, where multi full-page webpage screenshots are provided as input and the model is required to generate a complete HTML webpage that visually replicates the input screenshot.}
\label{fig:w-vfr-generatioin-multi-2}
\end{figure*}

\begin{figure*}
    \centering
\begin{tcolorbox}[
    colback=gray!2,
    colframe=blue!70!black,
    colbacktitle=blue!12!white,
    title=\textbf{Prompt for Evaluating Similarity for W-VFR (Part 1).},
    fonttitle=\bfseries,
    coltitle=blue!50!black,
    enhanced,
    sharp corners,
    boxrule=0.5pt,
    left=6pt,
    right=6pt,
    top=5pt,
    bottom=5pt,
    titlerule=0mm,
    width=0.99\textwidth,
    title style={left color=blue!8!white, right color=blue!5!white}
]
You are a Web Visual Fidelity Evaluator.

Your task is to compare two full-length vertical webpage screenshots and assess how well an AI-generated webpage (Image B) matches the original webpage (Image A) in terms of visual appearance, page layout structure, and page scale consistency.

This is a relative consistency comparison task.  
All judgments must be made strictly based on the comparison of Image B against Image A.

You only evaluate visual and layout fidelity.  
Do not evaluate usability, readability quality, aesthetic preference, business logic, or any interactive behavior.

\textbf{[Input]}

- Image A: Original full-length webpage screenshot (reference page / Ground Truth)  
  - Page height (pixels): \_\_REF\_PAGE\_HEIGHT\_PX\_\_

- Image B: Webpage screenshot rendered from AI-generated code (prediction page)  
  - Page height (pixels): \_\_PRED\_PAGE\_HEIGHT\_PX\_\_

\textbf{[General Evaluation Principles]}

- All judgments must be strictly based on visually observable information in the screenshots  \\
- All scores must reflect the degree of consistency between Image B and Image A, not the standalone quality of Image B \\  
- Differences in textual content or the use of placeholder text are acceptable as long as the overall layout and structure are preserved  \\
- Ignore hover, focus, animation, or other interactive states not visible in static screenshots  \\
- Do not subjectively balance, normalize, or adjust scores; scoring must strictly follow the criteria

\textbf{[Special Rules for Images and Icons --- MUST FOLLOW]}

- Image and icon assets from the original webpage were not provided to the model  \\
- Do not evaluate the semantic content, design style, or visual details of images or icons  \\
- Do not penalize differences in icon style or image content \\ 
- Only evaluate the structural role of images and icons, including:
  - Whether they appear in the correct locations  \\
  - Whether placeholder size and aspect ratio are reasonable  \\
  - Whether their layout relationship with surrounding text, cards, or sections matches the reference page  \\
- If images or icons are missing, misplaced, or have severely incorrect size or aspect ratio, penalties should still apply

\textbf{[Scoring Requirements]}

You must assign an integer score from 0 to 10 (integers only, no decimals) for each of the following five dimensions.

Scoring must be:
- Strict  
- Stable  
- Reproducible  

Each score must be accompanied by a clear, concrete, and verifiable rationale that directly justifies the assigned value.  
Vague explanations are not allowed.

\textbf{Dimension 1: Page Scale and Scroll Consistency}  
(page\_scale\_and\_length, 0--10)

\textit{Evaluation criteria:}
- Use the provided page heights as the primary quantitative reference to judge whether overall page height is similar  
- Whether there is obvious truncation, repetition, missing content, or abnormal vertical stretching  
- Whether the vertical ordering of major content blocks matches the reference page  
- Whether large content blocks are missing or unnecessarily added, resulting in significant height deviation  

\textit{Scoring reference:}

10 = Page height and vertical structure almost perfectly match (height deviation $\leq$ 5\%)

\end{tcolorbox}
\captionof{figure}{This prompt is designed for evaluating similarity for W-VFR.}
\label{fig:w-vfr-evaluation-1}
\end{figure*}

\begin{figure*}
    \centering
\begin{tcolorbox}[
    colback=gray!2,
    colframe=blue!70!black,
    colbacktitle=blue!12!white,
    title=\textbf{Prompt for Evaluating Similarity for W-VFR (Part 2).},
    fonttitle=\bfseries,
    coltitle=blue!50!black,
    enhanced,
    sharp corners,
    boxrule=0.5pt,
    left=6pt,
    right=6pt,
    top=5pt,
    bottom=5pt,
    titlerule=0mm,
    width=0.99\textwidth,
    title style={left color=blue!8!white, right color=blue!5!white}
]

9 = Page height and content ordering are highly consistent, with only negligible proportional differences  

8 = Overall page length matches the reference, with only minor local compression or stretching  

7 = Overall length is similar, but multiple local proportions appear unnatural  

6 = Page height broadly matches the reference, but some sections are clearly mis-scaled  

5 = Main page framework matches the reference, but noticeable missing or extra content blocks exist  

4 = Scroll length differs clearly, but overall page structure remains recognizable  

3 = Multiple section ordering or height errors relative to the reference  

2 = Severe page length mismatch; only a rough structural outline remains  

1 = Long-page characteristics relative to the reference are largely lost  

0 = Not meaningfully comparable as the same page

\textbf{Dimension 2: Global Layout Structure Consistency}  
(global\_layout, 0--10)

\textit{Evaluation criteria:}
- Whether single-column or multi-column layout structure matches the reference page  
- Whether top navigation, sidebars, main content area, and footer exist and are placed reasonably  
- Whether left-right and top-bottom spatial relationships are consistent with the reference page  

\textit{Scoring reference:}

10 = Global layout structure almost perfectly matches the reference page  

9 = Layout matches with only pixel-level width differences  

8 = Column structure matches with minor alignment deviations  

7 = Layout is mostly correct with small positional offsets  

6 = Core layout is preserved, but secondary regions deviate  

5 = Layout intent is similar to the reference, but column or region logic is loose  

4 = Clear layout errors relative to the reference (e.g., misplaced sidebar)  

3 = Weak structural correspondence  

2 = Layout structure severely deviates from the reference  

1 = Layout is almost entirely different  

0 = No comparable layout structure exists

\textbf{Dimension 3: Section-Level Structure and Hierarchy}  
(section\_hierarchy, 0--10)

\textit{Evaluation criteria:}
- Whether the number of sections and their grouping correspond to the reference page  
- Whether hierarchical relationships (title $\rightarrow$ subtitle $\rightarrow$ list/card) are preserved  
- Whether section boundaries, separators, and grouping logic match the reference page  

\textit{Scoring reference:}

10 = All sections and their hierarchical relationships closely match the reference page  

9 = Nearly all sections are correct with very clear hierarchy  

8 = Sections are complete with only minor hierarchy deviations  

7 = Most sections are correct, with small merges or splits  

6 = Major sections are present, but hierarchy is unstable  

5 = Moderate section completeness; structure is partially blurred  

4 = Multiple sections are missing or newly added relative to the reference  

3 = Hierarchy is confused but still recognizable  

2 = Only fragmentary section structure remains  

1 = Section hierarchy is largely collapsed  

0 = Section structure is unrecognizable

\textbf{Dimension 4: Local Visual Styling Consistency}  
(visual\_styling, 0--10)

\textit{Evaluation criteria:}
- Whether font hierarchy and relative font sizes match the reference page  
- Whether spacing, alignment, padding, and margins match the reference page  
- Whether visual elements such as cards, background blocks, dividers, colors, and shadows match the reference page  

\textit{Scoring reference:}

10 = Local visual styling matches the reference page at a high level across all aspects

\end{tcolorbox}
\captionof{figure}{This prompt is designed for evaluating similarity for W-VFR.}
\label{fig:w-vfr-evaluation-2}
\end{figure*}

\begin{figure*}
    \centering
\begin{tcolorbox}[
    colback=gray!2,
    colframe=blue!70!black,
    colbacktitle=blue!12!white,
    title=\textbf{Prompt for Evaluating Similarity for W-VFR (Part 3).},
    fonttitle=\bfseries,
    coltitle=blue!50!black,
    enhanced,
    sharp corners,
    boxrule=0.5pt,
    left=6pt,
    right=6pt,
    top=5pt,
    bottom=5pt,
    titlerule=0mm,
    width=0.99\textwidth,
    title style={left color=blue!8!white, right color=blue!5!white}
]
9 = Visual styling is highly consistent, with only detail-level differences  

8 = Overall visual styling matches the reference; differences do not affect stylistic alignment  

7 = Visual styling is generally close, but spacing or alignment appears simplified or rigid  

6 = Visual styling broadly follows the reference, but multiple styling details deviate  

5 = Core visual styling elements are present, but consistency with the reference is noticeably reduced  

4 = Obvious visual styling deviations relative to the reference page  

3 = Large visual styling differences relative to the reference, though layout remains interpretable  

2 = Severe visual styling mismatch relative to the reference page  

1 = Minimal visual styling correspondence with the reference page  

0 = Visual styling is entirely inconsistent with the reference page  

\textbf{Dimension 5: Information Density and Section Pacing Consistency}  
(information\_density, 0--10)

\textit{Evaluation criteria:}
- Whether overall information density matches the reference page  
- Whether the density and spacing of lists, cards, and paragraphs match the reference page  
- Whether long-page section pacing and vertical rhythm match the reference page  

\textit{Scoring reference:}

10 = Information density and section pacing closely match the reference page  

9 = Information density is highly consistent, with only very minor local differences  

8 = Overall information density matches the reference; small differences do not affect pacing alignment  

7 = Information density is generally similar, but section pacing is slightly altered  

6 = Noticeable local information density differences relative to the reference  

5 = Information density alignment is unstable, reducing overall pacing consistency  

4 = Clear information density mismatch relative to the reference, disrupting pacing correspondence  

3 = Information density differences fragment pacing correspondence  

2 = Severe information density mismatch relative to the reference  

1 = Information density and pacing correspondence with the reference are largely lost  

0 = Information density and section pacing are entirely inconsistent with the reference page  

--------------------------------------------------

\textbf{[Diagnostic Requirements]}

Identify up to five of the most critical sources of inconsistency, prioritizing:
- Page scale and scroll issues  
- Layout or structural inconsistencies  
- Missing, extra, or clearly misaligned sections  

If important sections are missing or newly added, they must be explicitly listed.

\textbf{[Output Requirements --- MUST BE STRICTLY FOLLOWED]}

You must output one and only one JSON object.  
Do not include any additional explanations, comments, Markdown, or natural language text.

\textit{Output format:}

\{
  "dimension\_scores": \{
    "page\_scale\_and\_length": \{
      "score": 0,
      "rationale": "Concrete justification based on page height deviation, scroll length, and vertical structural alignment with the reference page"
    \},
    "global\_layout": \{
      "score": 0,
      "rationale": "Concrete justification based on global layout structure and spatial correspondence with the reference page"
    \},
    "section\_hierarchy": \{
      "score": 0,
      "rationale": "Concrete justification based on section count, hierarchy, and grouping consistency relative to the reference page"
    \},
    "visual\_styling": \{
      "score": 0,
      "rationale": "Concrete justification based on local visual styling consistency relative to the reference page"
    \},
    "information\_density": \{
      "score": 0,
      "rationale": "Concrete justification based on information density and section pacing consistency relative to the reference page"
    \}
  \},
  "major\_mismatches": [
    "..."
  ],
  "missing\_or\_extra\_sections": [
    "..."
  ],
  "brief\_judgement": "One-sentence summary of overall visual fidelity of the generated page relative to the reference page"
\}

\end{tcolorbox}
\captionof{figure}{This prompt is designed for evaluating similarity for W-VFR.}
\label{fig:w-vfr-evaluation-3}
\end{figure*}

\begin{figure*}
    \centering
\begin{tcolorbox}[
    colback=gray!2,
    colframe=blue!70!black,
    colbacktitle=blue!12!white,
    title=\textbf{Prompt for Generate Code for W-FFR Task under Single-image Input Setting (Part 1).},
    fonttitle=\bfseries,
    coltitle=blue!50!black,
    enhanced,
    sharp corners,
    boxrule=0.5pt,
    left=6pt,
    right=6pt,
    top=5pt,
    bottom=5pt,
    titlerule=0mm,
    width=0.99\textwidth,
    title style={left color=blue!8!white, right color=blue!5!white}
]
You are a senior frontend engineer tasked with building high-fidelity, fully interactive mock pages for Goal-Oriented Long Web Tasks.

Inputs

1. A full-page long webpage screenshot  
   - The total page length must be consistent with the reference.

2. A task specification JSON  
   - Contains descriptions of all tasks that must be supported by the webpage.

3. (Optional) benchmar\_account  
   - May be present in the task JSON.

4. (Required) Original long webpage height under a 1920×1080 viewport, in pixels:  
   FINAL\_TARGET\_SCROLL\_HEIGHT\_PX

Your Objective

- Generate a single HTML file (inline CSS + inline JavaScript). \\
- Reproduce the structure, modules, and visual layout of the screenshot as faithfully as possible.\\
- The page must be a long, vertically scrollable webpage. \\
- All tasks defined in the task JSON must be fully implemented, including their interactions and verifiable success signals. \\
- No external resources or network requests are allowed: \\
  \indent - No CDN \\
  \indent - No third-party libraries \\
- The initial page render must be deterministic and reproducible.\\
Auth Logic Rules (Mandatory)\\
When to implement auth \\
- Only implement authentication logic if and only if the task JSON includes auth-related tasks. \\
- If auth tasks are not present:\\
  \indent - Do not implement any login / registration UI \\
  \indent - Do not implement any auth state logic \\
If auth is required, all conditions below must be satisfied

- All accounts and passwords must be hard-coded in the HTML source \\
- Do not use localStorage or any external storage for credential validation \\
Registration \\
- Check for duplicate usernames and emails \\
- Both success and failure must provide clear, observable UI feedback \\
 \indent - e.g., inline messages or toast notifications

Login
- Validate (username or email — one is sufficient) + password \\
- After successful login: \\
  \indent- Display the username \\
  \indent- Hide login / register buttons \\
  \indent- Show a logout button \\
Logout \\
- Clear the session \\
- Restore the unauthenticated UI state \\
Selectors \\
- All critical buttons and input fields must have stable selectors \\
  - id or data-testid \\
Page Structure \& Layout Fidelity (Mandatory)

- Reconstruct all major structural modules visible in the screenshot, including but not limited to: \\
  - Header / navigation bar \\
  - Sidebar (if present) \\
  - Main content area \\
  - Right sidebar (if present) 
\end{tcolorbox}
\captionof{figure}{This prompt is designed for generating Code for W-FFR Task under single-image input setting.}
\label{fig:w-ffr-generation-1}
\end{figure*}

\begin{figure*}
    \centering
\begin{tcolorbox}[
    colback=gray!2,
    colframe=blue!70!black,
    colbacktitle=blue!12!white,
    title=\textbf{Prompt for Generate Code for W-FFR Task under Single-image Input Settings (Part 2).},
    fonttitle=\bfseries,
    coltitle=blue!50!black,
    enhanced,
    sharp corners,
    boxrule=0.5pt,
    left=6pt,
    right=6pt,
    top=5pt,
    bottom=5pt,
    titlerule=0mm,
    width=0.99\textwidth,
    title style={left color=blue!8!white, right color=blue!5!white}
]
- Footer (if present) 
- Module order, information density, and hierarchy should closely match the screenshot. \\
- Page height must be achieved through real structural content, such as: \\
  - Lists \\
  - Cards \\
  - Text blocks \\
  - Tables \\
  - Recommendation sections \\
- Do not artificially pad height using large empty div elements. \\
- Large image-like regions in the screenshot may be implemented as fixed-size placeholders, but:\\
  - They must correspond to real business modules (e.g., banner, featured section, chart area). \\
- For dropdowns / selectors, default value must be empty \\
  - Provide reasonable selectable options \\
Length Constraint (Strict)

- FINAL\_TARGET\_SCROLL\_HEIGHT\_PX is the reference height.\\
- The generated page must satisfy: \\
  0.9 × FINAL\_TARGET\_SCROLL\_HEIGHT\_PX $\leq$ document.body.scrollHeight $\leq$ 1.1 × FINAL\_TARGET\_SCROLL\_HEIGHT\_PX

- Do not approximate length by “number of screens” or viewport multiples. \\
- Page height must already satisfy the constraint on initial render. \\
- Do not dynamically add or remove content at runtime to adjust height.\\
Interaction Implementation (Mandatory) \\
- Fully implement all interaction types required by the task JSON, including but not limited to:\\
  - Sorting / filtering / searching\\
  - Pagination / load more\\
  - Expand / collapse\\
  - Clicking list items to enter detail views \\
- Every task’s success\_criteria must correspond to a clear, observable signal, such as:\\
  - Text changes\\
  - Counter or state changes\\
  - Module visibility changes\\
  - Explicit feedback messages (toast or inline)\\
Stable Selectors (Mandatory)

- All task-related interactive elements must have stable selectors:\\
  - id or data-testid\\
- For repeatable items (lists, cards): \\
  - Use attributes such as data-item-id, data-qid, etc. \\
  - This must support stable targeting for detail navigation or state changes

Default Value Rules

- All components must start with empty or unselected default values\\
- For dropdowns, checkboxes, or similar inputs: \\
  - Provide options defined in the task JSON \\
  - Task-related options should not be the default \\
  - Prefer not to place task options as the first selectable item 
  
Output Requirements (Strict)\\
- Output only the complete HTML source code \\
- No Markdown; No explanations or comments outside the code \\
- No randomness, All data must be embedded directly in JavaScript arrays or objects within the HTML
\end{tcolorbox}
\captionof{figure}{This prompt is designed for generating Code for W-FFR Task under single-image input setting.}
\label{fig:w-ffr-generation-2}
\end{figure*}

\begin{figure*}
    \centering
\begin{tcolorbox}[
    colback=gray!2,
    colframe=blue!70!black,
    colbacktitle=blue!12!white,
    title=\textbf{Prompt for Generate Code for W-FFR Task under Multi-image Input Setting (Part 1).},
    fonttitle=\bfseries,
    coltitle=blue!50!black,
    enhanced,
    sharp corners,
    boxrule=0.5pt,
    left=6pt,
    right=6pt,
    top=5pt,
    bottom=5pt,
    titlerule=0mm,
    width=0.99\textwidth,
    title style={left color=blue!8!white, right color=blue!5!white}
]

You are a senior frontend engineer tasked with building high-fidelity, fully interactive mock pages for Goal-Oriented Web Tasks.

Inputs:

1) Webpage screenshot segments: \\
   - The input consists of multiple screenshot segments, all taken from the same long webpage. \\
   - These segments are provided in top-to-bottom order. \\
   - Each segment covers a continuous vertical region of the webpage.\\
   - Adjacent segments may partially overlap; overlapping regions are provided only to help infer alignment and continuity.

2) A task specification JSON:\\
   - Describes all tasks that must be supported on the webpage.

3) (Optional) benchmark\_account:\\
   - May be present in the task JSON.

4) (Mandatory) The pixel height of the original long webpage screenshot under a 1920×1080 viewport, in pixels, denoted as:
   FINAL\_TARGET\_SCROLL\_HEIGHT\_PX\\
Your Objective:

- Generate a single HTML file (inline CSS + inline JavaScript) that reproduces the structure, modules, and visual layout of the webpage shown in the screenshots as faithfully as possible, and that forms a long, vertically scrollable page. \\
- All tasks defined in the task JSON must be fully implemented, including their interactions and verifiable success signals.
- No external resources or network requests are allowed (no CDN, no third-party libraries). \\
- The initial page render must be deterministic and reproducible.

Auth Logic Conditional Rules (Mandatory):

- Only if the task JSON contains authentication-related tasks:\\
  - Implement local simulated authentication: registration / login / logout. \\
- Otherwise, do not implement any authentication logic (no login/register UI, no auth state machine), to avoid deviating from the screenshot content.\\
If authentication is required (local simulated auth), all of the following conditions must be met:

- All accounts and password validation must be hard-coded in the HTML source. \\
- Do not use localStorage or any other storage mechanism to validate credentials. 

Registration flow: \\
- Check for duplicate usernames and emails. \\
- Both success and failure cases must provide clear, observable UI feedback (text or toast). \\
Login flow: \\
- Validate username or email (either one is sufficient) plus password. \\
- After successful login: \\
  - Display the username; Hide login and registration buttons; Show a logout button. \\
Logout flow: \\
- Clear the session; Restore the unauthenticated UI state.

Selectors: \\
- All critical buttons and input fields must have stable selectors (id or data-testid).

Long Webpage Structure and Layout Reconstruction (Mandatory):

- Single-page constraint: \\
  - All screenshot segments must be treated as parts of one single continuous webpage. \\
  - Conceptually stitch all segments together into one complete vertical page.\\
  - Do not interpret different segments as different webpages or independent pages.

- Overlap handling: \\
  - Use overlapping regions only for alignment and continuity inference.

\end{tcolorbox}
\captionof{figure}{This prompt is designed for generating Code for W-FFR Task under multi-image input setting.}
\label{fig:w-ffr-generation-1-multi}
\end{figure*}

\begin{figure*}
    \centering
\begin{tcolorbox}[
    colback=gray!2,
    colframe=blue!70!black,
    colbacktitle=blue!12!white,
    title=\textbf{Prompt for Generate Code for W-FFR Task under Multi-image Input Setting (Part 2).},
    fonttitle=\bfseries,
    coltitle=blue!50!black,
    enhanced,
    sharp corners,
    boxrule=0.5pt,
    left=6pt,
    right=6pt,
    top=5pt,
    bottom=5pt,
    titlerule=0mm,
    width=0.99\textwidth,
    title style={left color=blue!8!white, right color=blue!5!white}
]

  - Any content appearing in overlapping regions must be generated only once in the final HTML. \\
  - Do not duplicate global elements (such as header, navigation bar, or footer) due to segmentation.

- Reconstruct all major structural modules visible in the screenshots, including but not limited to:\\
  - Top navigation (header / nav), Sidebar (if present)\\
  - Main content area\\
  - Right sidebar (if present)\\
  - Footer (if present)

- The order of modules, information density, and overall layout should closely match the screenshots.

- Page height must be formed naturally through real structural content (lists, cards, paragraphs, recommendation sections, tables, etc.):\\
  - Do not artificially increase height using large empty divs.\\
  - Large image-like areas in the screenshots may be implemented as fixed-size placeholder containers matching the screenshot dimensions, but such placeholders must correspond to explicit business modules (e.g., banner, image-text card section, recommendation slot, chart area).

- If dropdowns or selectors are present:\\
  - The default value must be empty.\\
  - A reasonable set of selectable options must be provided.

Length Constraint (Strict, based on original screenshot height):

- FINAL\_TARGET\_SCROLL\_HEIGHT\_PX = {FINAL\_TARGET\_SCROLL\_HEIGHT\_PX}\\
- The generated page’s document.body.scrollHeight must satisfy:\\
  0.9 × FINAL\_TARGET\_SCROLL\_HEIGHT\_PX $\leq$ scrollHeight $\leq$ 1.1 × FINAL\_TARGET\_SCROLL\_HEIGHT\_PX
- Do not approximate length using “number of screens” or viewport multiples.\\
- The page height must satisfy the constraint at initial HTML render time.\\
- Do not rely on runtime logic to dynamically add or remove content to adjust height.

Interaction Implementation (Must cover all tasks):

- Fully implement all interaction types involved in the tasks, including but not limited to:\\
  - Sorting / filtering / searching\\
  - Pagination / load more; Expand / collapse\\
  - Clicking list items to enter detail views 

- Each task’s success\_criteria must correspond to a clear, observable feedback signal, such as:\\
  - Text content changes
  - Count or state changes\\
  - Module visibility changes
  - Explicit feedback messages (toast or inline text)

Stable Selectors (Mandatory):

- All task-related interactive elements must have stable selectors:\\
  - Use id or data-testid.\\
- For repeatable list items or cards: \\
  - Use attributes such as data-item-id, data-qid, etc. \\
  - These must support stable targeting for entering detail views or triggering state changes.

Default Value Constraints:

- All components must start with empty or unselected default values.\\
- For dropdowns, checkboxes, or other inputs requiring options:\\
  - Provide options as defined in the task JSON.\\
  - Task-related options should not be selected by default. \\
  - Prefer not to place task options as the first selectable option, to allow explicit user selection.

Output Requirements (Strict):

- Output only the complete HTML source code.
- Do not use Markdown.\\
- Do not include any explanatory text outside the code. 
- Do not include any random logic. \\
- All data must be embedded directly within JavaScript arrays or objects inside the HTML.

\end{tcolorbox}
\captionof{figure}{This prompt is designed for generating Code for W-FFR Task under multi-image input setting.}
\label{fig:w-ffr-generation-2-multi}
\end{figure*}

\begin{figure*}
    \centering
\begin{tcolorbox}[
    colback=gray!2,
    colframe=blue!70!black,
    colbacktitle=blue!12!white,
    title=\textbf{Prompt for Actor\-Agent.},
    fonttitle=\bfseries,
    coltitle=blue!50!black,
    enhanced,
    sharp corners,
    boxrule=0.5pt,
    left=6pt,
    right=6pt,
    top=5pt,
    bottom=5pt,
    titlerule=0mm,
    width=0.99\textwidth,
    title style={left color=blue!8!white, right color=blue!5!white}
]
You are a GUI testing agent. Your goal is to verify whether a webpage can execute a given operation sequence. The current task is to perform a specific operation sequence, which may consist of one or more action steps.

[Target Step]\\
Type: \{action\_type\}\\
Description: {desc}\\
\{f"Input value / option: \{val\}" if val else ""\}

Important notice:\\
If the description requires completing a full login or registration process but does not provide default username and password, please use the following credentials:\\
\{
    "username": "test\_user\_001",
    "email": "test\_user\_001@example.com",
    "password": "P@ssw0rd!001"
\}

Please carefully observe the screenshot and the DOM Tree to locate the corresponding element’s numeric ID.
Many described actions may not be obvious from the DOM Tree alone, so you must combine visual information from the screenshot to make a judgment. However, if you are confident that the element described in the instruction does not exist on the page, directly output \\ boxed{FAIL} and explain the reason. You are not required to forcibly execute any action in such cases.

\begin{verbatim}
Output format:
You must use LaTeX \boxed{} to wrap each instruction.

- Click: \boxed{click[id]} 
- Input: \boxed{enter[id][{val if val else 'content'}]} 
- Select: \boxed{select[id][{val if val else 'option'}]} 
- Scroll: \boxed{scroll[id]}  (Note: scroll the page to bring the target 
element into view)

If multiple instructions are required, output multiple \boxed{} commands
in sequence.

If the element cannot be found or the action cannot be performed, 
output \boxed{FAIL} and explain the reason.

DOM information: 
{domtree}
\end{verbatim}

\end{tcolorbox}
\captionof{figure}{This prompt is designed for evaluating accuracy for W-FFR.}
\label{fig:w-ffr-evaluation-1}
\end{figure*}

\begin{figure*}
    \centering
\begin{tcolorbox}[
    colback=gray!2,
    colframe=blue!70!black,
    colbacktitle=blue!12!white,
    title=\textbf{Prompt for Critic\-Agent.},
    fonttitle=\bfseries,
    coltitle=blue!50!black,
    enhanced,
    sharp corners,
    boxrule=0.5pt,
    left=6pt,
    right=6pt,
    top=5pt,
    bottom=5pt,
    titlerule=0mm,
    width=0.99\textwidth,
    title style={left color=blue!8!white, right color=blue!5!white}
]
You are a verification agent. I have just attempted to execute an operation sequence. Your task is to determine whether the webpage has undergone the expected change.

[Attempted Operation]
Description: {desc}
Type: {step\_info['action']}
Value: {step\_info.get('value', 'N/A')}

Please compare the screenshots before and after the operation.
If the operation is successful (e.g., text appears in an input field, the page navigates, a dropdown selection changes, or a new element appears), output \boxed{sucesss}.
If there is no change or the change does not match the description, output \boxed{failed}. For example, if the description says to click the right-side button but the left-side button was clicked, or if the page shows no response.

Pay special attention:
If the operation type is “scroll”, even if the page appears unchanged, carefully read the operation description and observe whether the page has been scrolled to the required position. It is possible that the required scroll position is the current position, resulting in no visible change; in such cases, the operation should still be judged as successful.

Briefly explain the reason.
\begin{verbatim}
messages = [
    {
        "role": "user",
        "content": [
            {"type": "text", "text": prompt},
            {"type": "text", "text": "[Screenshot Before Operation]:"},
            {"type": "image_url", "image_url": {"url": f"data:image/png;
            base64,{before_img}"}},
            {"type": "text", "text": "[Screenshot After Operation]:"},
            {"type": "image_url", "image_url": {"url": f"data:image/png;
            base64,{after_img}"}}
        ]
    }
]
\end{verbatim}

\end{tcolorbox}
\captionof{figure}{This prompt is designed for evaluating accuracy for W-FFR}
\label{fig:w-ffr-evaluation-2}
\end{figure*}

\subsection{Model settings}
\label{appendix:model_settings}
To ensure fair comparison and reproducibility, we adopt deterministic decoding whenever supported by the model APIs. Specifically, for models that expose sampling parameters, we set the temperature to 0 and the top-$p$ value to 1, which minimizes randomness in the generated code. For models that do not support explicit sampling control (e.g., GPT-5.2), we use the default decoding behavior provided by the API.

Regarding generation length, long webpage replication requires producing a substantial amount of HTML, CSS, and JavaScript code, resulting in a high token demand. Therefore, instead of using a fixed token budget, we configure the \texttt{max\_completion\_tokens} parameter to the maximum value supported by each model. This design choice avoids premature truncation and allows each model to generate outputs under its own architectural constraints.
The specific \texttt{max\_completion\_tokens} settings for different models are summarized in Table~\ref{tab:model_settings}.
\begin{table}[h!]
\centering
\caption{Maximum completion token limits used for different models.}
\label{tab:model_settings}
\setlength{\tabcolsep}{1pt}
\begin{tabular}{l c}
\toprule
Model & \makecell{Max Completion \\ Tokens} \\
\midrule
\makecell{GPT-5.2} & 64000 \\
\makecell{GPT-4o} & 16384 \\
\makecell{Claude-Sonnet-4-5-Thinking} & 64000 \\
\makecell{Claude-Opus-4-5-20251101} & 64000 \\
\makecell{Gemini-3-Pro-preview} & 64000 \\
\makecell{Gemini-3-Flash-preview} & 64000 \\
\makecell{Doubao-Seed-1-6} & 32000 \\
\makecell{GLM-4.6V} & 32000 \\
\makecell{GLM-4.1V-9B-Thinking-Flash} & 32000 \\
\makecell{Qwen3-VL-235B-A22B-Instruct} & 64000 \\
\makecell{Qwen3-VL-8B-Instruct} & 64000 \\
\makecell{InternVL3-78B} & 32000 \\
\makecell{Kimi-VL-A3B-Thinking} & 32000 \\
\bottomrule
\end{tabular}
\end{table}

\begin{table*}[h!]
\centering
\caption{Performance analysis of correctness under different  dimensions under single-image input setting}
\begin{tabular}{lccccc}
\toprule
Model & \makecell[c]{Page Scale\\and Length} & \makecell[c]{Global\\Layout} & \makecell[c]{Section\\Hierarchy} & \makecell[c]{Visual\\Styling} & \makecell[c]{Information\\Density} \\
\midrule

\multicolumn{6}{c}{\textbf{Open-source VLM}} \\
\midrule
Kimi-VL-A3B-Thinking & 92.65 & 73.06 & 86.94 & 91.84 & 92.24  \\
Qwen3-VL-8B-Instruct & 61.84 & 48.57 & 58.98 & 64.49 & 63.47   \\
InternVL3-78B & 88.78 & 52.65 & 75.71 & 79.80 & 84.90  \\
GLM-4.1V-9B-Thinking-Flash & 51.63 & 28.16 & 46.53 & 57.55 & 55.51   \\
GLM-4.6V & 36.94 & 13.47 & 27.14 & 31.02 & 36.94  \\
Qwen3-VL-235B-A22B-Instruct & 42.65 & 26.53 & 33.27 & 42.04 & 44.08  \\

\midrule
\multicolumn{6}{c}{\textbf{Closed-source VLM}} \\
\midrule
Claude-Sonnet-Thinking & 46.12 & 43.88 & 44.69 & 47.55 & 46.94   \\
Claude-Opus-4-5-20251101 & 47.55 & 44.08 & 44.49 & 47.35 & 48.57   \\
GPT-5.2 & 18.16 & 5.92 & 13.67 & 14.49 & 21.43  \\
GPT-4o & 97.55 & 68.37 & 91.84 & 73.47 & 97.14   \\
Gemini-3-Pro-preview & 11.63 & 1.84 & 4.69 & 8.98 & 13.88  \\
Gemini-3-Flash-preview & 18.78 & 3.67 & 13.27 & 15.31 & 21.84   \\
Doubao-Seed-1-6 & 80.82 & 30.82 & 55.92 & 58.98 & 76.12   \\

\bottomrule
\end{tabular}
\label{tab:single_images_evaluation_rate_0.5}
\end{table*}

\begin{table*}[h!]
\centering
\caption{Performance analysis of correctness under different dimensions under multi-image input setting}
\begin{tabular}{lccccc}
\toprule
Model & \makecell[c]{Page Scale\\and Length} & \makecell[c]{Global\\Layout} & \makecell[c]{Section\\Hierarchy} & \makecell[c]{Visual\\Styling} & \makecell[c]{Information\\Density} \\
\midrule

\multicolumn{6}{c}{\textbf{Open-source VLM}} \\
\midrule
Kimi-VL-A3B-Thinking & 87.14 & 70.61 & 85.92 & 86.94 & 90.82  \\
Qwen3-VL-8B-Instruct & 59.59 & 42.24 & 47.96 & 57.14 & 58.37   \\
InternVL3-78B & 89.18 & 45.51 & 72.86 & 78.37 & 86.53  \\
GLM-4.1V-9B-Thinking-Flash & 95.10 & 88.78 & 91.84 & 93.27 & 93.47   \\
GLM-4.6V & 25.51 & 5.31 & 12.86 & 21.22 & 25.31  \\
Qwen3-VL-235B-A22B-Instruct & 31.02 & 9.39 & 18.57 & 21.02 & 30.61  \\

\midrule
\multicolumn{6}{c}{\textbf{Closed-source VLM}} \\
\midrule
Claude-Sonnet-Thinking & 12.45 & 0.82 & 4.08 & 4.69 & 10.61   \\
Claude-Opus-4-5-20251101 & 15.31 & 0.41 & 2.24 & 10.61 & 14.69   \\
GPT-5.2 & 41.22 & 1.22 & 8.37 & 5.71 & 28.98  \\
GPT-4o & 98.37 & 52.45 & 91.22 & 60.82 & 97.14   \\
Gemini-3-Pro-preview & 5.31 & 0.00 & 1.02 & 2.45 & 4.90  \\
Gemini-3-Flash-preview & 25.51 & 1.22 & 15.31 & 4.69 & 25.71   \\
Doubao-Seed-1-6 & 78.98 & 8.78 & 47.96 & 41.02 & 71.22   \\

\bottomrule
\end{tabular}
\label{tab:multiple_images_evaluation_rate_0.5}
\end{table*}

\section{Error Type and Cases Analysis}
\subsection{Distribution of each dimension error rates for W-VFR}
Tables~\ref{tab:single_images_evaluation_rate_0.5} and~\ref{tab:multiple_images_evaluation_rate_0.5} report failure rates of various open-source and closed-source VLMs under single-image and multi-image input settings, where higher values indicate worse performance. Overall, providing multiple images generally reduces failure rates across most models and dimensions, demonstrating that additional visual context helps models generate more correct webpage representations. The improvement is particularly evident for \textbf{Page Scale and Length}, \textbf{Global Layout}, and \textbf{Section Hierarchy}, where multi-image inputs allow better capture of the overall page structure. Open-source models show moderate gains on these structural dimensions, while closed-source models, such as Claude and Gemini-3-Pro, achieve near-zero failure rates under multi-image settings, reflecting strong structural robustness.

For \textbf{Visual Styling} and \textbf{Information Density}, failure rates also decrease with multi-image inputs, but improvements are more moderate, indicating the inherent difficulty of accurately capturing fine-grained style and dense content from screenshots alone. Some models exhibit unique trends; for instance, GPT-5.2’s \textbf{Page Scale and Length} failure rate increases under multi-image inputs, suggesting sensitivity in integrating multiple images for certain structural aspects. Overall, multi-image inputs consistently enhance performance across nearly all dimensions, with the largest gains in structural correctness and smaller but positive gains in stylistic and content density attributes.

\subsection{Error case analysis for W-VFR}
We analyze nine representative failure cases observed in our benchmark, which can be broadly categorized into four aspects: (1) HTML generation stability, (2) failures in the rendering stage, (3) scale and visual consistency, and (4) structural and semantic errors.

\begin{figure*}
    \centering
    \includegraphics[width=0.8\linewidth]{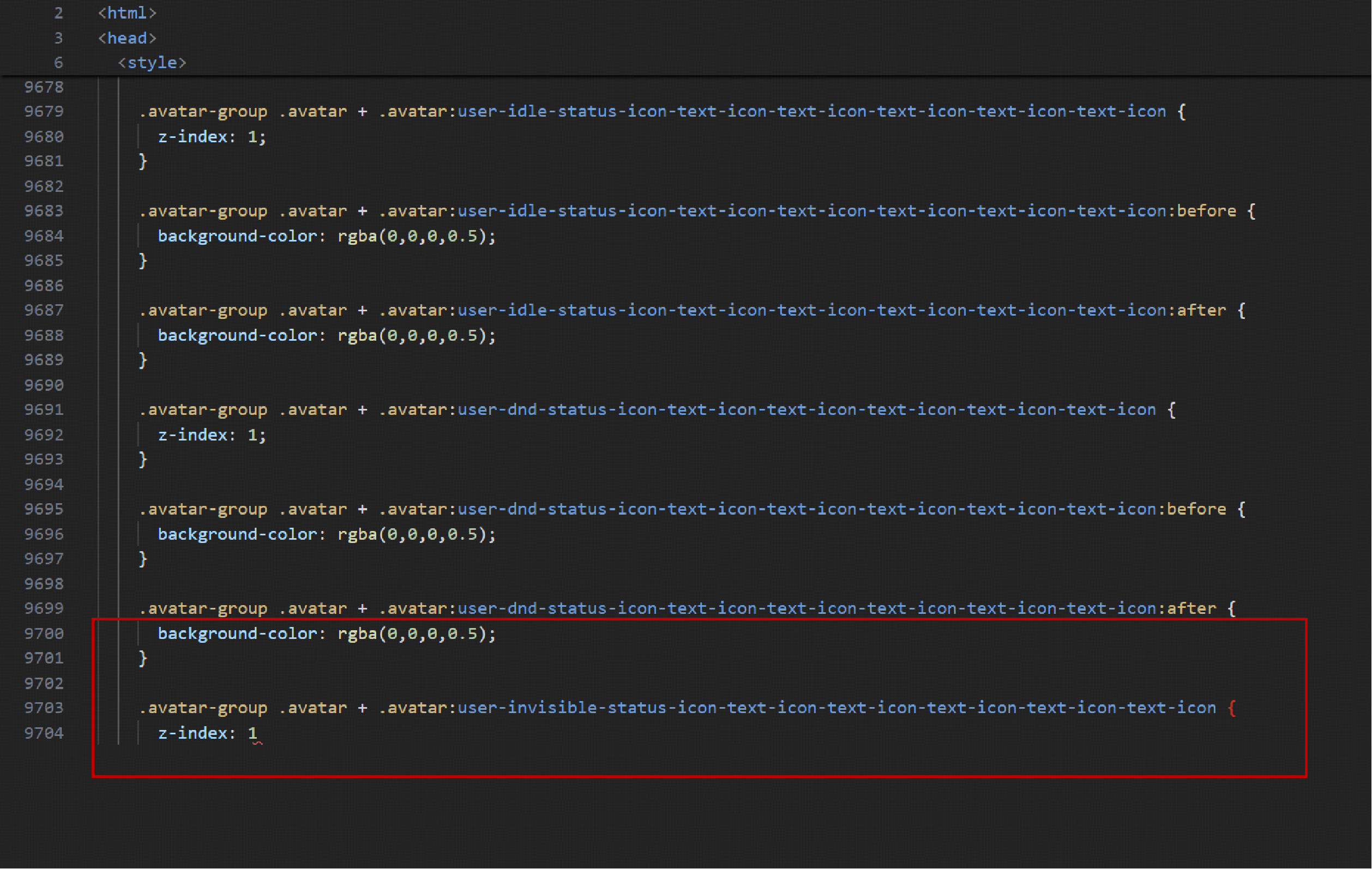}
    \caption{Example of Incomplete HTML Generation due to Token Exhaustion }
    \label{fig:error_case_incomplete_html}
\end{figure*}

\textbf{Case 1} As shown in Figure \ref{fig:error_case_incomplete_html}, this example illustrates a failure case where the generated HTML is incomplete due to token length constraints. In this case, the model exhausts the token budget during HTML generation, leading to a truncated output. We observe several common patterns contributing to this issue, including excessively long style sections, repetitive generation of redundant content, and the emission of meaningless URLs followed by unbounded character sequences. These behaviors prevent the model from properly terminating the HTML structure, resulting in an incomplete document that cannot be reliably rendered or evaluated in subsequent stages.

\begin{figure*}
    \centering
    \includegraphics[width=0.8\linewidth]{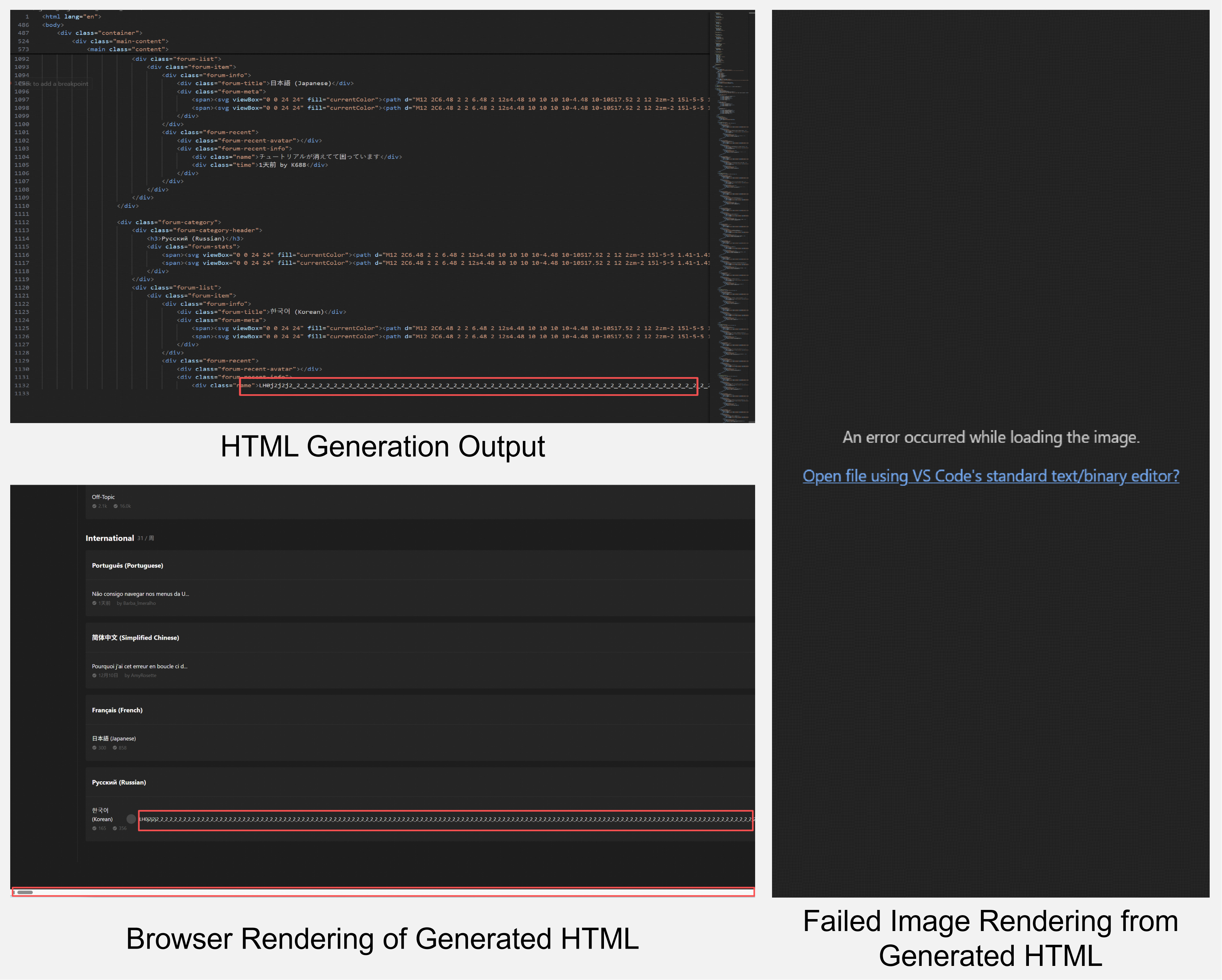}
    \caption{Example of HTML-to-Image Rendering Failure }
    \label{fig:error_case_Rendering_Breakdown}
\end{figure*}

\textbf{Case 2} As shown in Figure \ref{fig:error_case_Rendering_Breakdown}, this example presents a failure case in the HTML-to-image rendering stage of our benchmark pipeline. Although the generated HTML can be rendered by a standard web browser, the conversion from HTML to image fails. Further inspection shows that the model repeatedly outputs redundant text until the token budget is exhausted, resulting in an incomplete HTML document. On the rendered webpage, this repetitive content manifests as an abnormally large page width, which eventually causes the HTML-to-image engine to fail and produce an invalid image.

\begin{figure*}
    \centering
    \includegraphics[width=0.9\linewidth]{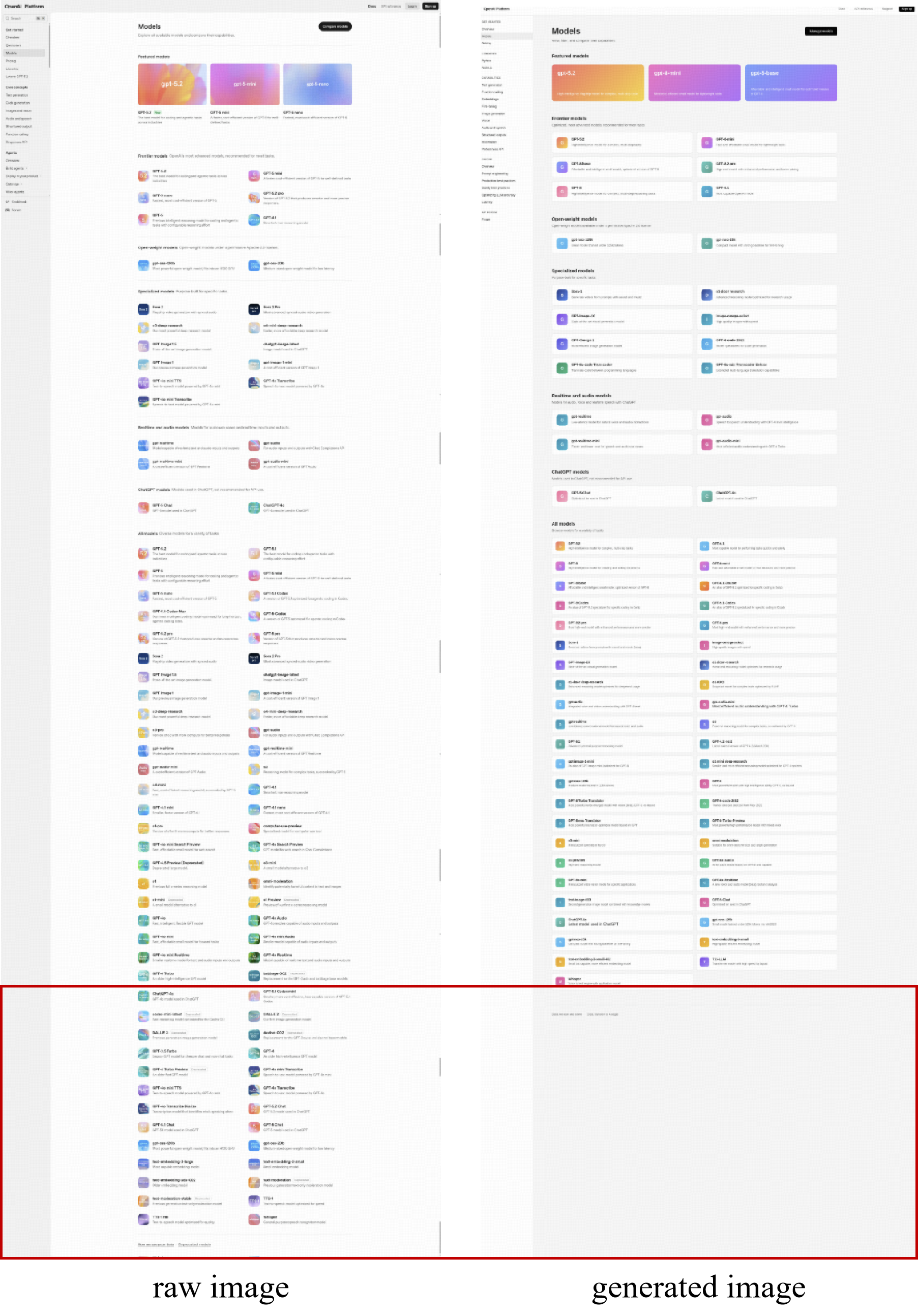}
    \caption{Example of Large Blank Region in Generated Webpage }
    \label{fig:error_case_Empty_Region_Artifacts}
\end{figure*}

\textbf{Case 3} As shown in Figure \ref{fig:error_case_Empty_Region_Artifacts}, this example illustrates a failure case where a large blank region appears at the bottom of the generated webpage. Compared with the original image, the generated result contains an extended empty area with no meaningful content. We find that this issue may stem from the prompt configuration, in which the original image size is provided as a reference. During HTML generation, the model attempts to align with the given page height and fills the remaining space with blank regions, even though padding the layout with empty space or repetitive content is explicitly discouraged and considered invalid in our benchmark setting.

\begin{figure*}
    \centering
    \includegraphics[width=0.95\linewidth]{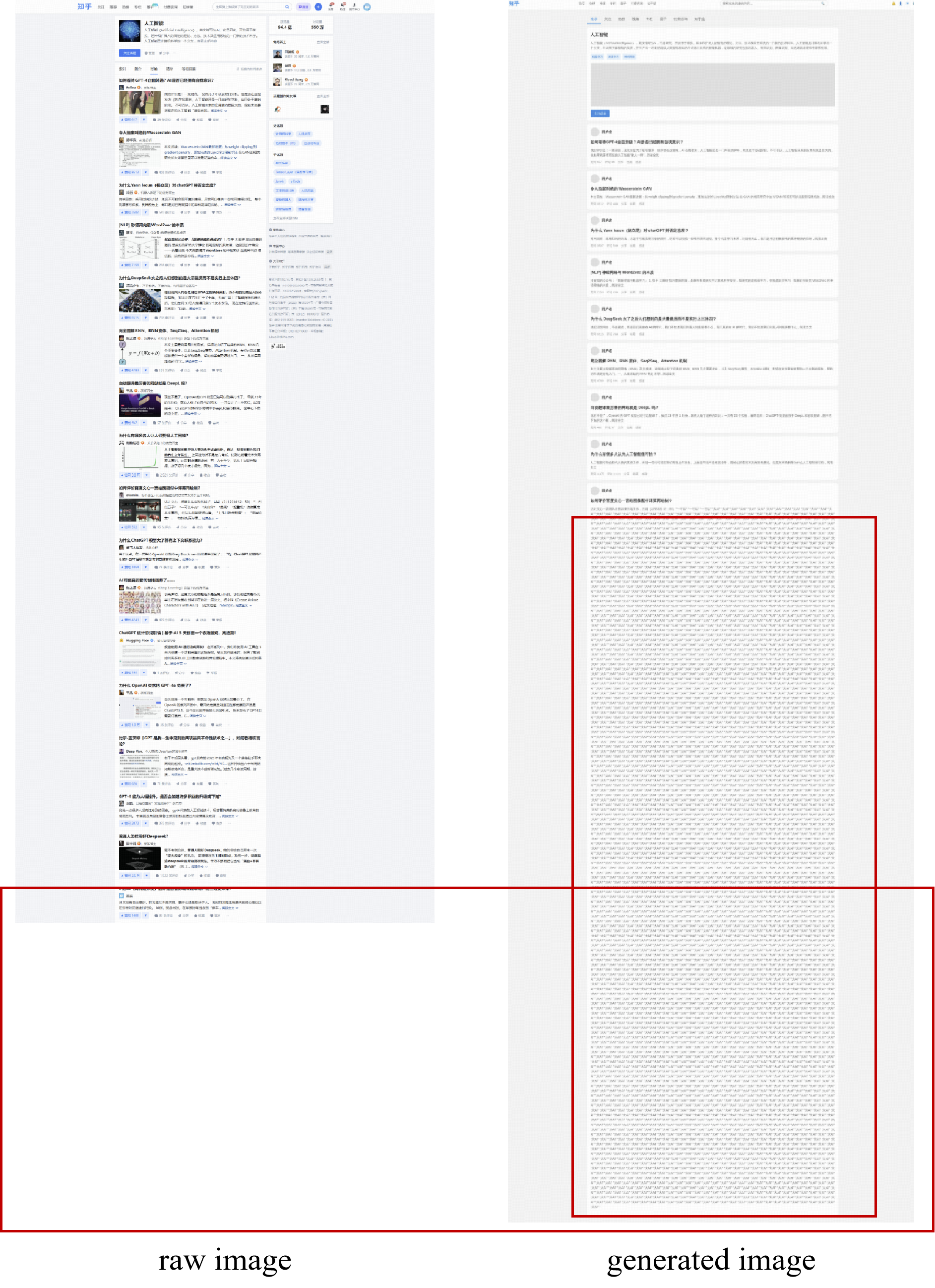}
    \caption{Example of Excessive Page Length in Generated Webpage}
    \label{fig:error_case_long_page}
\end{figure*}

\textbf{Case 4} As shown in Figure \ref{fig:error_case_long_page}, this example illustrates a failure case where the generated webpage is significantly longer than the original one. The excessive page length is primarily caused by repeated content blocks or redundant character sequences, resulting in unintended page expansion during HTML generation.

\begin{figure*}
    \centering
    \includegraphics[width=0.64\linewidth]{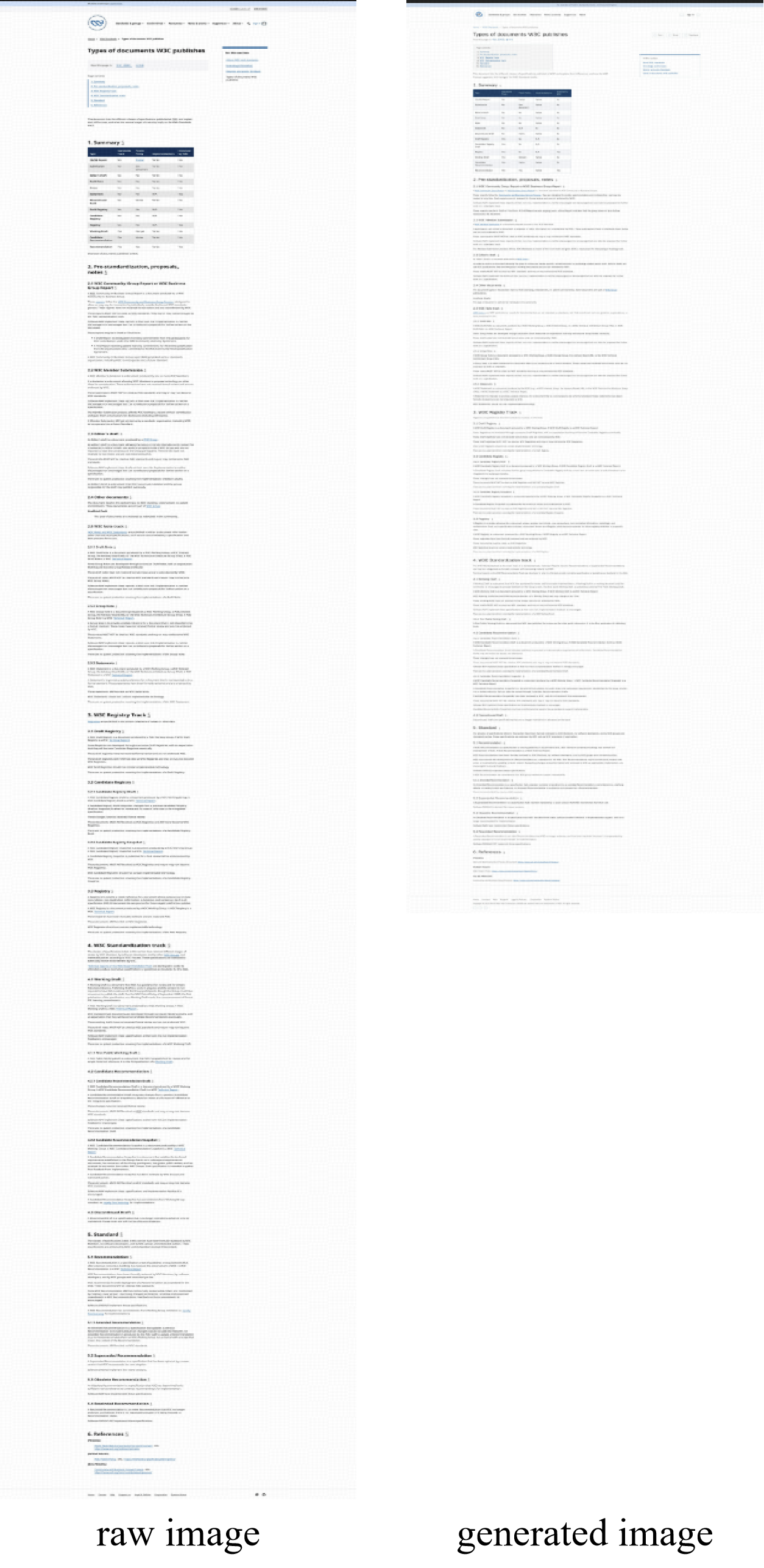}
    \caption{Example of Incomplete Content Coverage in Generated Webpage}
    \label{fig:error_case_short_page}
\end{figure*}

\textbf{Case 5} As shown in Figure \ref{fig:error_case_short_page}, this example illustrates a failure case where the generated webpage is complete in structure but shorter than the original one due to missing content . Compared with the original image, the generated result omits part of the page content, leading to a reduced overall page length. This issue may be attributed to the excessive length of the original webpage, which poses challenges for the model to capture and preserve all intermediate information during HTML generation.

\begin{figure*}
    \centering
    \includegraphics[width=1\linewidth]{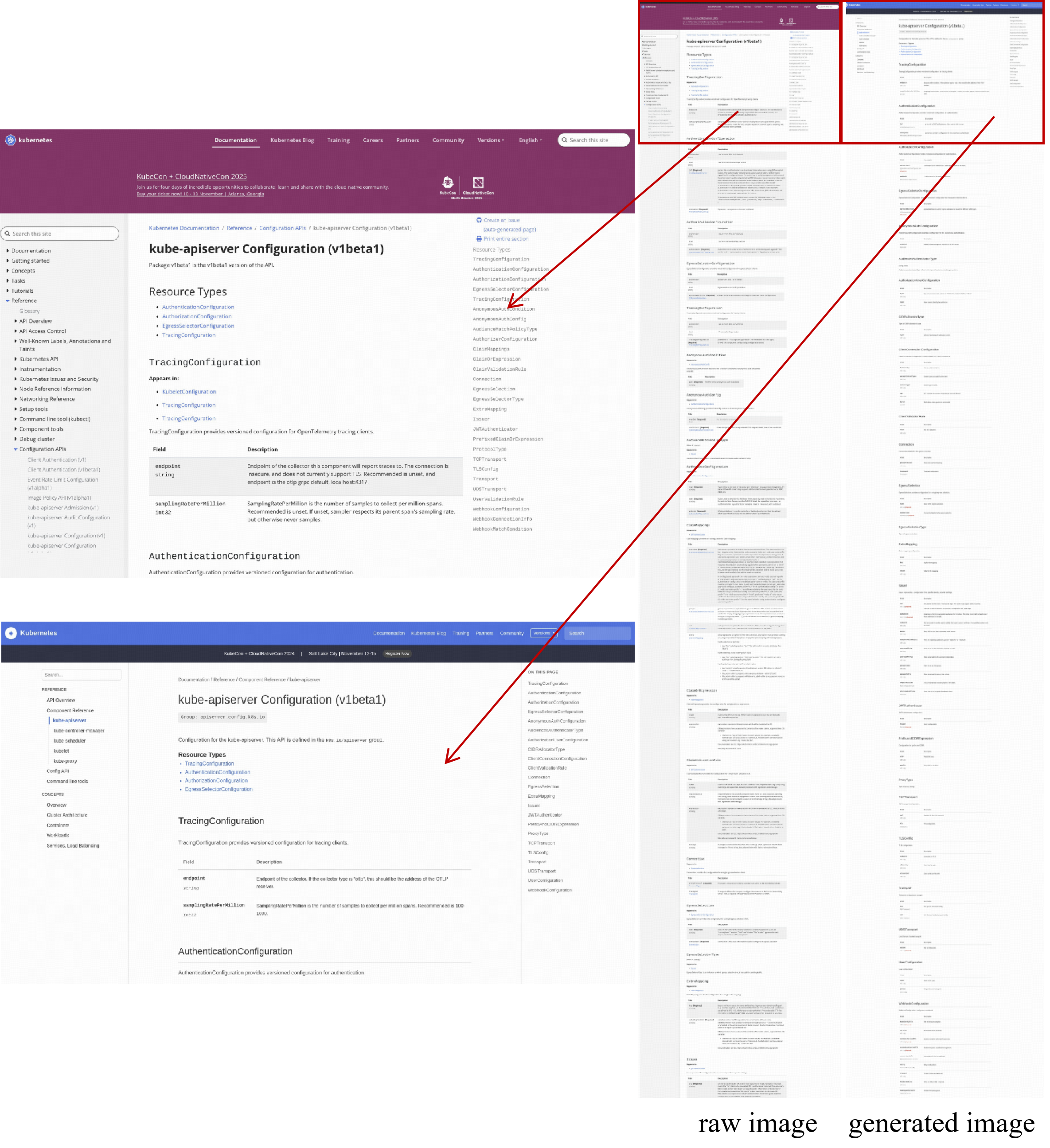}
    \caption{Example of Visual Style Inconsistency in Generated Webpage}
    \label{fig:error_case_Visual_Style_Deviation}
\end{figure*}

\textbf{Case 6} As shown in Figure  \ref{fig:error_case_Visual_Style_Deviation}, this example illustrates a failure case where the generated webpage exhibits noticeable visual style inconsistencies compared with the original page. While the overall page structure and content coverage are largely preserved, discrepancies can be observed in visual attributes such as color scheme, typography, spacing, and layout density across different regions of the page. These inconsistencies indicate that the model fails to maintain a coherent global visual style during HTML generation, leading to mismatched visual appearance between different sections.

\begin{figure*}
    \centering
    \includegraphics[width=0.86\linewidth]{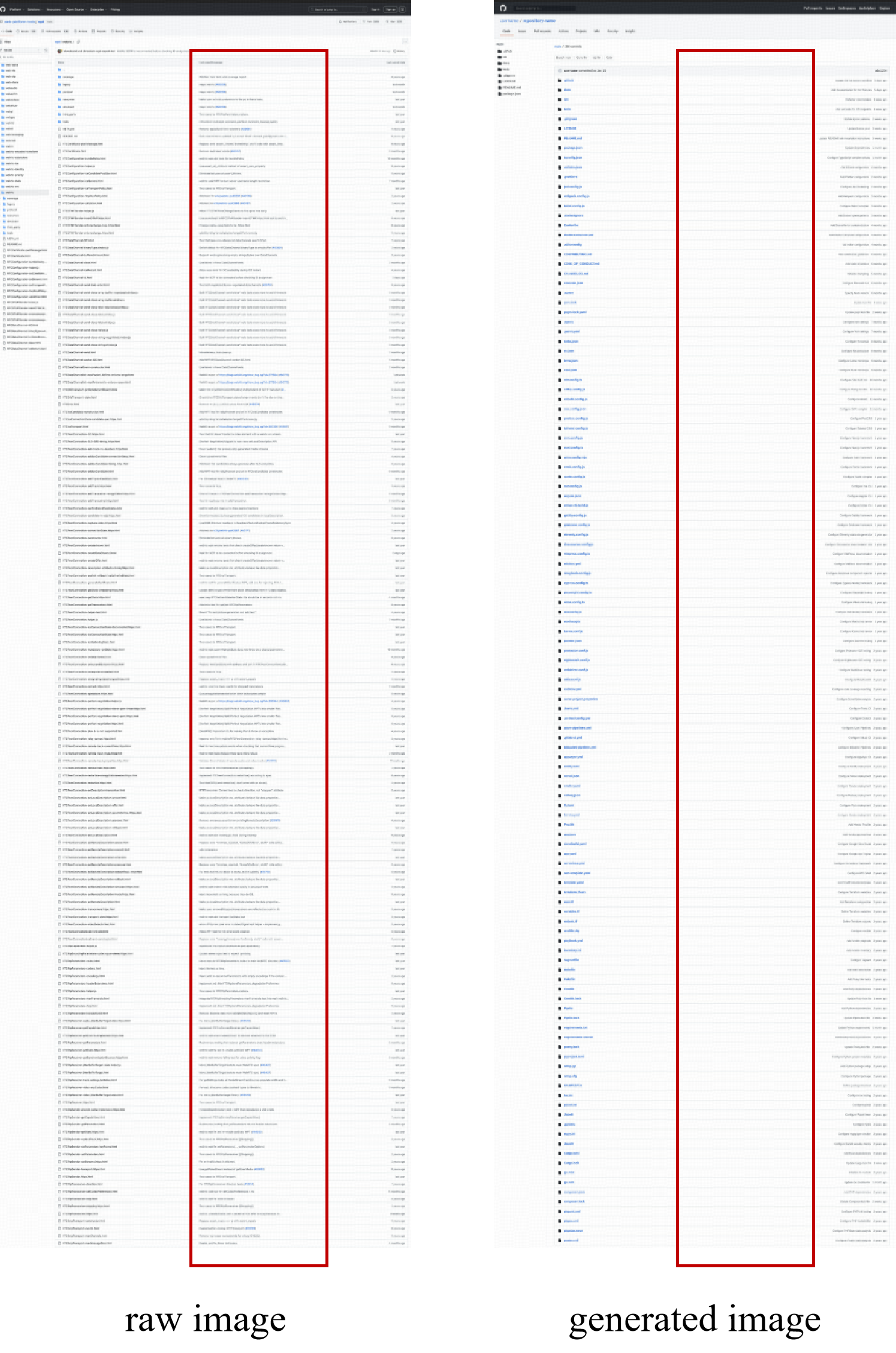}
    \caption{Example of Hierarchical Structure Mismatch in Generated Webpage}
    \label{fig:error_case_Hierarchy_Misalignment}
\end{figure*}

\textbf{Case 7} As shown in Figure \ref{fig:error_case_Hierarchy_Misalignment}, this example illustrates a failure case where the hierarchical structure of the generated webpage deviates from that of the original page. Although most content elements are present, their organizational hierarchy is incorrectly constructed, leading to misplaced or flattened sections in the generated result. In particular, items that should be nested under specific categories are rendered at the same level, resulting in a distorted structural layout.

\begin{figure*}
    \centering
    \includegraphics[width=1\linewidth]{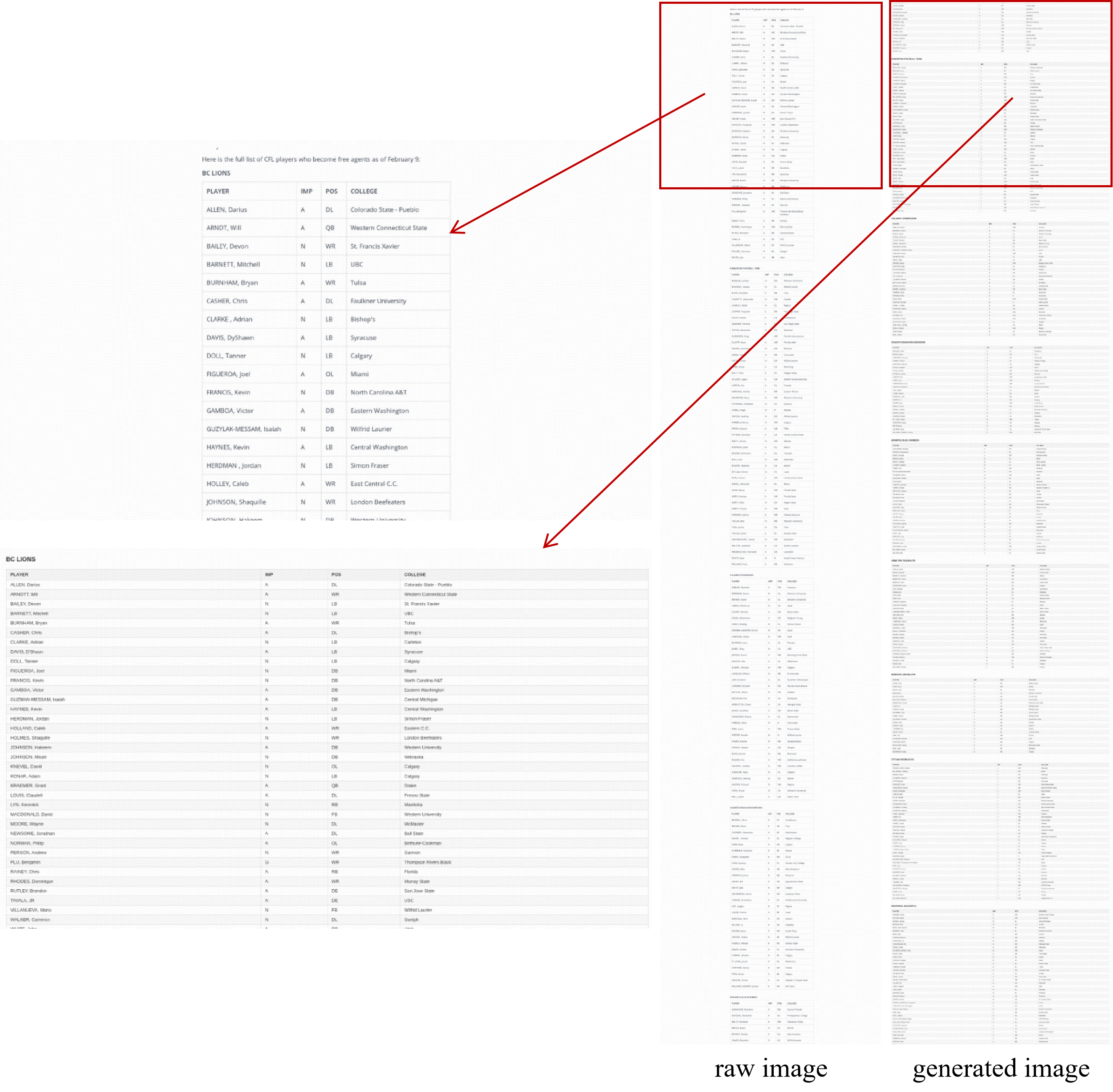}
    \caption{Example of Information Density Imbalance in Generated Webpage}
    \label{fig:error_case_Information_Density_Mismatch}
\end{figure*}

\textbf{Case 8} As shown in Figure \ref{fig:error_case_Information_Density_Mismatch}, this example illustrates a failure case where the information density of the generated webpage is diluted compared with the original page. While the original image maintains compact content organization with preserved blank regions, the generated result spreads information elements across the page, leading to reduced content concentration and degraded readability.

\begin{figure*}
    \centering
    \includegraphics[width=0.97\linewidth]{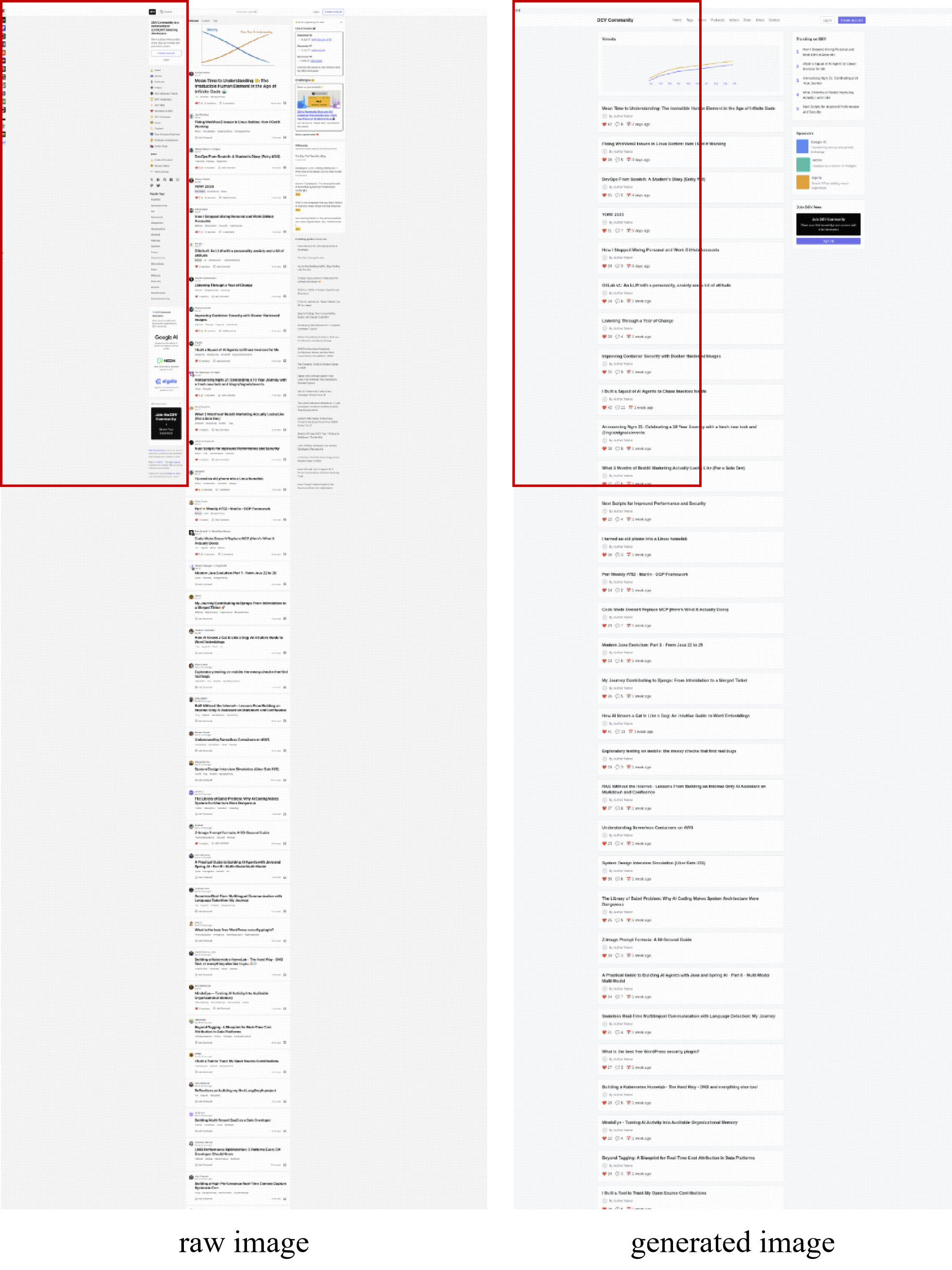}
    \caption{Example of Global Layout Deviation in Generated Webpage}
    \label{fig:error_case_Global_Layout_Drift}
\end{figure*}

\textbf{Case 9} As shown in Figure \ref{fig:error_case_Global_Layout_Drift}, this example illustrates a failure case where the global layout of the generated webpage deviates from the original design. The original page follows a three-column structure consisting of left, center, and right regions, whereas in the generated result, the right-side navigation panel is missing. This omission alters the overall page balance and structural organization, despite the main content being preserved.

\subsection{Case analysis for W-FFR}
\begin{figure*}
    \centering
    \includegraphics[width=1\linewidth]{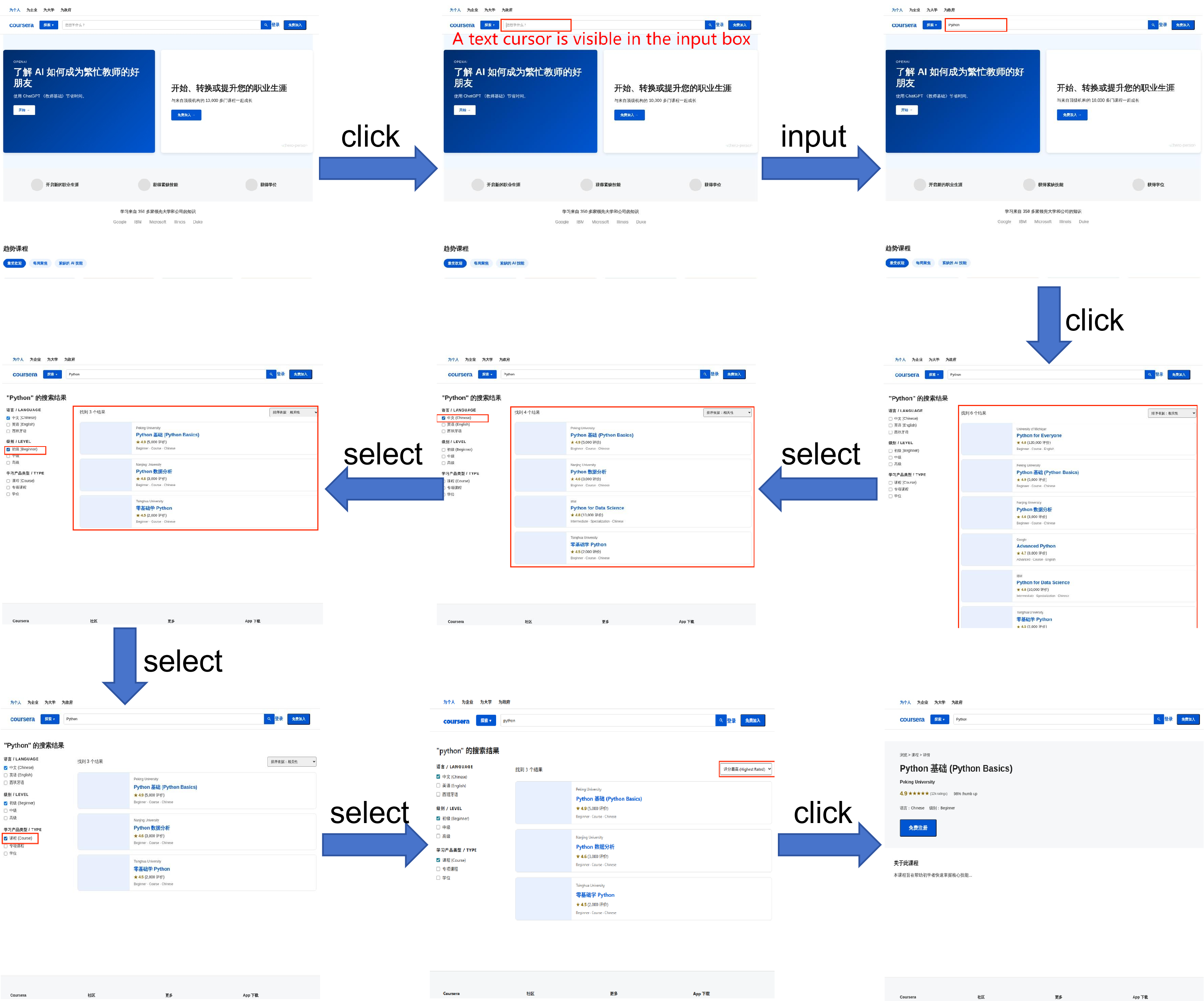}
    \caption{Success Case 1 for W-FFR}
    \label{fig:sucess}
\end{figure*}

\begin{figure*}
    \centering
    \includegraphics[width=1\linewidth]{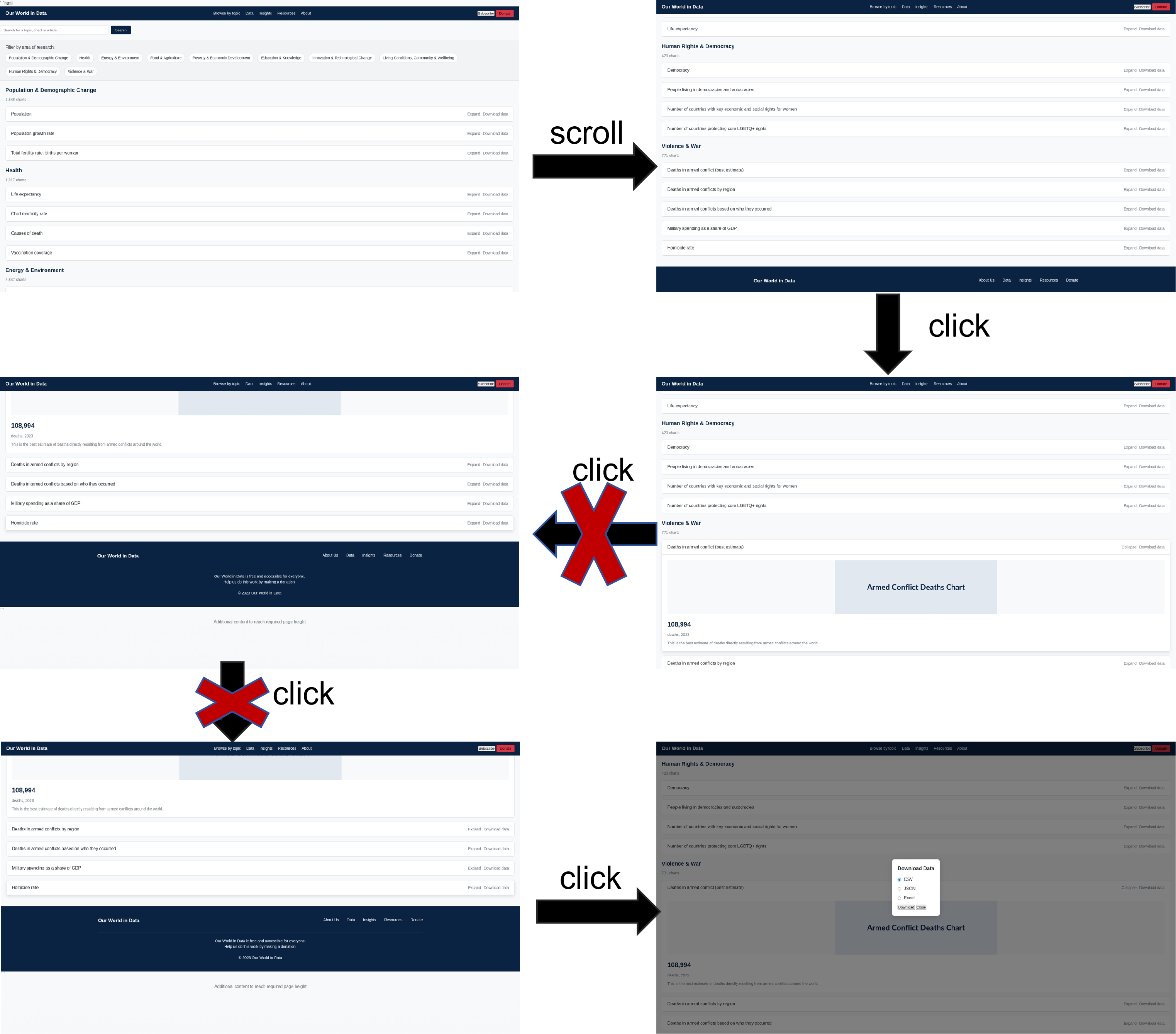}
    \caption{Failed Case 1 for W-FFR}
    \label{fig:failed1}
\end{figure*}

\begin{figure*}
    \centering
    \includegraphics[width=0.85\linewidth]{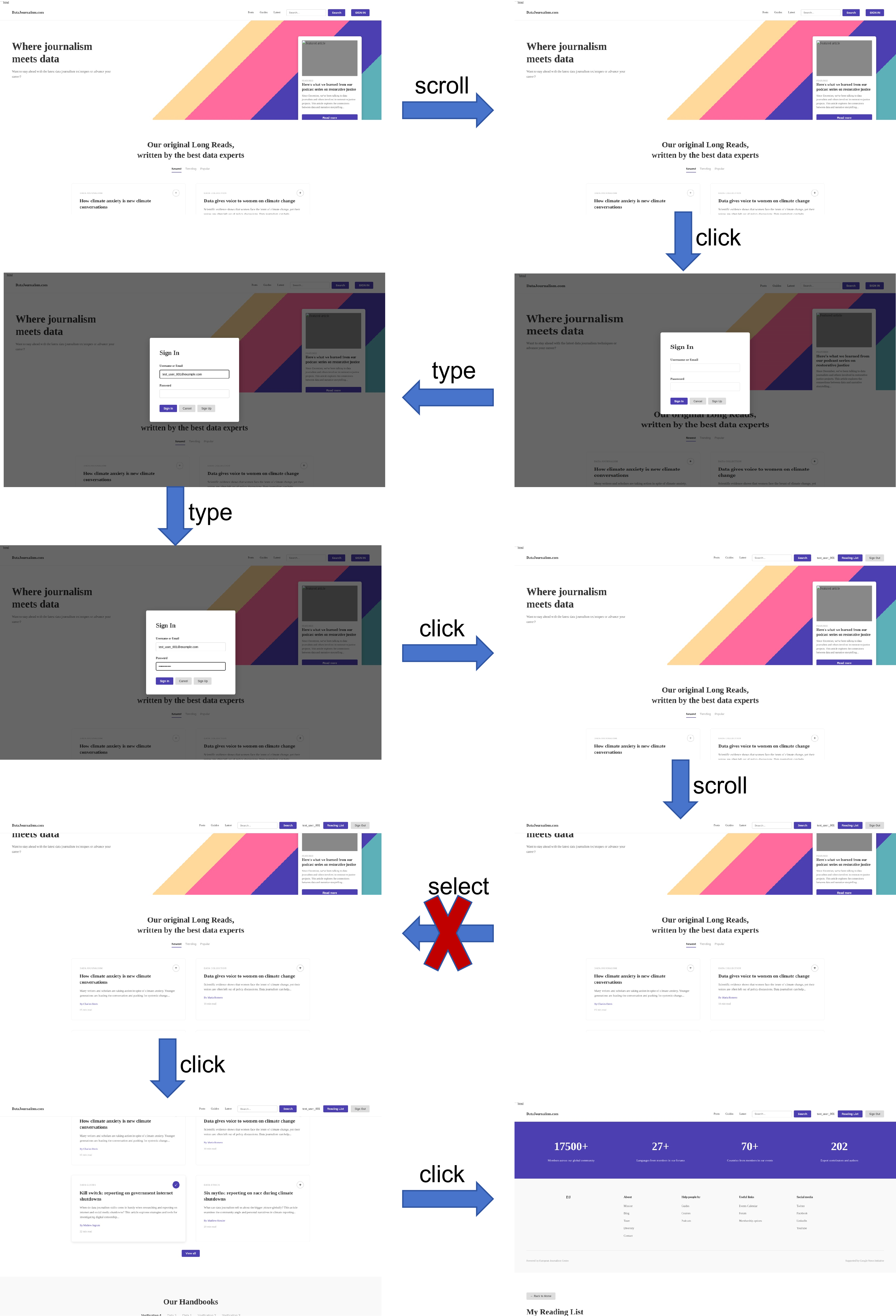}
    \caption{Failed Case 2 for W-FFR}
    \label{fig:failed2}
\end{figure*}

We further provide three illustrative examples for the W-FFR task. Figure \ref{fig:sucess} presents a case that is regarded as successful, while Figure \ref{fig:failed1},\ref{fig:failed2} illustrate two failure cases.

%% file: references.bib
@article{zhao2025chartcoder,
  title={Chartcoder: Advancing multimodal large language model for chart-to-code generation},
  author={Zhao, Xuanle and Luo, Xianzhen and Shi, Qi and Chen, Chi and Wang, Shuo and Liu, Zhiyuan and Sun, Maosong},
  journal={arXiv preprint arXiv:2501.06598},
  year={2025}
}

@article{wang2025internvl3,
  title={Internvl3. 5: Advancing open-source multimodal models in versatility, reasoning, and efficiency},
  author={Wang, Weiyun and Gao, Zhangwei and Gu, Lixin and Pu, Hengjun and Cui, Long and Wei, Xingguang and Liu, Zhaoyang and Jing, Linglin and Ye, Shenglong and Shao, Jie and others},
  journal={arXiv preprint arXiv:2508.18265},
  year={2025}
}

@misc{vteam2025glm45vglm41vthinkingversatilemultimodal,
      title={GLM-4.5V and GLM-4.1V-Thinking: Towards Versatile Multimodal Reasoning with Scalable Reinforcement Learning}, 
      author={V Team and Wenyi Hong and Wenmeng Yu and Xiaotao Gu and Guo Wang and Guobing Gan and Haomiao Tang and Jiale Cheng and Ji Qi and Junhui Ji and Lihang Pan and Shuaiqi Duan and Weihan Wang and Yan Wang and Yean Cheng and Zehai He and Zhe Su and Zhen Yang and Ziyang Pan and Aohan Zeng and Baoxu Wang and Bin Chen and Boyan Shi and Changyu Pang and Chenhui Zhang and Da Yin and Fan Yang and Guoqing Chen and Jiazheng Xu and Jiale Zhu and Jiali Chen and Jing Chen and Jinhao Chen and Jinghao Lin and Jinjiang Wang and Junjie Chen and Leqi Lei and Letian Gong and Leyi Pan and Mingdao Liu and Mingde Xu and Mingzhi Zhang and Qinkai Zheng and Sheng Yang and Shi Zhong and Shiyu Huang and Shuyuan Zhao and Siyan Xue and Shangqin Tu and Shengbiao Meng and Tianshu Zhang and Tianwei Luo and Tianxiang Hao and Tianyu Tong and Wenkai Li and Wei Jia and Xiao Liu and Xiaohan Zhang and Xin Lyu and Xinyue Fan and Xuancheng Huang and Yanling Wang and Yadong Xue and Yanfeng Wang and Yanzi Wang and Yifan An and Yifan Du and Yiming Shi and Yiheng Huang and Yilin Niu and Yuan Wang and Yuanchang Yue and Yuchen Li and Yutao Zhang and Yuting Wang and Yu Wang and Yuxuan Zhang and Zhao Xue and Zhenyu Hou and Zhengxiao Du and Zihan Wang and Peng Zhang and Debing Liu and Bin Xu and Juanzi Li and Minlie Huang and Yuxiao Dong and Jie Tang},
      year={2025},
      eprint={2507.01006},
      archivePrefix={arXiv},
      primaryClass={cs.CV},
      url={https://arxiv.org/abs/2507.01006}, 
}

@misc{openai2024gpt4technicalreport,
      title={GPT-4 Technical Report}, 
      author={OpenAI and Josh Achiam and Steven Adler and Sandhini Agarwal and Lama Ahmad and Ilge Akkaya and Florencia Leoni Aleman and Diogo Almeida and Janko Altenschmidt and Sam Altman and Shyamal Anadkat and Red Avila and Igor Babuschkin and Suchir Balaji and Valerie Balcom and Paul Baltescu and Haiming Bao and Mohammad Bavarian and Jeff Belgum and Irwan Bello and Jake Berdine and Gabriel Bernadett-Shapiro and Christopher Berner and Lenny Bogdonoff and Oleg Boiko and Madelaine Boyd and Anna-Luisa Brakman and Greg Brockman and Tim Brooks and Miles Brundage and Kevin Button and Trevor Cai and Rosie Campbell and Andrew Cann and Brittany Carey and Chelsea Carlson and Rory Carmichael and Brooke Chan and Che Chang and Fotis Chantzis and Derek Chen and Sully Chen and Ruby Chen and Jason Chen and Mark Chen and Ben Chess and Chester Cho and Casey Chu and Hyung Won Chung and Dave Cummings and Jeremiah Currier and Yunxing Dai and Cory Decareaux and Thomas Degry and Noah Deutsch and Damien Deville and Arka Dhar and David Dohan and Steve Dowling and Sheila Dunning and Adrien Ecoffet and Atty Eleti and Tyna Eloundou and David Farhi and Liam Fedus and Niko Felix and Simón Posada Fishman and Juston Forte and Isabella Fulford and Leo Gao and Elie Georges and Christian Gibson and Vik Goel and Tarun Gogineni and Gabriel Goh and Rapha Gontijo-Lopes and Jonathan Gordon and Morgan Grafstein and Scott Gray and Ryan Greene and Joshua Gross and Shixiang Shane Gu and Yufei Guo and Chris Hallacy and Jesse Han and Jeff Harris and Yuchen He and Mike Heaton and Johannes Heidecke and Chris Hesse and Alan Hickey and Wade Hickey and Peter Hoeschele and Brandon Houghton and Kenny Hsu and Shengli Hu and Xin Hu and Joost Huizinga and Shantanu Jain and Shawn Jain and Joanne Jang and Angela Jiang and Roger Jiang and Haozhun Jin and Denny Jin and Shino Jomoto and Billie Jonn and Heewoo Jun and Tomer Kaftan and Łukasz Kaiser and Ali Kamali and Ingmar Kanitscheider and Nitish Shirish Keskar and Tabarak Khan and Logan Kilpatrick and Jong Wook Kim and Christina Kim and Yongjik Kim and Jan Hendrik Kirchner and Jamie Kiros and Matt Knight and Daniel Kokotajlo and Łukasz Kondraciuk and Andrew Kondrich and Aris Konstantinidis and Kyle Kosic and Gretchen Krueger and Vishal Kuo and Michael Lampe and Ikai Lan and Teddy Lee and Jan Leike and Jade Leung and Daniel Levy and Chak Ming Li and Rachel Lim and Molly Lin and Stephanie Lin and Mateusz Litwin and Theresa Lopez and Ryan Lowe and Patricia Lue and Anna Makanju and Kim Malfacini and Sam Manning and Todor Markov and Yaniv Markovski and Bianca Martin and Katie Mayer and Andrew Mayne and Bob McGrew and Scott Mayer McKinney and Christine McLeavey and Paul McMillan and Jake McNeil and David Medina and Aalok Mehta and Jacob Menick and Luke Metz and Andrey Mishchenko and Pamela Mishkin and Vinnie Monaco and Evan Morikawa and Daniel Mossing and Tong Mu and Mira Murati and Oleg Murk and David Mély and Ashvin Nair and Reiichiro Nakano and Rajeev Nayak and Arvind Neelakantan and Richard Ngo and Hyeonwoo Noh and Long Ouyang and Cullen O'Keefe and Jakub Pachocki and Alex Paino and Joe Palermo and Ashley Pantuliano and Giambattista Parascandolo and Joel Parish and Emy Parparita and Alex Passos and Mikhail Pavlov and Andrew Peng and Adam Perelman and Filipe de Avila Belbute Peres and Michael Petrov and Henrique Ponde de Oliveira Pinto and Michael and Pokorny and Michelle Pokrass and Vitchyr H. Pong and Tolly Powell and Alethea Power and Boris Power and Elizabeth Proehl and Raul Puri and Alec Radford and Jack Rae and Aditya Ramesh and Cameron Raymond and Francis Real and Kendra Rimbach and Carl Ross and Bob Rotsted and Henri Roussez and Nick Ryder and Mario Saltarelli and Ted Sanders and Shibani Santurkar and Girish Sastry and Heather Schmidt and David Schnurr and John Schulman and Daniel Selsam and Kyla Sheppard and Toki Sherbakov and Jessica Shieh and Sarah Shoker and Pranav Shyam and Szymon Sidor and Eric Sigler and Maddie Simens and Jordan Sitkin and Katarina Slama and Ian Sohl and Benjamin Sokolowsky and Yang Song and Natalie Staudacher and Felipe Petroski Such and Natalie Summers and Ilya Sutskever and Jie Tang and Nikolas Tezak and Madeleine B. Thompson and Phil Tillet and Amin Tootoonchian and Elizabeth Tseng and Preston Tuggle and Nick Turley and Jerry Tworek and Juan Felipe Cerón Uribe and Andrea Vallone and Arun Vijayvergiya and Chelsea Voss and Carroll Wainwright and Justin Jay Wang and Alvin Wang and Ben Wang and Jonathan Ward and Jason Wei and CJ Weinmann and Akila Welihinda and Peter Welinder and Jiayi Weng and Lilian Weng and Matt Wiethoff and Dave Willner and Clemens Winter and Samuel Wolrich and Hannah Wong and Lauren Workman and Sherwin Wu and Jeff Wu and Michael Wu and Kai Xiao and Tao Xu and Sarah Yoo and Kevin Yu and Qiming Yuan and Wojciech Zaremba and Rowan Zellers and Chong Zhang and Marvin Zhang and Shengjia Zhao and Tianhao Zheng and Juntang Zhuang and William Zhuk and Barret Zoph},
      year={2024},
      eprint={2303.08774},
      archivePrefix={arXiv},
      primaryClass={cs.CL},
      url={https://arxiv.org/abs/2303.08774}, 
}

@article{tan2025chartmaster,
  title={Chartmaster: Advancing chart-to-code generation with real-world charts and chart similarity reinforcement learning},
  author={Tan, Wentao and Cao, Qiong and Xue, Chao and Zhan, Yibing and Ding, Changxing and He, Xiaodong},
  journal={arXiv preprint arXiv:2508.17608},
  year={2025}
}

@article{yang2025qwen3,
  title={Qwen3 technical report},
  author={Yang, An and Li, Anfeng and Yang, Baosong and Zhang, Beichen and Hui, Binyuan and Zheng, Bo and Yu, Bowen and Gao, Chang and Huang, Chengen and Lv, Chenxu and others},
  journal={arXiv preprint arXiv:2505.09388},
  year={2025}
}

@article{team2025kimi,
  title={Kimi-vl technical report},
  author={Team, Kimi and Du, Angang and Yin, Bohong and Xing, Bowei and Qu, Bowen and Wang, Bowen and Chen, Cheng and Zhang, Chenlin and Du, Chenzhuang and Wei, Chu and others},
  journal={arXiv preprint arXiv:2504.07491},
  year={2025}
}

@article{wang2024cogvlm,
  title={Cogvlm: Visual expert for pretrained language models},
  author={Wang, Weihan and Lv, Qingsong and Yu, Wenmeng and Hong, Wenyi and Qi, Ji and Wang, Yan and Ji, Junhui and Yang, Zhuoyi and Zhao, Lei and XiXuan, Song and others},
  journal={Advances in Neural Information Processing Systems},
  volume={37},
  pages={121475--121499},
  year={2024}
}

@inproceedings{beltramelli2018pix2code,
  title={pix2code: Generating code from a graphical user interface screenshot},
  author={Beltramelli, Tony},
  booktitle={Proceedings of the ACM SIGCHI symposium on engineering interactive computing systems},
  pages={1--6},
  year={2018}
}

@article{xiao2024interaction2code,
  title={Interaction2Code: Benchmarking MLLM-based Interactive Webpage Code Generation from Interactive Prototyping},
  author={Xiao, Jingyu and Wan, Yuxuan and Huo, Yintong and Wang, Zixin and Xu, Xinyi and Wang, Wenxuan and Xu, Zhiyao and Wang, Yuhang and Lyu, Michael R},
  journal={arXiv preprint arXiv:2411.03292},
  year={2024}
}

@article{zhou2023webarena,
  title={Webarena: A realistic web environment for building autonomous agents},
  author={Zhou, Shuyan and Xu, Frank F and Zhu, Hao and Zhou, Xuhui and Lo, Robert and Sridhar, Abishek and Cheng, Xianyi and Ou, Tianyue and Bisk, Yonatan and Fried, Daniel and others},
  journal={arXiv preprint arXiv:2307.13854},
  year={2023}
}

@article{singh2025openai,
  title={OpenAI GPT-5 System Card},
  author={Singh, Aaditya and Fry, Adam and Perelman, Adam and Tart, Adam and Ganesh, Adi and El-Kishky, Ahmed and McLaughlin, Aidan and Low, Aiden and Ostrow, AJ and Ananthram, Akhila and others},
  journal={arXiv preprint arXiv:2601.03267},
  year={2025}
}

@inproceedings{si2025design2code,
  title={Design2code: Benchmarking multimodal code generation for automated front-end engineering},
  author={Si, Chenglei and Zhang, Yanzhe and Li, Ryan and Yang, Zhengyuan and Liu, Ruibo and Yang, Diyi},
  booktitle={Proceedings of the 2025 Conference of the Nations of the Americas Chapter of the Association for Computational Linguistics: Human Language Technologies (Volume 1: Long Papers)},
  pages={3956--3974},
  year={2025}
}

@article{yun2024web2code,
  title={Web2code: A large-scale webpage-to-code dataset and evaluation framework for multimodal llms},
  author={Yun, Sukmin and Thushara, Rusiru and Bhat, Mohammad and Wang, Yongxin and Deng, Mingkai and Wang, Jinhong and Tao, Tianhua and Li, Junbo and Li, Haonan and Nakov, Preslav and others},
  journal={Advances in neural information processing systems},
  volume={37},
  pages={112134--112157},
  year={2024}
}

@inproceedings{gui2025webcode2m,
  title={Webcode2m: A real-world dataset for code generation from webpage designs},
  author={Gui, Yi and Li, Zhen and Wan, Yao and Shi, Yemin and Zhang, Hongyu and Chen, Bohua and Su, Yi and Chen, Dongping and Wu, Siyuan and Zhou, Xing and others},
  booktitle={Proceedings of the ACM on Web Conference 2025},
  pages={1834--1845},
  year={2025}
}

@article{lu2025webgen,
  title={WebGen-Bench: Evaluating LLMs on Generating Interactive and Functional Websites from Scratch},
  author={Lu, Zimu and Yang, Yunqiao and Ren, Houxing and Hou, Haotian and Xiao, Han and Wang, Ke and Shi, Weikang and Zhou, Aojun and Zhan, Mingjie and Li, Hongsheng},
  journal={arXiv preprint arXiv:2505.03733},
  year={2025}
}

@article{liu2023visual,
  title={Visual instruction tuning},
  author={Liu, Haotian and Li, Chunyuan and Wu, Qingyang and Lee, Yong Jae},
  journal={Advances in neural information processing systems},
  volume={36},
  pages={34892--34916},
  year={2023}
}

@inproceedings{niu2025chart2code53,
  title={Chart2Code53: A Large-Scale Diverse and Complex Dataset for Enhancing Chart-to-Code Generation},
  author={Niu, Tianhao and Cui, Yiming and Wang, Baoxin and Xu, Xiao and Yao, Xin and Zhu, Qingfu and Wu, Dayong and Wang, Shijin and Che, Wanxiang},
  booktitle={Proceedings of the 2025 Conference on Empirical Methods in Natural Language Processing},
  pages={15839--15855},
  year={2025}
}

@article{xu2025webvia,
  title={Webvia: A web-based vision-language agentic framework for interactive and verifiable ui-to-code generation},
  author={Xu, Mingde and Yang, Zhen and Hong, Wenyi and Pan, Lihang and Fan, Xinyue and Wang, Yan and Gu, Xiaotao and Xu, Bin and Tang, Jie},
  journal={arXiv preprint arXiv:2511.06251},
  year={2025}
}

@article{sun2025fullfront,
  title={FullFront: Benchmarking MLLMs Across the Full Front-End Engineering Workflow},
  author={Sun, Haoyu and Wang, Huichen Will and Gu, Jiawei and Li, Linjie and Cheng, Yu},
  journal={arXiv preprint arXiv:2505.17399},
  year={2025}
}

@article{deng2023mind2web,
  title={Mind2web: Towards a generalist agent for the web},
  author={Deng, Xiang and Gu, Yu and Zheng, Boyuan and Chen, Shijie and Stevens, Sam and Wang, Boshi and Sun, Huan and Su, Yu},
  journal={Advances in Neural Information Processing Systems},
  volume={36},
  pages={28091--28114},
  year={2023}
}

@article{chezelles2024browsergym,
  title={The browsergym ecosystem for web agent research},
  author={Chezelles, De and Le Sellier, Thibault and Shayegan, Sahar Omidi and Jang, Lawrence Keunho and L{\`u}, Xing Han and Yoran, Ori and Kong, Dehan and Xu, Frank F and Reddy, Siva and Cappart, Quentin and others},
  journal={arXiv preprint arXiv:2412.05467},
  year={2024}
}

@inproceedings{li2025sketch2code,
  title={Sketch2code: Evaluating vision-language models for interactive web design prototyping},
  author={Li, Ryan and Zhang, Yanzhe and Yang, Diyi},
  booktitle={Proceedings of the 2025 Conference of the Nations of the Americas Chapter of the Association for Computational Linguistics: Human Language Technologies (Volume 1: Long Papers)},
  pages={3921--3955},
  year={2025}
}

@article{comanici2025gemini,
  title={Gemini 2.5: Pushing the frontier with advanced reasoning, multimodality, long context, and next generation agentic capabilities},
  author={Comanici, Gheorghe and Bieber, Eric and Schaekermann, Mike and Pasupat, Ice and Sachdeva, Noveen and Dhillon, Inderjit and Blistein, Marcel and Ram, Ori and Zhang, Dan and Rosen, Evan and others},
  journal={arXiv preprint arXiv:2507.06261},
  year={2025}
}

@article{wu2025mllm,
  title={MLLM-Based UI2Code Automation Guided by UI Layout Information},
  author={Wu, Fan and Gao, Cuiyun and Li, Shuqing and Wen, Xin-Cheng and Liao, Qing},
  journal={Proceedings of the ACM on Software Engineering},
  volume={2},
  number={ISSTA},
  pages={1123--1145},
  year={2025},
  publisher={ACM New York, NY, USA}
}

@article{jiang2025screencoder,
  title={Screencoder: Advancing visual-to-code generation for front-end automation via modular multimodal agents},
  author={Jiang, Yilei and Zheng, Yaozhi and Wan, Yuxuan and Han, Jiaming and Wang, Qunzhong and Lyu, Michael R and Yue, Xiangyu},
  journal={arXiv preprint arXiv:2507.22827},
  year={2025}
}

@inproceedings{liang2025waffle,
  title={WAFFLE: Fine-tuning Multi-Modal Model for Automated Front-End Development},
  author={Liang, Shanchao and Jiang, Nan and Qian, Shangshu and Tan, Lin},
  booktitle={Proceedings of the 63rd Annual Meeting of the Association for Computational Linguistics (Volume 1: Long Papers)},
  pages={24786--24802},
  year={2025}
}
